\documentclass[runningheads]{llncs}

\usepackage{eccv}

\usepackage{eccvabbrv}

\usepackage{graphicx}
\usepackage{booktabs}

\usepackage[accsupp]{axessibility}  %

\usepackage{hyperref}

\usepackage{orcidlink}

\newcommand{\cR}{\mathcal{R}}
\newcommand{\cJ}{\mathcal{J}}
\newcommand{\cC}{\mathcal{C}}
\newcommand{\cQ}{\mathcal{Q}}

\usepackage{graphicx}
\usepackage{eso-pic}
\graphicspath{ {images/} }
\usepackage{multirow}
\usepackage{colortbl} 
\usepackage{multicol}
\usepackage{subcaption}

\definecolor{vlmcolor}{RGB}{217, 131, 36}
\definecolor{supervisedcolor}{RGB}{107, 142, 35}
\definecolor{probcolor}{RGB}{148, 113, 220}
\definecolor{supervisedanothercolor}{RGB}{10, 192, 235}
\definecolor{sslcolor}{RGB}{74, 119, 184}
\definecolor{vlmanothercolor}{RGB}{255, 105, 180}
\definecolor{vlmanothercolor2}{RGB}{32, 178, 170}

\definecolor{augcolor}{RGB}{102, 45, 145}
\definecolor{noaugcolor}{RGB}{0, 180, 180}

\definecolor{lightred}{RGB}{255, 0, 0}   
\definecolor{lightblue}{RGB}{54, 69, 79}
\definecolor{lightgreen}{RGB}{56, 118, 29}

\definecolor{raminuscolor}{RGB}{205, 92, 92}
\definecolor{rapluscolor}{RGB}{255, 255, 255}

\def\lsp{\hspace{12pt}}

\definecolor{appleblue}{RGB}{80,122,255}  %
\definecolor{applepink}{RGB}{217,127,174}  %
\definecolor{appleteal}{RGB}{85,163,152}  %
\definecolor{applepurple}{RGB}{138,107,221}  %
\definecolor{applemustard}{RGB}{234,179,77}  %
\definecolor{appleorange}{RGB}{255,127,80}  %
\definecolor{applegreen}{RGB}{52,199,89}
\definecolor{appleyellow}{RGB}{255,204,0}

\usepackage[many]{tcolorbox}

\tcbset{
    plotgroup/.style={
        enhanced,
        boxrule=0pt,      %
        frame hidden,     %
        arc=4pt,          %
        boxsep=2pt,       %
        left=2pt, right=2pt, top=2pt, bottom=2pt,
        nobeforeafter     %
    }
}

\usepackage{tikz}
\usetikzlibrary{arrows,shapes,calc,matrix,fit,backgrounds}
\usepackage{pgfplots}
\usepackage{pgfplotstable}
\pgfplotsset{compat=1.9}
\usepgfplotslibrary{fillbetween}
\usetikzlibrary{patterns}
\usetikzlibrary{calc}
\usetikzlibrary{overlay-beamer-styles}
\usepgfplotslibrary{external}
\usepackage[export]{adjustbox}
\AtBeginEnvironment{tikzpicture}{%
}
\usepackage{algorithm}
\usepackage{algpseudocode}
\usepackage{makecell}
\usepackage{pifont}

\begin{document}

\title{Invisible Shortcuts:\\ Why Vision Encoders Know Your Camera}

\titlerunning{Invisible Shortcuts}

\author{Vladan Stojnić\inst{1} \and
Ryan Ramos\inst{2}\and
Giorgos Kordopatis-Zilos\inst{1} \and \\
Noa Garcia\inst{2} \and Giorgos Tolias\inst{1}}

\authorrunning{V.~Stojnić et al.}

\institute{VRG, FEE, Czech Technical University in Prague \and
The University of Osaka}

\maketitle

\begin{center}
\includegraphics[width=0.90\textwidth]{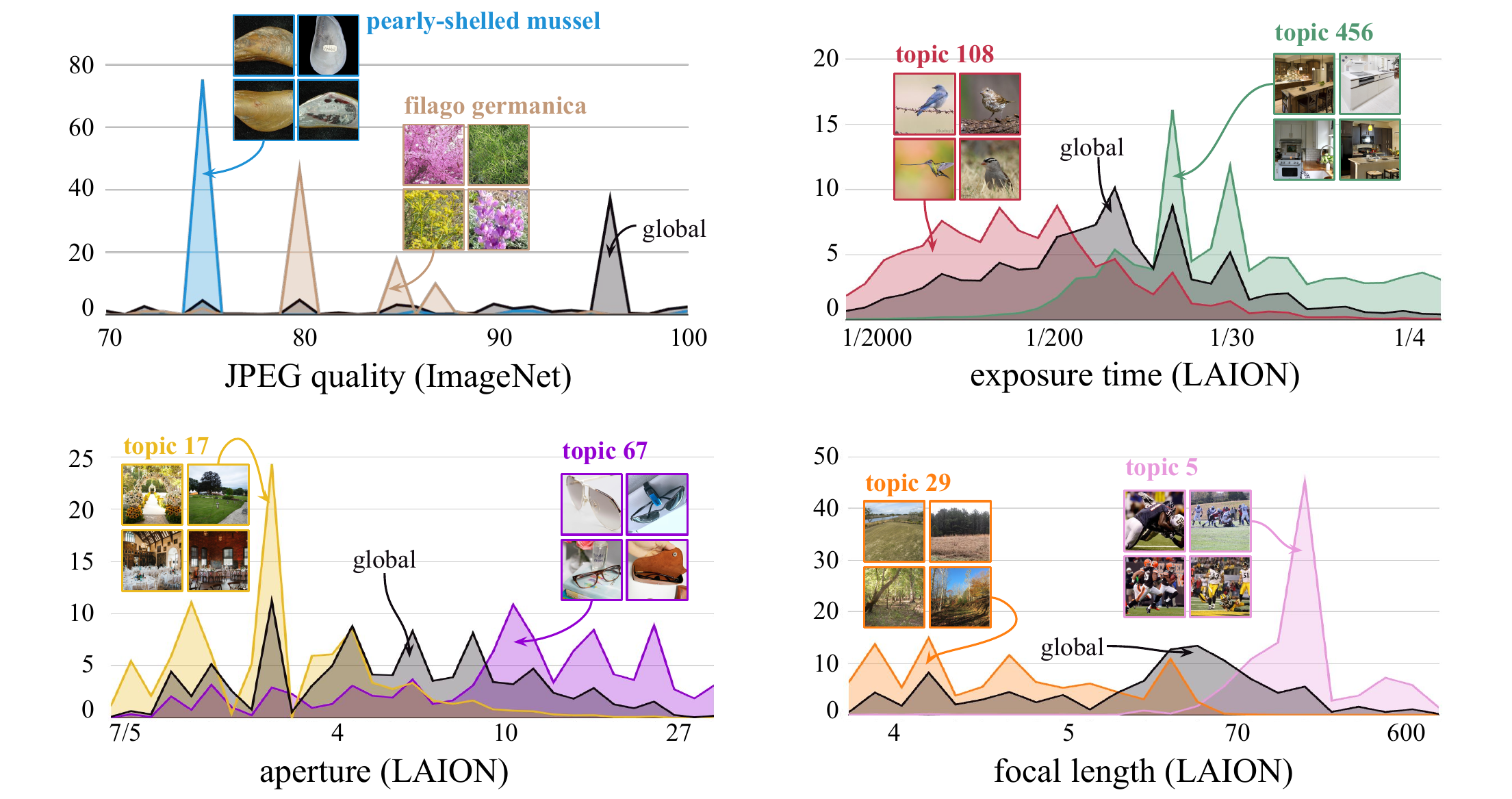}
\vspace{-8pt}
\captionof{figure}{\textbf{Metadata distribution per semantic category}. 
Different classes (ImageNet21k) and topics (Re-LAION-2B) have distinct distributions, indicating correlations between metadata and semantics. 
These correlations create shortcut signals that models exploit during large-scale pretraining. 
Global: overall dataset distribution.}
    \label{fig:fig1}
    \vspace{-8pt}
\end{center}

\begin{abstract}
Deep vision models exploit shortcuts, relying on cues that
correlate with supervision signals. Prior work has focused on visible biases, such as object-background or texture correlations. We identify a
different source of shortcut learning: invisible metadata traces embedded at the pixel level, for metadata such as image processing and photo acquisition. We hypothesize that large-scale semantic supervision, whether through categorical labels (ImageNet) or billion-scale captions (LAION), naturally
induces metadata-semantics correlations during pretraining, leading models to convert low-level signals into predictive features. 
By introducing controlled metadata-semantics correlations, we show that stronger ones produce  systematically higher sensitivity to metadata traces and larger performance degradation under metadata distribution shifts. We further explore mitigation strategies applied during
and after pretraining that reduce sensitivity not only to targeted metadata but also to unseen
ones,
without sacrificing performance on
downstream tasks. 
Metadata sensitivity also has a positive side: it partly explains the strong generated-image detection ability of some encoders, while its mitigation can improve out-of-distribution generalization.
Code: \url{https://github.com/ryan-caesar-ramos/visual-encoder-traces}

\end{abstract}

\newpage
\section{Introduction}
\label{sec:intro}
Deep neural networks are known to rely on shortcuts~\cite{geirhos2020shortcut,suhail2025shortcut,beery2018recognition,wang2024sober,shah2020pitfalls,zech2018variable}.
When a predictive cue is present in training data, models exploit it regardless of whether it aligns with the intended task, since the objective is simply to minimize error.

Whether this is harmful depends on the stability of these correlations; when they shift, performance degrades.
This phenomenon, studied under shortcut learning~\cite{geirhos2020shortcut} and spurious correlations~\cite{sagawa2020distributionally,liu2021just}, has been extensively analyzed in vision, including object–background dependencies~\cite{hendrycks2021natural,Neuhaus_2023_ICCV,singla2022salient}, color–label correlations~\cite{Meister_2023_ICCV}, texture versus shape bias~\cite{geirhos2018imagenet}, watermark artifacts~\cite{li2023whac}, and 
synthetic benchmarks~\cite{sagawa2020distributionally,InvariantRiskMinimization}.
These cues are visible and interpretable to humans. \looseness=-1

We study shortcuts largely invisible to the human eye, namely metadata traces, including processing parameters such as JPEG compression and sharpening, and acquisition parameters such as focal length and camera model. Although imperceptible at a semantic level, they leave low-level signals.

Recent work shows that distribution shifts in image metadata affect the predictions of pretrained vision models~\cite{ramos2025traces}, revealing that representations encode metadata traces, but the origin remains unclear.
We argue that metadata sensitivity arises from metadata--semantics correlations in pretraining datasets: images from certain semantic categories are often captured or processed under similar conditions, creating systematic associations between labels or captions and metadata attributes such as camera parameters or JPEG quality (\cref{fig:fig1}).
Models trained with semantic supervision exploit these correlations as shortcuts, relying on metadata-derived pixel signals rather than semantic content, as such signals are simpler than high-level concepts and optimization favors them~\cite{shah2020pitfalls}.

We test this hypothesis across both regimes of large-scale semantic supervision: categorical label supervision on ImageNet and caption supervision on LAION, measuring the metadata-semantics correlations in each and introducing controlled ones during training. As correlation strength increases, models become progressively more sensitive to metadata. Motivated by this, we investigate mitigation both during and after pretraining. Interestingly, using a single attribute, \eg JPEG compression, as the correlation source or mitigation target also changes sensitivity to other metadata, \eg resizing or camera model, suggesting that models exploit a broad class of low-level traces rather than attribute-specific artifacts. Notably, this contradicts prior work~\cite{li2023whac}, where mitigating one type of bias amplified others.

Our main contribution is to show that metadata sensitivity is a result of shortcut learning arising from metadata-semantics correlations in training data, manifesting at scale across both supervised ImageNet and caption-guided LAION pretraining, rather than being specific to a single supervision paradigm. We further show it can be mitigated, both during and after pretraining, with mitigation targeting one attribute generalizing to unseen ones. Finally, such sensitivity is not %
only a disadvantage: since it reflects a model's tendency to capture pixel-level signal, which is key to separating generated from real images, it partly explains the strong generated-image detection ability of some encoders~\cite{ojha2023towards}. Meanwhile its mitigation can improve out-of-distribution generalization.

\section{Diagnostic metrics}
\label{sec:preliminaries}
We use metrics that quantify (i) the extent to which frozen model representations encode metadata and (ii) the degree to which such encoding interferes with semantic prediction ability. 
We follow Ramos \etal~\cite{ramos2025traces}, see original work for details. 
We consider two metadata categories:
(i) \emph{image processing attributes}: JPEG compression parameters (quality, chroma subsampling), sharpening (strength), resizing (factor), and interpolation (algorithm type), and
(ii) \emph{acquisition attributes}: camera make/model, exposure time, aperture, ISO, and focal length.
\looseness=-1

\noindent\textbf{Metadata Prediction (MP).}
We measure the strength of metadata encoding in frozen model representations by training a linear classifier to predict metadata labels.
For processing metadata, we use ImageNet1k (IN1k)~\cite{russakovsky15imagenet}, randomly assign processing labels to training and test images, and process them accordingly.
For acquisition metadata, we use FlickrExif~\cite{ramos2025traces} with predefined train/test splits per acquisition attribute.
Accuracy is denoted by MP$_\text{p}$ and MP$_\text{a}$ for processing and acquisition metadata, respectively.
Accuracy substantially above chance indicates that the representation contains linearly separable metadata information, while accuracy close to chance suggests invariance to metadata labels.

\noindent\textbf{Semantic Prediction (SP)}
ability is evaluated by the accuracy of a $k$-NN classifier predicting semantic labels with original IN1k train/val sets.

\noindent\textbf{Semantic Prediction Distraction -- processing (SPD$_\text{p}$).}
To quantify the impact of processing-related metadata on semantic predictions, we use IN1k with a $k$-NN classifier for semantic labels, where metadata labels of training/exemplar images are modified per test image guided by semantic labels. Images are processed according to the assigned metadata label.
Two evaluation setups:
(i) \emph{Pos-same}: the test image and its semantic positives are assigned to the same processing label, while negatives are assigned to a different one. Accuracy in this setup is denoted by $A_{\text{pos-same}}$.
(ii) \emph{Neg-same}: the test image and its semantic negatives are assigned to the same processing label, while positives are assigned to a different one. Accuracy in this setup is denoted by $A_{\text{neg-same}}$.
We evaluate
distraction with
$\Delta_p = |A_{\text{pos-same}} - A_{\text{neg-same}}|$ quantifying the extent to which processing metadata impacts semantic recognition.
Lower values indicate stronger invariance to processing metadata, while higher values indicate stronger sensitivity.

\noindent\textbf{Semantic Prediction Distraction -- acquisition (SPD$_\text{a}$).}
To measure how acquisition metadata interferes with semantic similarity, we use the PairCams~\cite{ramos2025traces} dataset.
Given a query image from one camera class \eg smartphones, the task is to retrieve a semantically positive image captured with a different camera class \eg digital cameras.
Using recall@1, denoted by $R$, we evaluate two cases where semantically negative images vary:
(i) \emph{same}: negatives originate from the same camera class as the query; performance denoted by $R_{\text{same}}$, and
(ii) \emph{different}: negatives originate from a different camera class than the query; performance denoted by  $R_{\text{diff}}$.
We define the distraction score as $\Delta_a = |R_{\text{same}} - R_{\text{diff}}|$, quantifying the extent to which acquisition metadata impacts semantic recognition.

\noindent\textbf{Overall.} MP quantifies metadata encoding, SPD quantifies metadata interference with semantic recognition, and SP captures whether training data correlations or mitigation strategies degrade semantic prediction performance.

\section{Why are vision encoders sensitive to image metadata}
\label{sec:causes}

The work of Ramos \etal \cite{ramos2025traces} shows that vision encoders are sensitive to image metadata, and that the presence of metadata traces in features of pretrained models has consequences on downstream performance. However, the causes of these traces remain unclear.
We hypothesize that they stem from correlations between metadata and semantic labels in pretraining datasets: such correlations let metadata labels and their pixel-level signal act as a shortcut for predicting semantic labels, leading the model to encode metadata information.
We corroborate our hypothesis in several steps.
We first show that these correlations,
with both processing and acquisition metadata,
exist across standard pretraining datasets, spanning 
categorical (ImageNet) and caption-based (LAION) supervision.
We then use controlled experiments to verify that an encoder's metadata sensitivity grows as a direct function of correlation strength.

\subsection{Correlations of semantics and processing metadata in ImageNet}

\noindent\textbf{Measuring existing correlations.}
We focus on JPEG compression parameters, specifically the quality and chroma subsampling, as processing metadata. Each combination of values for these two parameters defines a discrete categorical label for each image.
We analyze the association between these metadata and semantic labels in ImageNet1k (IN1k) and ImageNet21k (IN21k) using Cramér's V~\cite{cramer1999mathematical} measure of association between categorical variables.
The values are 0.047 and  0.067 in IN1k and IN21k, respectively.
Even though the value itself is low and not easily interpretable, we focus on the relative comparison between the two datasets; correlations are stronger in IN21k than in IN1k.

\begin{figure}[th]
    \centering
    \input{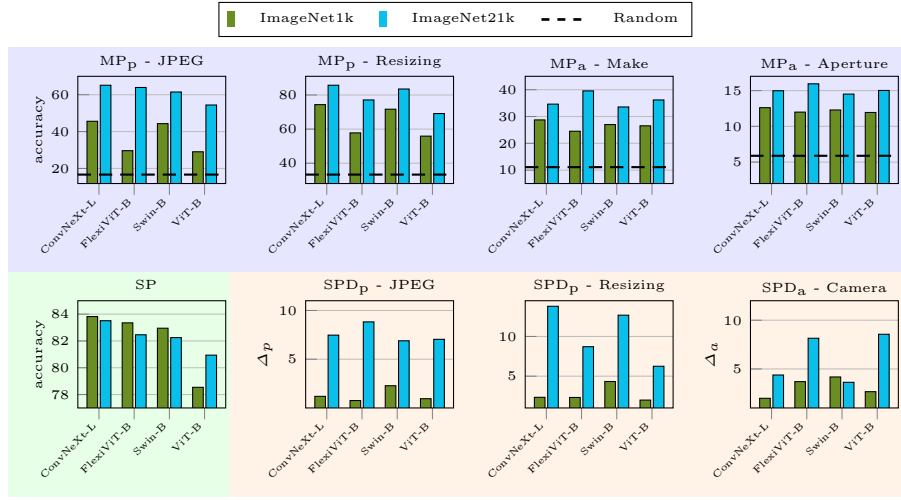}
\pgfplotsset{
    mybarstyle/.style={
        ybar=1pt, 
        bar width=4pt,
        width=0.29\linewidth,
        height=3cm,
        enlarge x limits=0.2,
        symbolic x coords={ConvNeXt-L, FlexiViT-B, Swin-B, ViT-B},
        xtick=data,
        x tick label style={rotate=45, anchor=east, font=\tiny, scale=0.75},
        title style={font=\tiny, yshift=-7pt},
        label style={font=\tiny},
        tick label style={font=\tiny},
        y label style={yshift=-4pt},
        ymajorgrids=true,
        xtick pos=bottom,
        ytick pos=left,
    },
    imagenet1k/.style={restrict expr to domain={mod(\coordindex,2)}{0:0}},
    imagenet21k/.style={restrict expr to domain={mod(\coordindex,2)}{1:1}}
}

\pgfplotsset{
    ybar legend/.style={
        legend image code/.code={
            \fill[#1] (0pt,-3pt) rectangle (3pt,3pt);
        }
    }
}

\begin{tikzpicture}
    \begin{axis}[
        hide axis,
        scale only axis,
        height=0cm,
        width=0cm,
        legend style={
            draw=black,
            cells={anchor=west},
            legend columns=3,
            column sep=0.8em,
            font=\tiny,
            anchor=north west
        },
        legend entries={
            {ImageNet1k},
            {ImageNet21k},
            {Random}
        }
    ]
        \addlegendimage{ybar, fill=supervisedcolor, draw=none, ybar legend}
        \addlegendimage{ybar, fill=supervisedanothercolor, draw=none, ybar legend}
        \addlegendimage{color=black, line width=1, dashed}
        
        \addplot[draw=none] coordinates {(0,0)};
    \end{axis}
\end{tikzpicture}
\vspace{3pt}

\centering
    \begin{tabular}{cccc}
    
        \cellcolor{blue!10}
        
        \begin{tikzpicture}
            \begin{axis}[mybarstyle, title={MP$_\text{p}$ - JPEG}, ylabel={accuracy}]
            \addplot[fill=supervisedcolor, draw=black] table[x=model, y=jpeg, imagenet1k] {\imagenetcomparison};
            \addplot[fill=supervisedanothercolor, draw=black] table[x=model, y=jpeg, imagenet21k] {\imagenetcomparison};
            \addplot[
                black, 
                dashed, 
                thick, 
                sharp plot, %
                dash pattern=on 5pt off 3pt,
                mark=none,  %
                shorten <=-8pt, %
                shorten >=-8pt  %
            ] coordinates {(ConvNeXt-L, 16.67) (ViT-B, 16.67)};
            \end{axis}
        \end{tikzpicture} &

        \cellcolor{blue!10}
        \begin{tikzpicture}
            \begin{axis}[mybarstyle, title={MP$_\text{p}$ - Resizing}, ylabel={\phantom{$\Delta_p$}}]
            \addplot[fill=supervisedcolor, draw=black] table[x=model, y=resizing, imagenet1k] {\imagenetcomparison};
            \addplot[fill=supervisedanothercolor, draw=black] table[x=model, y=resizing, imagenet21k] {\imagenetcomparison};
            \addplot[
                black, 
                dashed, 
                thick, 
                sharp plot,
                dash pattern=on 5pt off 3pt,
                mark=none,
                shorten <=-8pt,
                shorten >=-8pt 
            ] coordinates {(ConvNeXt-L, 33.33) (ViT-B, 33.33)};
            \end{axis}
        \end{tikzpicture} &

        \cellcolor{blue!10}
        \begin{tikzpicture}
            \begin{axis}[mybarstyle, title={MP$_\text{a}$ - Make}, ymin=5, ymax=45]
            \addplot[fill=supervisedcolor, draw=black] table[x=model, y=make, imagenet1k] {\imagenetcomparison};
            \addplot[fill=supervisedanothercolor, draw=black] table[x=model, y=make, imagenet21k] {\imagenetcomparison};
            \addplot[
                black, 
                dashed, 
                thick, 
                sharp plot, %
                dash pattern=on 5pt off 3pt,
                mark=none,  %
                shorten <=-8pt, %
                shorten >=-8pt  %
            ] coordinates {(ConvNeXt-L, 11.11) (ViT-B, 11.11)};
            \end{axis}
        \end{tikzpicture} &

        \cellcolor{blue!10}
        \begin{tikzpicture}
            \begin{axis}[mybarstyle, title={MP$_\text{a}$ - Aperture}, ymin=2, ymax=17, ylabel={\phantom{$\Delta_a$}}]
            \addplot[fill=supervisedcolor, draw=black] table[x=model, y=aperture, imagenet1k] {\imagenetcomparison};
            \addplot[fill=supervisedanothercolor, draw=black] table[x=model, y=aperture, imagenet21k] {\imagenetcomparison};
            \addplot[
                black, 
                dashed, 
                thick, 
                sharp plot, %
                dash pattern=on 5pt off 3pt,
                mark=none,  %
                shorten <=-8pt, %
                shorten >=-8pt  %
            ] coordinates {(ConvNeXt-L, 5.88) (ViT-B, 5.88)};
            \end{axis}
        \end{tikzpicture} \\

        \cellcolor{green!10}
        \begin{tikzpicture}
            \begin{axis}[mybarstyle, title={SP\phantom{$_\text{p}$}}, ylabel={accuracy}, ymin=77, ymax=85]
            \addplot[fill=supervisedcolor, draw=black] table[x=model, y=val-knn, imagenet1k] {\imagenetcomparison};
            \addplot[fill=supervisedanothercolor, draw=black] table[x=model, y=val-knn, imagenet21k] {\imagenetcomparison};
            \end{axis}
        \end{tikzpicture} &

        \cellcolor{orange!10}
        \begin{tikzpicture}
            \begin{axis}[mybarstyle, title={SPD$_\text{p}$ - JPEG}, ylabel={$\Delta_p$}, ymin=0, ymax=11, ytick={5, 10}]
            \addplot[fill=supervisedcolor, draw=black] table[x=model, y=jpeg-sem, imagenet1k] {\imagenetcomparison};
            \addplot[fill=supervisedanothercolor, draw=black] table[x=model, y=jpeg-sem, imagenet21k] {\imagenetcomparison};
            \end{axis}
        \end{tikzpicture} &

        \cellcolor{orange!10}
        \begin{tikzpicture}
            \begin{axis}[mybarstyle, title={SPD$_\text{p}$ - Resizing}, ymin=1, ymax=14.5]
            \addplot[fill=supervisedcolor, draw=black] table[x=model, y=resizing-sem, imagenet1k] {\imagenetcomparison};
            \addplot[fill=supervisedanothercolor, draw=black] table[x=model, y=resizing-sem, imagenet21k] {\imagenetcomparison};
            \end{axis}
        \end{tikzpicture}  &

        \cellcolor{orange!10}
        \begin{tikzpicture}
            \begin{axis}[mybarstyle, title={SPD$_\text{a}$ - Camera}, ylabel={$\Delta_a$}, ymin=1, ymax=12]
            \addplot[fill=supervisedcolor, draw=black] table[x=model, y=spd-a, imagenet1k] {\imagenetcomparison};
            \addplot[fill=supervisedanothercolor, draw=black] table[x=model, y=spd-a, imagenet21k] {\imagenetcomparison};
            \end{axis}
        \end{tikzpicture} \\

    \end{tabular}
    \caption{\textbf{IN1k \vs IN21k pretraining: which training set causes higher metadata sensitivity?} 
    Reporting Metadata Prediction (\textcolor{blue!30}{\textbf{MP}}), Semantic Prediction (\textcolor{green!30}{\textbf{SP}}), and Semantic Prediction Distraction (\textcolor{orange!30}{\textbf{SPD}}) for ConvNeXt~\cite{liu2022convnet}, FlexiViT~\cite{beyer23flexivit}, Swin~\cite{liu21swin}, and ViT~\cite{dosovitskiy2020image} pretrained on IN1k and IN21k.
    The amount of correlation between JPEG compression parameters and semantic labels is measured to be 0.047 and 0.067 in IN1k and IN21k, respectively, according to Cramér's V measure.
    The stronger correlations in IN21k result in consistently more sensitive models.}
    \label{fig:imagenet_models}
\end{figure}

\noindent\textbf{Impact of correlations.}
We use models\footnote{We use pretrained models available in the \href{https://github.com/huggingface/pytorch-image-models}{timm} library.} pretrained on IN1k and IN21k to evaluate the impact of correlation strength. 
In \cref{fig:imagenet_models}, we present all the diagnostic metrics, SP, MP, and SPD, for different backbones.
We observe that both types of models have a good ability to predict metadata labels (MP), both processing and acquisition ones, at rates substantially above random chance.
While there exist minor differences in training recipes,
the models trained on IN21k consistently demonstrate stronger predictive ability.
This indicates that features trained on IN21k encode more metadata information than those trained on IN1k.
The SPD measurements further show that metadata affects the semantic prediction abilities of both models, but more so for IN21k.
Together, these results show that IN21k features not only encode more metadata information but also let it interfere more strongly with semantic prediction.
Overall, these results suggest that the stronger correlations between metadata and semantic labels in IN21k, relative to IN1k, have a greater impact on both the models' ability to encode metadata information and the sensitivity of their semantic predictions to metadata.

\subsection{Controlling correlations of semantics and processing metadata}
\label{sec:processing_correlations}
To further verify that the presence and extent of metadata-semantics correlations impacts model's sensitivity to metadata, we synthetically introduce correlations into the IN1k training set, train models on these correlated datasets, and evaluate all diagnostic metrics.\looseness=-1

\noindent\textbf{Creating stronger correlations.}
We define 22 distinct metadata labels by combining 11 JPEG quality levels with 2 chroma subsampling modes.
All images are then recompressed according to one of these labels, creating a correlation between semantic labels and metadata labels. 
We randomly assign one specific JPEG compression, \ie metadata label, to each semantic class.
We introduce a probability $p_i$ that determines whether an image is processed using the metadata label assigned to its semantic class or using a randomly sampled metadata label.
When $p_i=100$, correlation is maximal, while when $p_i=0$, metadata is independent of semantics.
Note that Cramér's V is roughly equal to the value of $p_i$, \eg 0.1 and 0.8 for $p_i=10$ and $p_i=80$, respectively.
\looseness=-1

\begin{figure}[t]
\vspace{-5pt}
    \centering
    \input{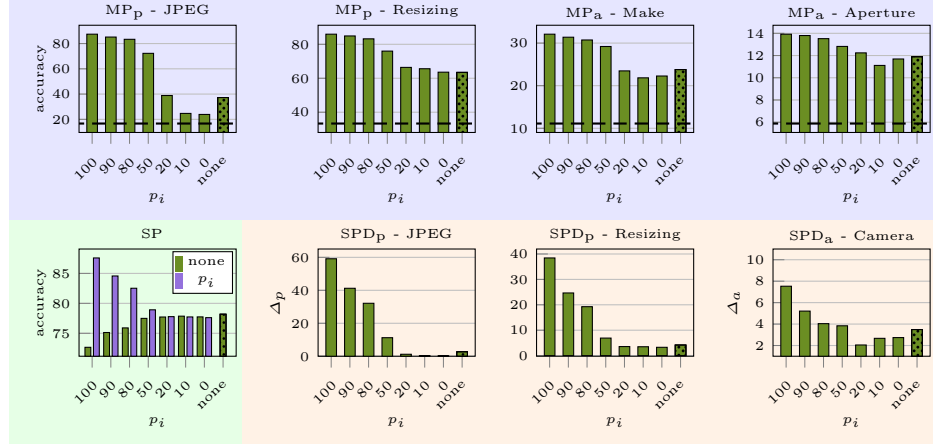}
\pgfplotsset{
    mybarstyle2/.style={
        ybar=1pt, 
        bar width=4pt,
        width=0.3\linewidth,
        height=3cm,
        enlarge x limits=0.1,
        xtick=data,
        xticklabels from table={\resnetimagenetjpegprobimageinversion}{p},
        x tick label style={font=\tiny, rotate=45, anchor=north east},
        title style={font=\tiny, yshift=-7pt},
        label style={font=\tiny},
        tick label style={font=\tiny},
        y label style={yshift=-4pt},
        x label style={yshift=5pt},
        ymajorgrids=true,
        xtick pos=bottom,
        ytick pos=left,
    }
}

\centering
    \begin{tabular}{cccc}

        \cellcolor{blue!10}
        \begin{tikzpicture}
            \begin{axis}[mybarstyle2, title={MP$_\text{p}$ - JPEG}, ylabel={accuracy}, xlabel={$p_{i}$}]
            \addplot[fill=supervisedcolor, draw=black, bar shift=0pt] table[x expr=\coordindex, y=jpeg] {\resnetimagenetjpegprobimageinversion};
            \addplot[fill=supervisedcolor, draw=black, postaction={pattern=crosshatch dots}, bar shift=0pt, restrict x to domain=6.5:7.5] table[x expr=\coordindex, y=jpeg] {\resnetimagenetjpegprobimageinversion};
            \addplot[
                black, 
                dashed, 
                thick, 
                sharp plot, %
                dash pattern=on 5pt off 3pt,
                mark=none,  %
                shorten <=-8pt, %
                shorten >=-8pt  %
            ] coordinates {(0, 16.67) (7, 16.67)};
            \end{axis}
        \end{tikzpicture} &

        \cellcolor{blue!10}
        \begin{tikzpicture}
            \begin{axis}[mybarstyle2, title={MP$_\text{p}$ - Resizing}, xlabel={$p_{i}$}, ylabel={\phantom{$\Delta_p$}}]
            \addplot[fill=supervisedcolor, draw=black, bar shift=0pt] table[x expr=\coordindex, y=resizing] {\resnetimagenetjpegprobimageinversion};
            \addplot[fill=supervisedcolor, draw=black, postaction={pattern=crosshatch dots}, bar shift=0pt, restrict x to domain=6.5:7.5] table[x expr=\coordindex, y=resizing] {\resnetimagenetjpegprobimageinversion};
            \addplot[
                black, 
                dashed, 
                thick, 
                sharp plot, %
                dash pattern=on 5pt off 3pt,
                mark=none,  %
                shorten <=-8pt, %
                shorten >=-8pt  %
            ] coordinates {(0, 33.33) (7, 33.33)};
            \end{axis}
        \end{tikzpicture} &

        \cellcolor{blue!10}
        \begin{tikzpicture}
            \begin{axis}[mybarstyle2, title={MP$_\text{a}$ - Make}, xlabel={$p_{i}$}]
            \addplot[fill=supervisedcolor, draw=black, bar shift=0pt] table[x expr=\coordindex, y=make] {\resnetimagenetjpegprobimageinversion};
            \addplot[fill=supervisedcolor, draw=black, postaction={pattern=crosshatch dots}, bar shift=0pt, restrict x to domain=6.5:7.5] table[x expr=\coordindex, y=make] {\resnetimagenetjpegprobimageinversion};
            \addplot[
                black, 
                dashed, 
                thick, 
                sharp plot, %
                dash pattern=on 5pt off 3pt,
                mark=none,  %
                shorten <=-8pt, %
                shorten >=-8pt  %
            ] coordinates {(0, 11.11) (7, 11.11)};
            \end{axis}
        \end{tikzpicture} &

        \cellcolor{blue!10}
        \begin{tikzpicture}
            \begin{axis}[mybarstyle2, title={MP$_\text{a}$ - Aperture}, xlabel={$p_{i}$}, ylabel={\phantom{$\Delta_a$}}]
            \addplot[fill=supervisedcolor, draw=black, bar shift=0pt] table[x expr=\coordindex, y=aperture] {\resnetimagenetjpegprobimageinversion};
            \addplot[fill=supervisedcolor, draw=black, postaction={pattern=crosshatch dots}, bar shift=0pt, restrict x to domain=6.5:7.5] table[x expr=\coordindex, y=aperture] {\resnetimagenetjpegprobimageinversion};
            \addplot[
                black, 
                dashed, 
                thick, 
                sharp plot, %
                dash pattern=on 5pt off 3pt,
                mark=none,  %
                shorten <=-8pt, %
                shorten >=-8pt  %
            ] coordinates {(0, 5.88) (7, 5.88)};
            \end{axis}
        \end{tikzpicture} \\

        \cellcolor{green!10}
        \begin{tikzpicture}
            \begin{axis}[mybarstyle2, title={SP\phantom{$_\text{p}$}}, ylabel={accuracy}, xlabel={$p_{i}$}, bar width=2.2pt, bar shift=0pt, legend style={font=\tiny, scale=0.6, /tikz/every even column/.append style={column sep=0.1cm}, inner sep=0.8pt, row sep=-1pt}, legend image code/.code={
                    \draw[#1, draw=none] (0cm,-0.1cm) rectangle (0.1cm,0.1cm); %
                }]
            \addplot[draw=none, fill=none, bar shift=0pt, forget plot] table[x expr=\coordindex, y=val-knn] {\resnetimagenetjpegprobimageinversion};
            \addplot[fill=supervisedcolor, draw=black, bar shift=-1.6pt, restrict x to domain=-0.5:6.5] table[x expr=\coordindex, y=val-knn] {\resnetimagenetjpegprobimageinversion};\addlegendentry{none};
            \addplot[fill=probcolor, draw=black, bar shift=1.6pt, restrict x to domain=-0.5:6.5] table[x expr=\coordindex, y=val-bias]{\resnetimagenetjpegprobimageinversion};\addlegendentry{$p_i$};
            \addplot[fill=supervisedcolor, draw=black, bar shift=0pt, restrict x to domain=6.5:7.5, forget plot] table[x expr=\coordindex, y=val-knn] {\resnetimagenetjpegprobimageinversion};
            \addplot[fill=supervisedcolor, draw=black, postaction={pattern=crosshatch dots}, bar shift=0pt, restrict x to domain=6.5:7.5, forget plot] table[x expr=\coordindex, y=val-knn] {\resnetimagenetjpegprobimageinversion};
            \end{axis}
        \end{tikzpicture} &

        \cellcolor{orange!10}
        \begin{tikzpicture}
            \begin{axis}[mybarstyle2, title={SPD$_\text{p}$ - JPEG}, ylabel={$\Delta_p$}, xlabel={$p_{i}$}, ymin=0]
            \addplot[fill=supervisedcolor, draw=black, bar shift=0pt] table[x expr=\coordindex, y=jpeg] {\resnetimagenetjpegprobimagesemantics};
            \addplot[fill=supervisedcolor, draw=black, postaction={pattern=crosshatch dots}, bar shift=0pt, restrict x to domain=6.5:7.5] table[x expr=\coordindex, y=jpeg] {\resnetimagenetjpegprobimagesemantics};
            \end{axis}
        \end{tikzpicture} &

        \cellcolor{orange!10}
        \begin{tikzpicture}
            \begin{axis}[mybarstyle2, title={SPD$_\text{p}$ - Resizing}, xlabel={$p_{i}$}]
            \addplot[fill=supervisedcolor, draw=black, bar shift=0pt] table[x expr=\coordindex, y=resizing] {\resnetimagenetjpegprobimagesemantics};
            \addplot[fill=supervisedcolor, draw=black, postaction={pattern=crosshatch dots}, bar shift=0pt, restrict x to domain=6.5:7.5] table[x expr=\coordindex, y=resizing] {\resnetimagenetjpegprobimagesemantics};
            \end{axis}
        \end{tikzpicture} &

        \cellcolor{orange!10}
        \begin{tikzpicture}
            \begin{axis}[mybarstyle2, title={SPD$_\text{a}$ - Camera}, ylabel={$\Delta_a$}, xlabel={$p_{i}$}, ymin=1, ymax=11]
            \addplot[fill=supervisedcolor, draw=black, bar shift=0pt] table[x expr=\coordindex, y=spd-a] {\resnetimagenetjpegprobimageinversion};
            \addplot[fill=supervisedcolor, draw=black, postaction={pattern=crosshatch dots}, bar shift=0pt, restrict x to domain=6.5:7.5] table[x expr=\coordindex, y=spd-a] {\resnetimagenetjpegprobimageinversion};
            \end{axis}
        \end{tikzpicture} \\

    \end{tabular}
\vspace{-8pt}    
    \caption{\textbf{Impact of correlations between semantics and processing metadata on the model's sensitivity to metadata.
    } ResNet50 trained on different versions of IN1k with 
    manipulated JPEG labels. 
    The larger the $p_i$, the stronger the correlation between JPEG labels and semantic labels. 
    \emph{None}: model trained on original IN1k. 
    Difference between $p_i=0$ and \emph{none}: the former has no JPEG-semantics correlations, while the latter has the original IN1k correlations.
    SP: two types of test sets are used, the original one (none, \textcolor{supervisedcolor}{green} color) which is fixed across values of $p_i$, and 7 different versions (\textcolor{probcolor}{purple} color) created by manipulating the JPEG labels of the original test set in the same way as the corresponding training set for each value of $p_i$.
\vspace{-20pt}    
    }
    \label{fig:imagenet_image_prob}
\end{figure}

\noindent\textbf{Impact of correlations.}
We train ResNet50~\cite{he2016deep}\footnote{ViT-S/16~\cite{dosovitskiy2020image}, and ViT-B/16~\cite{dosovitskiy2020image} results are presented in the supplementary.} on these versions of the IN1k training set created using the mechanism described above, and evaluate the models using the diagnostic metrics.
\cref{fig:imagenet_image_prob}
presents the results, where the larger values of $p_i$ mean stronger correlation between metadata and semantics.

We observe that accuracy of predicting JPEG metadata labels (MP) increases noticeably for stronger correlations. For zero correlations it is close to random chance, and lower than for the model trained on the original IN1k,
verifying
that the original IN1k has some correlations, though weaker than the ones we introduce with $p_i\geq 20$.
Since
the model is trained on a set where JPEG compression drives the correlation, its ability to predict JPEG labels is 
expected. More notably,
the ability to predict other processing and acquisition attributes (resizing, make, aperture) also increases with the JPEG-semantics correlation, suggesting that the model learns to encode pixel-level signal carrying such metadata.

The semantic prediction ability (SP) is affected by the correlation strength (\textcolor{supervisedcolor}{green} bars of bottom-left plot).
There is no noticeable drop for weak correlations compared to zero correlations, but as the correlation becomes stronger, the drop becomes more significant.
This suggests that 
strong correlations make the model capture pixel-level signal related to metadata labels rather than complex semantic-level features.
Performance is additionally reported for testing on 7 modified versions (\textcolor{probcolor}{purple} bars of bottom-left plot) of the IN1k validation, each processed using the same metadata-to-semantics associations as the training set for the corresponding value of $p_i$. 
Observe that when evaluating on these transformed test sets, the accuracy increases for increased correlations; this is because the test set is transformed identically to the training set. The model uses the same pixel-level signal that it exploited during training, which is not the case when testing on the original set.

Finally, we observe that the distraction of semantic prediction abilities by metadata (SPD) also increases with the increase in correlation, which suggests that the model relies more on the metadata information for semantic prediction when the correlation is stronger, and thus the interference of metadata information with semantic prediction becomes stronger as well.

\subsection{Correlations of semantics and acquisition metadata in LAION}
We additionally investigate whether correlations exist between semantics and acquisition metadata.
The subset of Re-LAION-2B~\cite{relaion} carrying Exif tags, about 40M images, provides an image set annotated with acquisition metadata labels.
Since Re-LAION-2B contains captions rather than categorical class labels, we apply topic modeling~\cite{grootendorst2022bertopic} 
to cluster caption embeddings and assign each caption to one of the resulting $6,238$ topics, which serves as a discrete semantic label per image.
\cref{fig:fig1} illustrates examples of such correlations, where for a given acquisition attribute, we plot the frequency distribution of its metadata labels within Re-LAION-2B for two topics with clearly distinct distributions.
For instance, images whose captions fall under the topic of football tend to have longer focal lengths than those under plots of land. Similar distinctions arise between weddings and eyewear for aperture, and between birds and kitchens for exposure time.
We presume that such correlations can act as a source of shortcut learning, analogous to the processing metadata case, which we verify in the following.

\subsection{Controlling correlations of semantics and acquisition metadata}
\label{sec:causes_laion}

\noindent\textbf{Creating stronger and weaker correlations.}
We create three Re-LAION-2B subsets of equal size, \ie 6.4M images, with varying amounts of correlation between semantic topics and acquisition metadata labels.
Metadata labels are defined as tuples comprising the labels for the following attributes: camera manufacturer, aperture, exposure, and ISO speed.
(i) \emph{baseline}: we sample uniformly, reflecting the natural correlations present in Re-LAION-2B.
(ii) \emph{stronger}: for each metadata label, we select images belonging to its 40 most frequent topics, concentrating each label on a few topics and thus strengthening the topic--metadata association.
(iii) \emph{weaker}: for each metadata label, we identify its least frequent topic and downsample all other topics to that size, equalizing the topic distribution within each metadata label and thus weakening the association.

\begin{figure}[t]
    \centering
    \input{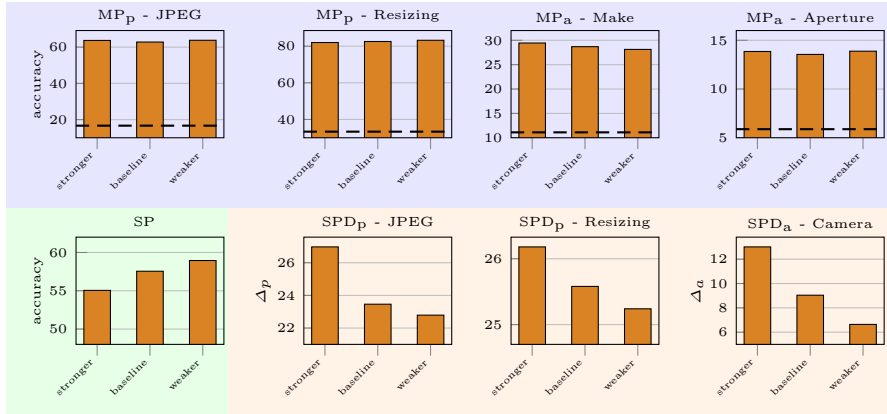}
\pgfplotsset{
    mybarstyle/.style={
        ybar=1pt, 
        width=0.29\linewidth,
        height=3cm,
        enlarge x limits=0.2,
        ymin=0, 
        symbolic x coords={stronger, baseline, weaker},
        xtick=data,
        x tick label style={rotate=45, anchor=east, font=\tiny, scale=0.75},
        title style={font=\tiny, yshift=-7pt},
        label style={font=\tiny},
        tick label style={font=\tiny},
        y label style={yshift=-4pt},
        ymajorgrids=true,
        xtick pos=bottom,
        ytick pos=left,
    },
    stronger/.style={restrict expr to domain={mod(\coordindex,3)}{0:0}},
    baseline/.style={restrict expr to domain={mod(\coordindex,3)}{1:1}},
    weaker/.style={restrict expr to domain={mod(\coordindex,3)}{2:2}},
}

\centering
    \begin{tabular}{cccc}
    
        \cellcolor{blue!10}
        
        \begin{tikzpicture}
            \begin{axis}[mybarstyle, title={MP$_\text{p}$ - JPEG}, ylabel={accuracy}, ymin=10]
            \addplot[fill=vlmcolor, draw=black] table[x=subset, y=jpeg] {\customclipcomparison};
            \addplot[
                black, 
                dashed, 
                thick, 
                sharp plot, %
                dash pattern=on 5pt off 3pt,
                mark=none,  %
                shorten <=-8pt, %
                shorten >=-8pt  %
            ] coordinates {(stronger, 16.67) (weaker, 16.67)};
            \end{axis}
        \end{tikzpicture} &

        \cellcolor{blue!10}
        \begin{tikzpicture}
            \begin{axis}[mybarstyle, title={MP$_\text{p}$ - Resizing}, ylabel={\phantom{$\Delta_p$}}, ymin=30]
            \addplot[fill=vlmcolor, draw=black] table[x=subset, y=resizing] {\customclipcomparison};
            \addplot[
                black, 
                dashed, 
                thick, 
                sharp plot,
                dash pattern=on 5pt off 3pt,
                mark=none,
                shorten <=-8pt,
                shorten >=-8pt 
            ] coordinates {(stronger, 33.33) (weaker, 33.33)};
            \end{axis}
        \end{tikzpicture} &

        \cellcolor{blue!10}
        \begin{tikzpicture}
            \begin{axis}[mybarstyle, title={MP$_\text{a}$ - Make}, ymin=10, ymax=32]
            \addplot[fill=vlmcolor, draw=black] table[x=subset, y=make] {\customclipcomparison};
            \addplot[
                black, 
                dashed, 
                thick, 
                sharp plot, %
                dash pattern=on 5pt off 3pt,
                mark=none,  %
                shorten <=-8pt, %
                shorten >=-8pt  %
            ] coordinates {(stronger, 11.11) (weaker, 11.11)};
            \end{axis}
        \end{tikzpicture} &

        \cellcolor{blue!10}
        \begin{tikzpicture}
            \begin{axis}[mybarstyle, title={MP$_\text{a}$ - Aperture}, ymin=5, ymax=16, ylabel={\phantom{$\Delta_a$}}]
            \addplot[fill=vlmcolor, draw=black] table[x=subset, y=aperture] {\customclipcomparison};
            \addplot[
                black, 
                dashed, 
                thick, 
                sharp plot, %
                dash pattern=on 5pt off 3pt,
                mark=none,  %
                shorten <=-8pt, %
                shorten >=-8pt  %
            ] coordinates {(stronger, 5.88) (weaker, 5.88)};
            \end{axis}
        \end{tikzpicture} \\

        \cellcolor{green!10}
        \begin{tikzpicture}
            \begin{axis}[mybarstyle, title={SP\phantom{$_\text{p}$}}, ylabel={accuracy}, ymin=48, ymax=62]
            \addplot[fill=vlmcolor, draw=black] table[x=subset, y=val-knn] {\customclipcomparison};
            \end{axis}
        \end{tikzpicture} &

        \cellcolor{orange!10}
        \begin{tikzpicture}
            \begin{axis}[mybarstyle, title={SPD$_\text{p}$ - JPEG}, ylabel={$\Delta_p$}, ymin=21]
            \addplot[fill=vlmcolor, draw=black] table[x=subset, y=jpeg-sem] {\customclipcomparison};
            \end{axis}
        \end{tikzpicture} &

        \cellcolor{orange!10}
        \begin{tikzpicture}
            \begin{axis}[mybarstyle, title={SPD$_\text{p}$ - Resizing}, ymin=24.7, ytick={25, 26}]
            \addplot[fill=vlmcolor, draw=black] table[x=subset, y=resizing-sem] {\customclipcomparison};
            \end{axis}
        \end{tikzpicture}  &

        \cellcolor{orange!10}
        \begin{tikzpicture}
            \begin{axis}[mybarstyle, title={SPD$_\text{a}$ - Camera}, ylabel={$\Delta_a$}, ymin=5]
            \addplot[fill=vlmcolor, draw=black] table[x=subset, y=spd-a] {\customclipcomparison};
            \end{axis}
        \end{tikzpicture} \\

    \end{tabular}
    \caption{
    \textbf{Impact of correlations between semantics and acquisition metadata on the model's sensitivity to metadata.} ResNet50 trained from scratch using CLIP loss~\cite{radford2021learning} on three Re-LAION-2B subsets with Cram\'er's V correlations of 0.396, 0.255, and 0.166, referred to as stronger, baseline, and weaker, respectively.
    }
    \label{fig:clip_training}
\end{figure}

\noindent\textbf{Impact of correlations.}
We train
ResNet50 models\footnote{ViT-S/16~\cite{dosovitskiy2020image} and ConvNeXt-T~\cite{liu2022convnet} results are presented in the supplementary.} using CLIP loss~\cite{radford2021learning}, with results
in \cref{fig:clip_training}. Training on the subset with stronger correlations consistently leads to higher SPD, while training on the weaker one tends to yield lower SPD, indicating that acquisition metadata correlations are also exploited as shortcuts. Furthermore, consistently with \cref{sec:processing_correlations}, models trained on the weaker subset exhibit the strongest semantic prediction ability, and those trained on the stronger subset the weakest. The MP results are less conclusive; nevertheless,
interference with semantic predictions reflected by SPD is the key signal of metadata sensitivity.

\section{Analysis of mitigation approaches}
\label{sec:mitigation}
We investigate mitigation in two settings: during training on a standard benchmark (IN1k), and as post-processing of released, widely used foundational models that already exhibit metadata sensitivity. Unlike \cref{sec:causes}, these experiments do not rely on artificially introduced correlations, but instead target sensitivity reduction under naturally occurring ones.

\subsection{Mitigation during training with augmentations}
\label{sec:mitigation_train}
\noindent\textbf{Augmentations during training.} Ramos \etal~\cite{ramos2025traces} argue that strong data augmentations during training reduce metadata sensitivity.
We explore this by training and testing ResNet50 on original IN1k.
The default training recipe includes RandomAugment (RA)~\cite{cubuk20randaugment},
while we remove some or add additional
augmentations to test their impact.

\begin{figure}[t]
\vspace{-11pt}
    \centering
    \input{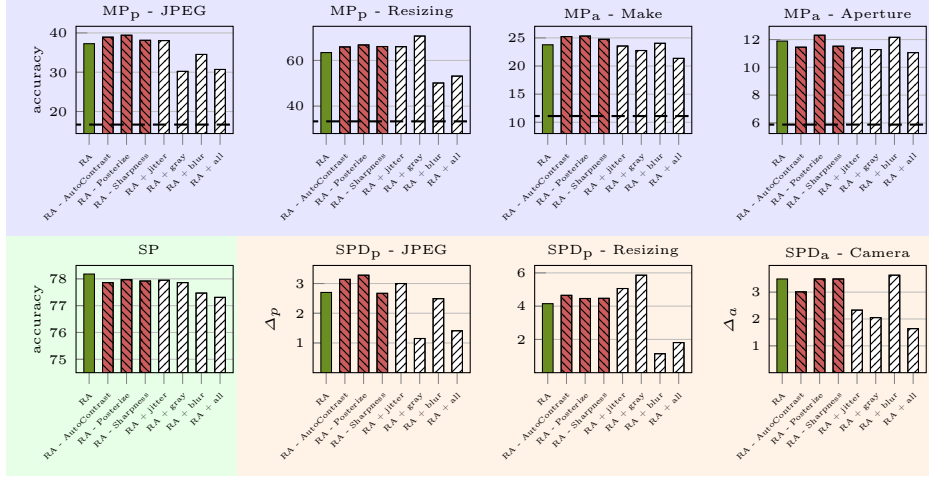}

\pgfplotsset{
    mybarstyle3/.style={
        ybar=1pt, 
        bar width=4pt,
        width=0.3\linewidth,
        height=3cm,
        enlarge x limits=0.1,
        xtick=data,
        xticklabels from table={\resnetraaugmentations}{aug},
        x tick label style={font=\tiny, rotate=45, anchor=north east, scale=0.6},
        title style={font=\tiny, yshift=-7pt},
        label style={font=\tiny},
        tick label style={font=\tiny},
        y label style={yshift=-4pt},
        x label style={yshift=5pt},
        ymajorgrids=true,
        xtick pos=bottom,
        ytick pos=left,
    }
}

\pgfplotsset{
    mybarstyle4/.style={
        ybar=1pt, 
        bar width=4pt,
        width=0.3\linewidth,
        height=3cm,
        enlarge x limits=0.1,
        xtick=data,
        xticklabels from table={\resnetraaugmentations}{aug},
        x tick label style={font=\tiny, rotate=45, anchor=north east, scale=0.6},
        title style={font=\tiny, yshift=-7pt},
        label style={font=\tiny},
        tick label style={font=\tiny},
        y label style={yshift=-4pt},
        x label style={yshift=5pt},
        ymajorgrids=true,
        xtick pos=bottom,
        ytick pos=left,
    }
}

\newcommand{\addhatchplot}[2]{
    \addplot[draw=none, fill=none, forget plot] table[x expr=\coordindex, y=#1] {#2};
    
    \addplot[bar shift=0pt, fill=supervisedcolor, draw=black, restrict expr to domain={\thisrow{type}}{0:0}] table[x expr=\coordindex, y=#1] {#2};
    \addplot[bar shift=0pt, fill=supervisedcolor, postaction={pattern=north west lines}, draw=black, restrict expr to domain={\thisrow{type}}{1:1}] table[x expr=\coordindex, y=#1] {#2};
    \addplot[bar shift=0pt, fill=supervisedcolor, postaction={pattern=north east lines}, draw=black, restrict expr to domain={\thisrow{type}}{2:2}] table[x expr=\coordindex, y=#1] {#2};
}

\newcommand{\addhatchplotalt}[2]{
    \addplot[draw=none, fill=none, forget plot] table[x expr=\coordindex, y=#1] {#2};
    
    \addplot[bar shift=0pt, fill=supervisedcolor, draw=black, restrict expr to domain={\thisrow{type}}{0:0}] table[x expr=\coordindex, y=#1] {#2};
    \addplot[bar shift=0pt, fill=raminuscolor, postaction={pattern=north west lines}, draw=black, restrict expr to domain={\thisrow{type}}{1:1}] table[x expr=\coordindex, y=#1] {#2};
    \addplot[bar shift=0pt, fill=rapluscolor, postaction={pattern=north east lines}, draw=black, restrict expr to domain={\thisrow{type}}{2:2}] table[x expr=\coordindex, y=#1] {#2};
}

\centering
    \begin{tabular}{cccc}

        \cellcolor{blue!10}
        \begin{tikzpicture}
            \begin{axis}[mybarstyle3, title={MP$_\text{p}$ - JPEG}, ylabel={accuracy}]
                \addhatchplotalt{jpeg}{\resnetraaugmentations};
                \addplot[
                black, 
                dashed, 
                thick, 
                sharp plot, %
                dash pattern=on 5pt off 3pt,
                mark=none,  %
                shorten <=-8pt, %
                shorten >=-8pt  %
            ] coordinates {(0, 16.67) (7, 16.67)};
            \end{axis}
        \end{tikzpicture} &

        \cellcolor{blue!10}
        \begin{tikzpicture}
            \begin{axis}[mybarstyle3, title={MP$_\text{p}$ - Resizing}, ymin=28, ylabel={\phantom{$\Delta_p$}}]
                \addhatchplotalt{resizing}
                {\resnetraaugmentations};
                \addplot[
                black, 
                dashed, 
                thick, 
                sharp plot, %
                dash pattern=on 5pt off 3pt,
                mark=none,  %
                shorten <=-8pt, %
                shorten >=-8pt  %
            ] coordinates {(0, 33.33) (7, 33.33)};
            \end{axis}
        \end{tikzpicture} &

        \cellcolor{blue!10}
        \begin{tikzpicture}
            \begin{axis}[mybarstyle3, title={MP$_\text{a}$ - Make}, ymin=8]
                \addhatchplotalt{make}{\resnetraaugmentations};
                \addplot[
                black, 
                dashed, 
                thick, 
                sharp plot, %
                dash pattern=on 5pt off 3pt,
                mark=none,  %
                shorten <=-8pt, %
                shorten >=-8pt  %
            ] coordinates {(0, 11.11) (7, 11.11)};
            \end{axis}
        \end{tikzpicture} &

        \cellcolor{blue!10}
        \begin{tikzpicture}
            \begin{axis}[mybarstyle3, title={MP$_\text{a}$ - Aperture}, ylabel={\phantom{$\Delta_a$}}]
                \addhatchplotalt{aperture}{\resnetraaugmentations};
                \addplot[
                black, 
                dashed, 
                thick, 
                sharp plot, %
                dash pattern=on 5pt off 3pt,
                mark=none,  %
                shorten <=-8pt, %
                shorten >=-8pt  %
            ] coordinates {(0, 5.88) (7, 5.88)};
            \end{axis}
        \end{tikzpicture} \\

        \cellcolor{green!10}
        \begin{tikzpicture}
            \begin{axis}[mybarstyle3, title={SP\phantom{$_\text{p}$}}, ylabel={accuracy}, ymin=74.5, ymax=78.5]
                \addhatchplotalt{knn-val}{\resnetraaugmentations};
            \end{axis}
        \end{tikzpicture} &

        \cellcolor{orange!10}
         \begin{tikzpicture}
            \begin{axis}[mybarstyle3, title={SPD$_\text{p}$ - JPEG}, ylabel={$\Delta_p$}, ymin=0, ytick={1,2,3}, yticklabels={\phantom{0}1,2,3}]
                \addhatchplotalt{jpeg}{\resnetraaugmentationssemantics}
            \end{axis}
        \end{tikzpicture} &

        \cellcolor{orange!10}
        \begin{tikzpicture}
            \begin{axis}[mybarstyle3, title={SPD$_\text{p}$ - Resizing}, ymin=0, ytick={2,4,6}, yticklabels={\phantom{0}2,4,6}]
                \addhatchplotalt{resizing}{\resnetraaugmentationssemantics}
            \end{axis}
        \end{tikzpicture} &

        \cellcolor{orange!10}
        \begin{tikzpicture}
            \begin{axis}[mybarstyle3, title={SPD$_\text{a}$ - Camera}, ylabel={$\Delta_a$}, ymin=0, ytick={1,2,3}, yticklabels={\phantom{0}1,2,3}]
                \addhatchplotalt{spd-a}{\resnetraaugmentationssemantics};
            \end{axis}
        \end{tikzpicture} \\

    \end{tabular}
\vspace{-10pt}    
    \caption{\textbf{Mitigation during training with augmentations -- impact on model sensitivity to metadata}. ResNet50 trained on IN1k with a varying set of augmentations. 
    The default case is Random Augment (RA)~\cite{cubuk20randaugment}, while we remove or add augmentations from/to the default.
    \emph{all}: color jitter, grayscale, and blur. 
\vspace{-20pt}
    }
    \label{fig:resnet_augmentations}
\end{figure}

\noindent\textbf{Impact of augmentations.} 
\cref{fig:resnet_augmentations} presents the results of this experiment. Removing any of the RA augmentations commonly increases the model's ability to capture metadata and its sensitivity. Adding Gaussian blur noticeably reduces sensitivity to processing metadata, but has little impact on acquisition metadata; blurring is thus relatively effective at removing the pixel-level signal that metadata leaves, while harming semantic prediction (SP) only slightly. Most notably, the combination of color jittering, blurring, and grayscale conversion consistently reduces sensitivity by a large factor for both metadata types. This is the set of augmentations used by DINOv2~\cite{oquab2024dinov2}, which is reported to be among the least sensitive foundational models~\cite{ramos2025traces}.

\subsection{Post-hoc adversarial mitigation }
\label{sec:mitigation_posthoc}

\noindent\textbf{Adversarial mitigation.}
Given a pretrained model that maps an image to a $D$-dimensional representation space, we train a linear layer $f: \mathbb{R}^{D} \rightarrow \mathbb{R}^{D}$, as a mitigation component, to transform the representation space and reduce metadata sensitivity.
A linear classifier $h: \mathbb{R}^{D} \rightarrow \mathbb{R}^{M}$, applied to the transformed space, predicts among $M$ metadata labels.
Training is adversarial~\cite{alvi2018turning}: $h$ tries to predict metadata labels, while $f$ tries to make the predictions of $h$ as bad as possible.
It proceeds in two steps. (i) \emph{Train the metadata prediction model $h$ while keeping the mitigation component frozen}, via cross-entropy loss and supervision from metadata labels. (ii) \emph{Train the mitigation component $f$ while keeping the metadata prediction model frozen}, using three loss terms: one that keeps the transformed feature $f(x)$ close to the pretrained feature by maximizing their dot product $f(x)^{\top}x$; one that pushes apart features $f(x)$ and $f(y)$ of images sharing the same metadata label by minimizing their dot product $f(x)^{\top}f(y)$; and one that maximizes the entropy of the metadata prediction $h(f(x))$ for each feature $x$.
These two steps are repeated $L$ times.
The process requires access to metadata labels, which we obtain by reprocessing IN1k with different JPEG compression parameters;  can be used with other attributes similarly.

\begin{figure}[t]
\vspace{-2pt}
    \centering
    \input{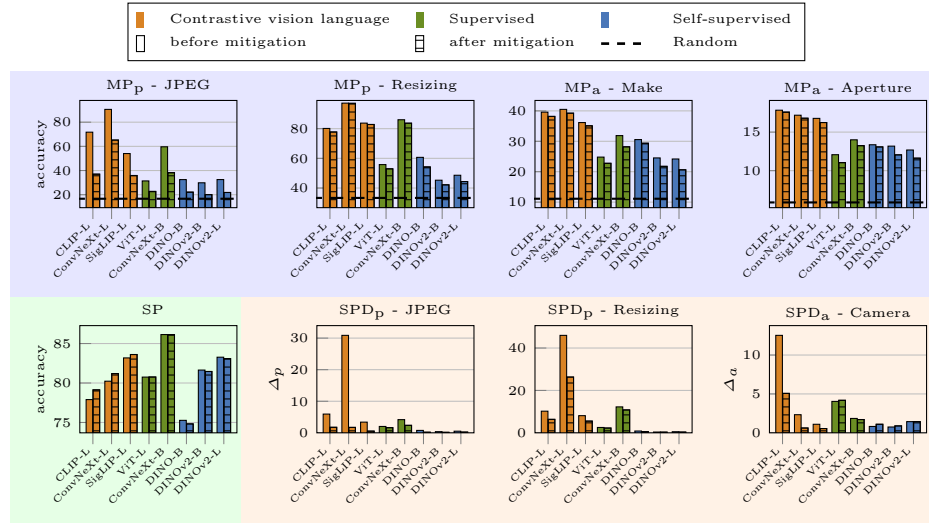}
\pgfplotsset{
    my 4bar style/.style={
        ybar,
        width=0.3\linewidth,
        bar width=2.5pt,
        bar shift=-1.3pt,
        height=3cm,
        enlarge x limits=0.1,
        xtick={0,1,2,3,4,5,6,7}, %
        label style={font=\tiny},
        tick label style={font=\tiny},
        xticklabels = {CLIP-L, ConvNeXt-L, SigLIP-L, ViT-L, ConvNeXt-B, DINO-B, DINOv2-B, DINOv2-L},
        x tick label style={rotate=45, anchor=east, font=\tiny, scale=0.75},
        title style={font=\tiny, yshift=-7pt},
        y label style={yshift=-4pt},
        x label style={yshift=5pt},
        ymajorgrids=true,
        xtick pos=bottom,
        ytick pos=left,
    }
}

\newcommand{\addplotpair}[1]{
    \addplot[fill=vlmcolor, draw=black, restrict x to domain=0:2] table[x=idx, y={#1 orig}] {\debiasprocessinginversion};
    \addplot[preaction={fill=vlmcolor}, draw=black, pattern=horizontal lines, bar shift=1.3pt, restrict x to domain=0:2] table[x=idx, y={#1 deb}] {\debiasprocessinginversion};
    
    \addplot[fill=supervisedcolor, draw=black, restrict x to domain=3:4] table[x=idx, y={#1 orig}] {\debiasprocessinginversion};
    \addplot[preaction={fill=supervisedcolor}, draw=black, pattern=horizontal lines, bar shift=1.3pt, restrict x to domain=3:4] table[x=idx, y={#1 deb}] {\debiasprocessinginversion};
    
    \addplot[fill=sslcolor, draw=black, restrict x to domain=5:7] table[x=idx, y={#1 orig}] {\debiasprocessinginversion};
    \addplot[preaction={fill=sslcolor}, draw=black, pattern=horizontal lines, bar shift=1.3pt, restrict x to domain=5:7] table[x=idx, y={#1 deb}] {\debiasprocessinginversion};
}

\newcommand{\addplotpairsem}[1]{
    \addplot[fill=vlmcolor, draw=black, restrict x to domain=0:2] table[x=idx, y={#1 orig}] {\debiasprocessingsemantics};
    \addplot[preaction={fill=vlmcolor}, draw=black, pattern=horizontal lines, bar shift=1.3pt, restrict x to domain=0:2] table[x=idx, y={#1 deb}] {\debiasprocessingsemantics};
    
    \addplot[fill=supervisedcolor, draw=black, restrict x to domain=3:4] table[x=idx, y={#1 orig}] {\debiasprocessingsemantics};
    \addplot[preaction={fill=supervisedcolor}, draw=black, pattern=horizontal lines, bar shift=1.3pt, restrict x to domain=3:4] table[x=idx, y={#1 deb}] {\debiasprocessingsemantics};
    
    \addplot[fill=sslcolor, draw=black, restrict x to domain=5:7] table[x=idx, y={#1 orig}] {\debiasprocessingsemantics};
    \addplot[preaction={fill=sslcolor}, draw=black, pattern=horizontal lines, bar shift=1.3pt, restrict x to domain=5:7] table[x=idx, y={#1 deb}] {\debiasprocessingsemantics};
}

\pgfplotsset{
    ybar legend/.style={
        legend image code/.code={
            \fill[#1] (0pt,-3pt) rectangle (3pt,3pt);
        }
    }
}

\begin{tikzpicture}
    \begin{axis}[
        hide axis,
        scale only axis,
        height=0cm,
        width=0cm,
        legend style={
            draw=black,
            cells={anchor=west},
            legend columns=3,
            column sep=0.8em,
            font=\tiny,
            anchor=north west
        },
        legend entries={
            {Contrastive vision language},
            {Supervised},
            {Self-supervised},
            {before mitigation},
            {after mitigation},
            {Random}
        }
    ]
        \addlegendimage{ybar, fill=vlmcolor, draw=none, ybar legend}
        \addlegendimage{ybar, fill=supervisedcolor, draw=none, ybar legend}
        \addlegendimage{ybar, fill=sslcolor, draw=none, ybar legend}
        \addlegendimage{ybar, fill=white, draw=black, ybar legend}
        \addlegendimage{ybar, fill=white, draw=black, pattern=horizontal lines, ybar legend}
        \addlegendimage{color=black, line width=1, dashed}
        
        \addplot[draw=none] coordinates {(0,0)};
    \end{axis}
\end{tikzpicture}
\vspace{3pt}

\centering
    \begin{tabular}{cccc}

        \cellcolor{blue!10}
        \begin{tikzpicture}
            \begin{axis}[my 4bar style, title={MP$_\text{p}$ - JPEG}, ylabel={accuracy}]
                \addplotpair{jpeg};
                \addplot[
                black, 
                dashed, 
                thick, 
                sharp plot, %
                dash pattern=on 5pt off 3pt,
                mark=none,  %
                shorten <=-8pt, %
                shorten >=-8pt  %
            ] coordinates {(0, 16.67) (7, 16.67)};
            \end{axis}
        \end{tikzpicture} &

        \cellcolor{blue!10}
        \begin{tikzpicture}
            \begin{axis}[my 4bar style, title={MP$_\text{p}$ - Resizing}, ymax=99, ylabel={\phantom{$\Delta_p$}}]
                \addplotpair{resizing};
                \addplot[
                black, 
                dashed, 
                thick, 
                sharp plot, %
                dash pattern=on 5pt off 3pt,
                mark=none,  %
                shorten <=-8pt, %
                shorten >=-8pt  %
            ] coordinates {(0, 33.33) (7, 33.33)};
            \end{axis}
        \end{tikzpicture} &

        \cellcolor{blue!10}
        \begin{tikzpicture}
            \begin{axis}[my 4bar style, title={MP$_\text{a}$ - Make}]
                \addplotpair{make};
                \addplot[
                black, 
                dashed, 
                thick, 
                sharp plot, %
                dash pattern=on 5pt off 3pt,
                mark=none,  %
                shorten <=-8pt, %
                shorten >=-8pt  %
            ] coordinates {(0, 11.11) (7, 11.11)};
            \end{axis}
        \end{tikzpicture} &

        \cellcolor{blue!10}
        \begin{tikzpicture}
            \begin{axis}[my 4bar style, title={MP$_\text{a}$ - Aperture}, ymin=5.2, ylabel={\phantom{$\Delta_a$}}]
                \addplotpair{aperture};
                \addplot[
                black, 
                dashed, 
                thick, 
                sharp plot, %
                dash pattern=on 5pt off 3pt,
                mark=none,  %
                shorten <=-8pt, %
                shorten >=-8pt  %
            ] coordinates {(0, 5.88) (7, 5.88)};
            \end{axis}
        \end{tikzpicture} \\

        \cellcolor{green!10}
        \begin{tikzpicture}
            \begin{axis}[my 4bar style, title={SP\phantom{$_\text{p}$}}, ylabel={accuracy}]
                \addplotpair{knn-val};
            \end{axis}
        \end{tikzpicture} &

        \cellcolor{orange!10}
        \begin{tikzpicture}
            \begin{axis}[my 4bar style, title={SPD$_\text{p}$ - JPEG}, ylabel={$\Delta_p$}, ymin=0]
                \addplotpairsem{jpeg sem}
            \end{axis}
        \end{tikzpicture} &

        \cellcolor{orange!10}
         \begin{tikzpicture}
            \begin{axis}[my 4bar style, title={SPD$_\text{p}$ - Resizing}, ymin=0]
                \addplotpairsem{resizing sem}
            \end{axis}
        \end{tikzpicture} &

        \cellcolor{orange!10}
        \begin{tikzpicture}
            \begin{axis}[my 4bar style, title={SPD$_\text{a}$ - Camera}, ylabel={$\Delta_a$}, ymin=0]
            \addplotpair{spd-a};
            \end{axis}
        \end{tikzpicture} \\

    \end{tabular}
\vspace{-3pt}
    \caption{\textbf{Post-hoc adversarial mitigation with a linear layer -- impact on model sensitivity to metadata}. 
    Performance is reported for the original frozen foundational models and for the models adjusted by an additional linear layer trained to mitigate the metadata sensitivity guided by manipulated JPEG labels on IN1k. 
\vspace{-5pt}
    }
    \label{fig:debias}
\end{figure}

\noindent\textbf{Impact of adversarial mitigation.} \cref{fig:debias} presents the results for this mitigation strategy  
for different contrastive vision language (CLIP~\cite{radford2021learning}, ConvNeXt~\cite{liu2022convnet}, SigLIP~\cite{zhai2023sigmoid}), supervised (ViT~\cite{dosovitskiy2020image}, ConvNeXt~\cite{liu2022convnet}), and self-supervised (DINO~\cite{caron2021emerging}, DINOv2~\cite{oquab2024dinov2}) models.
We observe that such process consistently reduces the sensitivity for a variety of models, both in terms of metadata prediction and also in terms of semantic prediction distraction. 
The most interesting fact is that the mitigation training is guided by the JPEG compression parameters, but it also reduces the sensitivity to other types of metadata.
Regarding the overall semantic prediction ability, it is either maintained or even slightly improved.

\section{Further analysis}
\label{sec:analysis}

\noindent\textbf{Do Stable Diffusion-generated images carry metadata traces?}
We hypothesize that images generated by models such as Stable Diffusion~\cite{rombach2022stablediffusion} may not contain pixel-level metadata traces, since they do not arise from a physical image formation process and the post-processing pipeline can be kept uniform after generation.
To test this, we train models on synthetically generated images for IN1k classes~\cite{sariyildiz23fakeit,fan24scaling} compressed with fixed high-quality JPEG parameters. 
The resulting models still encode and rely on metadata, with MP$_p$ -- JPEG: 29.91, MP$_p$ -- Resizing: 61.58, SPD$_p$ -- JPEG: 2.00, and SPD$_p$ -- Resizing: 6.06, comparable to training on the original ImageNet (\emph{none}) in \cref{fig:imagenet_image_prob}. 
This suggests generative models may implicitly reproduce pixel-level traces correlated with the semantics in their training data.

\noindent\textbf{Metadata sensitive models are good detectors of generated images.}
Recent work~\cite{ojha2023towards,sha2023fake,smeu2024declip} shows that frozen CLIP features are effective at detecting generated images, which we hypothesize is due to their strong metadata encoding ability. To test this, we apply a $k$-NN classifier ($k=1$) to detect generated images, as in Ojha \etal~\cite{ojha2023towards}, using features from the models in \cref{sec:processing_correlations} with varying metadata sensitivity. \cref{tab:generated_detection} shows that more sensitive models, trained on strong JPEG-semantics correlations, \ie $p_i=50$ or $100$, detect generated images noticeably better than the one trained on the original IN1k. This validates our hypothesis: metadata-semantics correlations encourage the model to focus on low-level details, a  beneficial trait for synthetic image detection.

\begin{table}[t]
    \centering
    \resizebox{\linewidth}{!}{

\begin{tabular}{c cccccc c cc cc c ccc ccc c c}
\toprule
\multirow{2}{*}{Model}
  & \multicolumn{6}{c}{Generative Adversarial Networks}
  & \multirow{2}{*}{\makecell{Deep\\fakes}}
  & \multicolumn{2}{c}{Low level vision}
  & \multicolumn{2}{c}{Perceptual loss}
  & \multirow{2}{*}{Guided}
  & \multicolumn{3}{c}{LDM}
  & \multicolumn{3}{c}{Glide}
  & \multirow{2}{*}{DALL-E}
  & \multirow{2}{*}{Avg} \\
\cmidrule(lr){2-7} \cmidrule(lr){9-10} \cmidrule(lr){11-12} \cmidrule(lr){14-16} \cmidrule(lr){17-19}
  & \makecell{Pro-\\GAN} & \makecell{Cycle-\\GAN} & \makecell{Big-\\GAN}
  & \makecell{Style-\\GAN} & \makecell{Gau-\\GAN} & \makecell{Star-\\GAN}
  &
  & SITD & SAN
  & CRN & IMLE
  &
  & \makecell{200\\steps} & \makecell{200\\w/ CFG} & \makecell{100\\steps}
  & \makecell{100\\27} & \makecell{50\\27} & \makecell{100\\10}
  &
  & \\
\midrule
$p_i = \text{none}$ & 69.1 & 65.3 & \textbf{49.5} & 53.5 & \textbf{58.7} & 55.4 & 51.1 & \underline{55.0} & 49.8 & 57.3 & 59.6 & \underline{51.2} & \underline{62.2} & 53.4 & \textbf{62.6} & 58.3 & \underline{60.7} & \underline{58.7} & \underline{57.4} & 57.3 \\
$p_i = 50\%$ & \underline{71.4} & \underline{66.4} & \underline{49.0} & \textbf{58.6} & \underline{58.1} & \underline{60.1} & \underline{56.1} & 54.7 & \underline{52.3} & \textbf{64.7} & \textbf{68.2} & 49.9 & 61.3 & \underline{53.7} & \underline{61.3} & \underline{59.5} & 60.4 & 57.9 & \underline{57.4} & \underline{59.0} \\
$p_i = 100\%$ & \textbf{77.2} & \textbf{67.0} & 48.4 & \underline{58.4} & 51.4 & \textbf{76.0} & \textbf{58.7} & \textbf{58.3} & \textbf{53.9} & \underline{61.3} & \underline{61.9} & \textbf{55.0} & \textbf{62.9} & \textbf{56.2} & 60.3 & \textbf{60.2} & \textbf{62.5} & \textbf{61.1} & \textbf{57.5} & \textbf{60.4} \\
\bottomrule
\end{tabular}

}

    \caption{\textbf{Metadata sensitive models are good detectors of generated images.} Used models are ResNet50 trained on different versions of IN1k with manipulated JPEG labels from \cref{sec:processing_correlations} (original IN1k/\emph{none}, $p_i=50$ and $p_i=100$ with low, strong and maximal correlations, respectively). Binary classification accuracy reported.}
    \label{tab:generated_detection}
\end{table}

\begin{figure}[t]
\vspace{-15pt}
    \centering
    \input{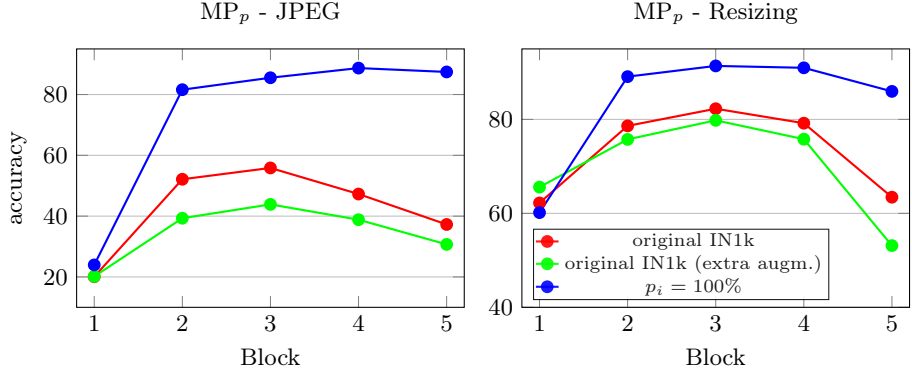}
\centering
\begin{tabular}{cc}
    \begin{tikzpicture}
\begin{axis}[
    title={MP$_p$ - JPEG},
    xlabel={Block},
    ylabel={accuracy},
    xmin=0.8, xmax=5.2,
    ymin=10, ymax=95,
    xtick=data, %
    ymajorgrids=true,
    legend image code/.code={
        \draw[mark repeat=2,mark phase=2,#1]
        plot coordinates {
            (0cm,0cm)
            (0.15cm,0cm) %
            (0.3cm,0cm)  %
        };
    },
    legend style={font=\scriptsize, scale=0.8, opacity=0.8, inner sep=0.5pt, outer sep=-2pt, row sep=-2pt, inner ysep=-0.5pt, at={(0.78, 0.03)}, anchor=south east},
    width=0.55\linewidth,
    height=5cm,
]

\addplot[
    color=red,
    mark=*,
    thick
] table[x=block, y=orig] {\perblock};

\addplot[
    color=green,
    mark=*,
    thick
] table[x=block, y=dino] {\perblock};

\addplot[
    color=blue,
    mark=*,
    thick
] table[x=block, y=bias] {\perblock};

\end{axis}
\end{tikzpicture}
&
\begin{tikzpicture}
\begin{axis}[
    title={MP$_p$ - Resizing},
    xlabel={Block},
    xmin=0.8, xmax=5.2,
    ymin=40, ymax=95,
    xtick=data, %
    legend pos=south west, %
    ymajorgrids=true,
    legend image code/.code={
        \draw[mark repeat=2,mark phase=2,#1]
        plot coordinates {
            (0cm,0cm)
            (0.15cm,0cm) %
            (0.3cm,0cm)  %
        };
    },
    legend style={font=\scriptsize, scale=0.8, opacity=0.9, inner sep=0.5pt, outer sep=0pt, row sep=-2pt, inner ysep=-0.5pt},
    width=0.55\linewidth,
    height=5cm,
]

\addplot[
    color=red,
    mark=*,
    thick
] table[x=block, y=orig] {\perblockresize};
\addlegendentry{original IN1k}

\addplot[
    color=green,
    mark=*,
    thick
] table[x=block, y=dino] {\perblockresize};
\addlegendentry{original IN1k (extra augm.)}

\addplot[
    color=blue,
    mark=*,
    thick
] table[x=block, y=bias] {\perblockresize};
\addlegendentry{$p_i = 100\%$}

\end{axis}
\end{tikzpicture}
\end{tabular}
\vspace{-15pt}    
    \caption{\textbf{Ability of internal model layers to predict metadata}. Metadata prediction (MP) accuracy across blocks of ResNet50 trained on: (i) original IN1k, (ii) original IN1k using extra augmentations (color jitter, grayscale, and blur augmentations), and (iii) modified IN1k with strong JPEG-semantics correlations. In all cases Random Augment is used.}
    \label{fig:blocks}
\end{figure}

\noindent\textbf{Layer-wise sensitivity analysis.}
In \cref{fig:blocks}, we examine the metadata prediction ability across the 5 blocks of ResNet50 models trained in different ways.
All models reach a strong metadata prediction ability already after two blocks, since metadata traces are low-level signals easily captured by early layers~\cite{zeiler2014visualizing}.
The ability further improves after the third block for all models, but then diverges: it starts dropping in deeper layers for the model trained on the original IN1k, while continuing to improve for the model trained on the modified IN1k with strong correlations.
This shows that the former learns to ignore metadata in deeper layers, delegating semantic feature extraction to them, whereas the latter learns to rely on metadata and cannot ignore it, suggesting that semantic discrimination is compromised in favor of metadata reliance.
The use of mitigation effective augmentations, as identified in \cref{sec:mitigation_train}, reduces the MP accuracy.  
Interestingly, the model trained on strong JPEG-semantics correlations behaves similarly for resizing: it captures the resizing metadata in early layers and cannot ignore it in deeper ones, even though it was not trained with strong resizing-semantics correlations.
This is further evidence that the model learns to rely on low-level traces in general, not only on the specific attribute strongly correlated with semantics during training.

\begin{figure}[t]
    \vspace{-5pt}
    \centering
    \setlength{\tabcolsep}{-0pt}

\centering
\begin{tabular}{cccc}
    & $p_i =$ none & $p_i = 80\%$ & $p_i = 100\%$ \\
    \rotatebox[origin=c]{90}{before mitigation} & 
    \includegraphics[width=0.3\linewidth, valign=m]{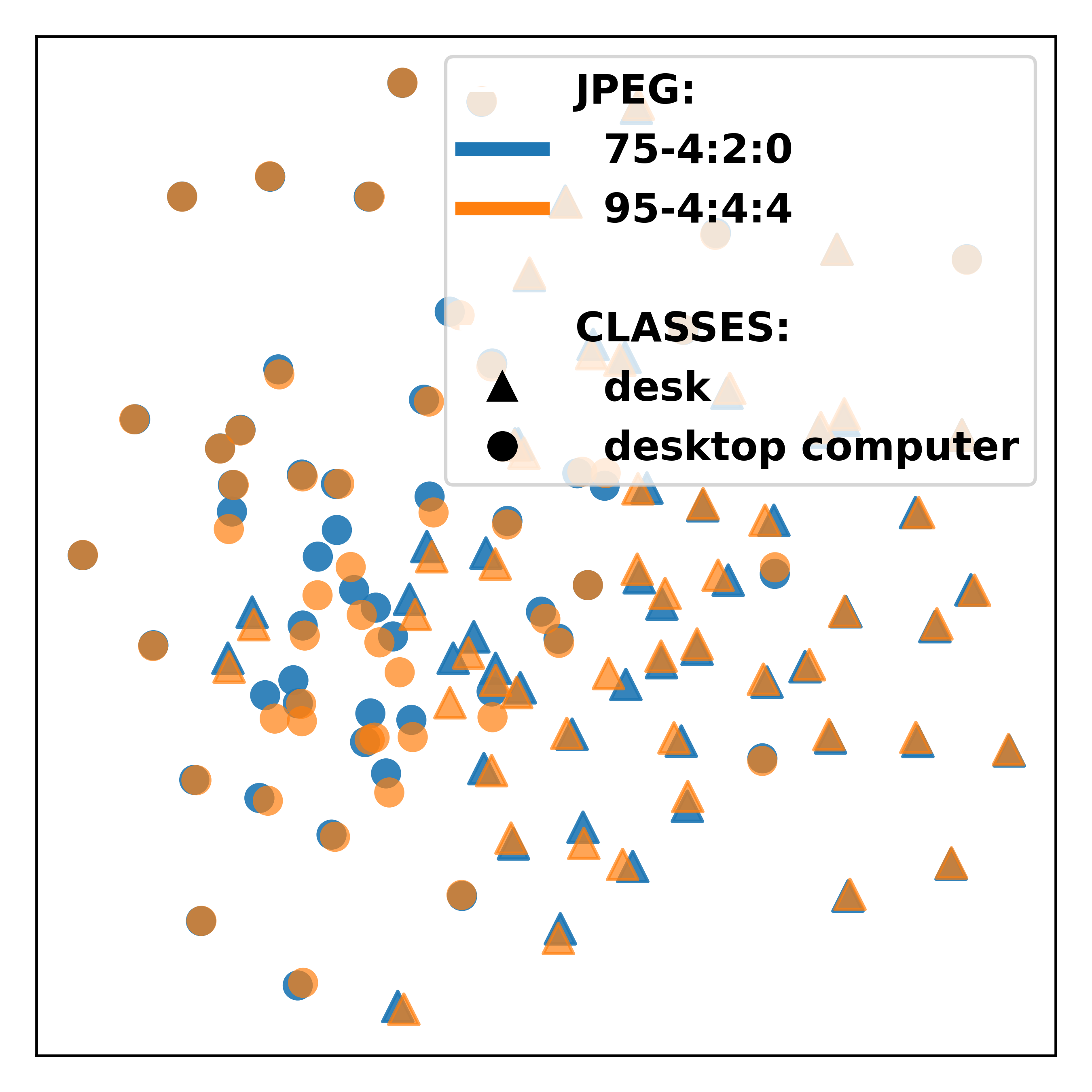} &
    \includegraphics[width=0.3\linewidth, valign=m]{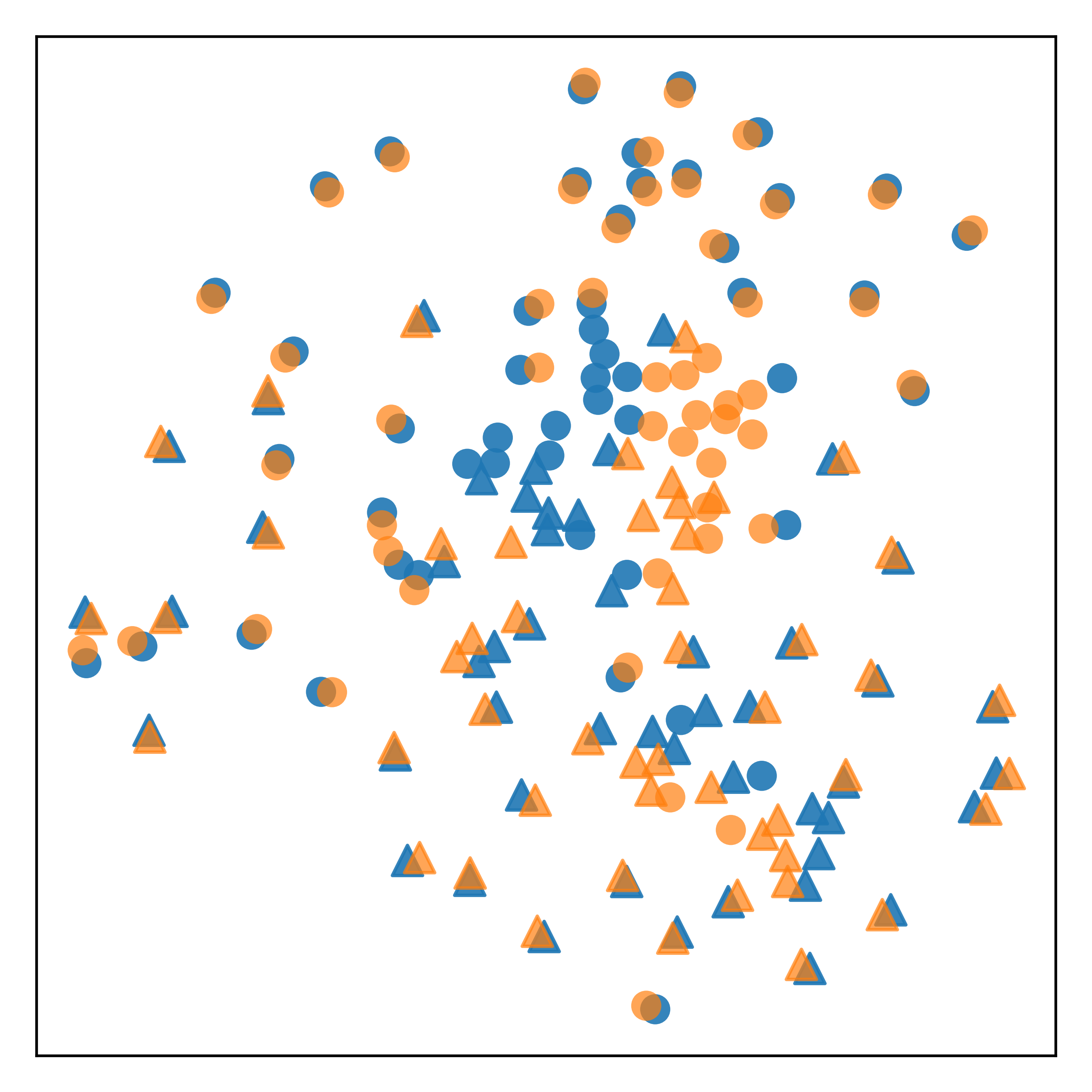} &
    \includegraphics[width=0.3\linewidth, valign=m]{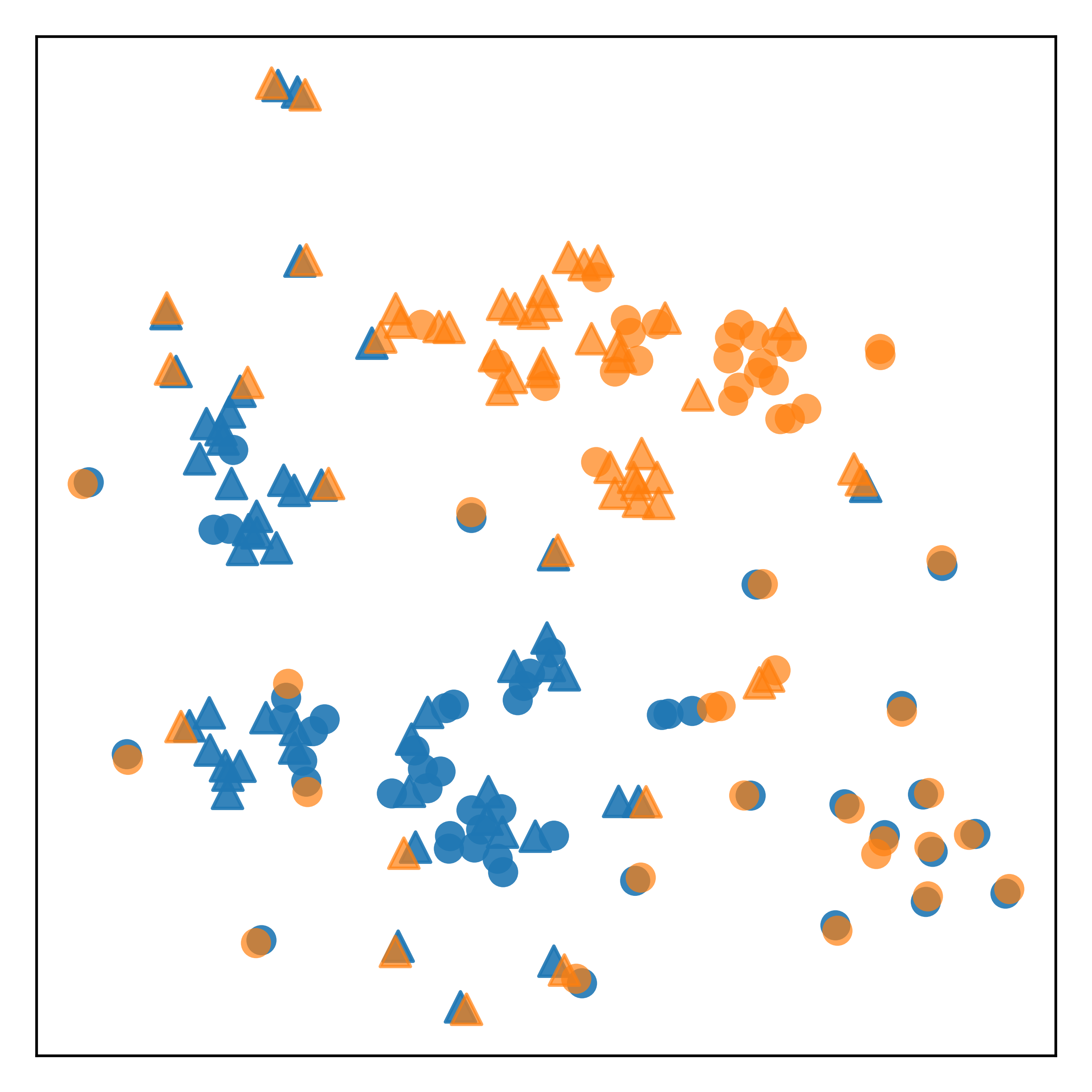}
    \\
    \rotatebox[origin=c]{90}{after mitigation} & 
    \includegraphics[width=0.3\linewidth, valign=m]{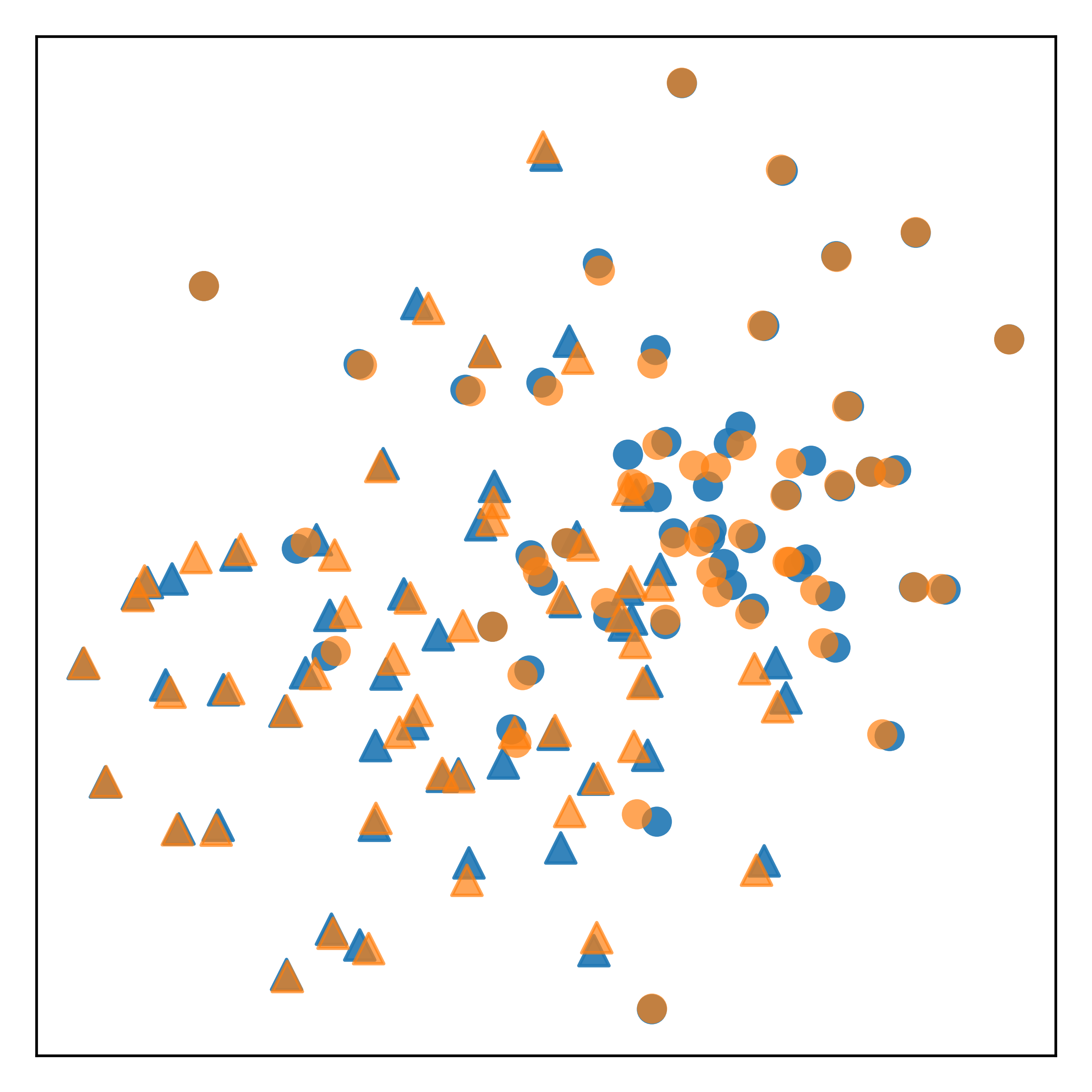} &
    \includegraphics[width=0.3\linewidth, valign=m]{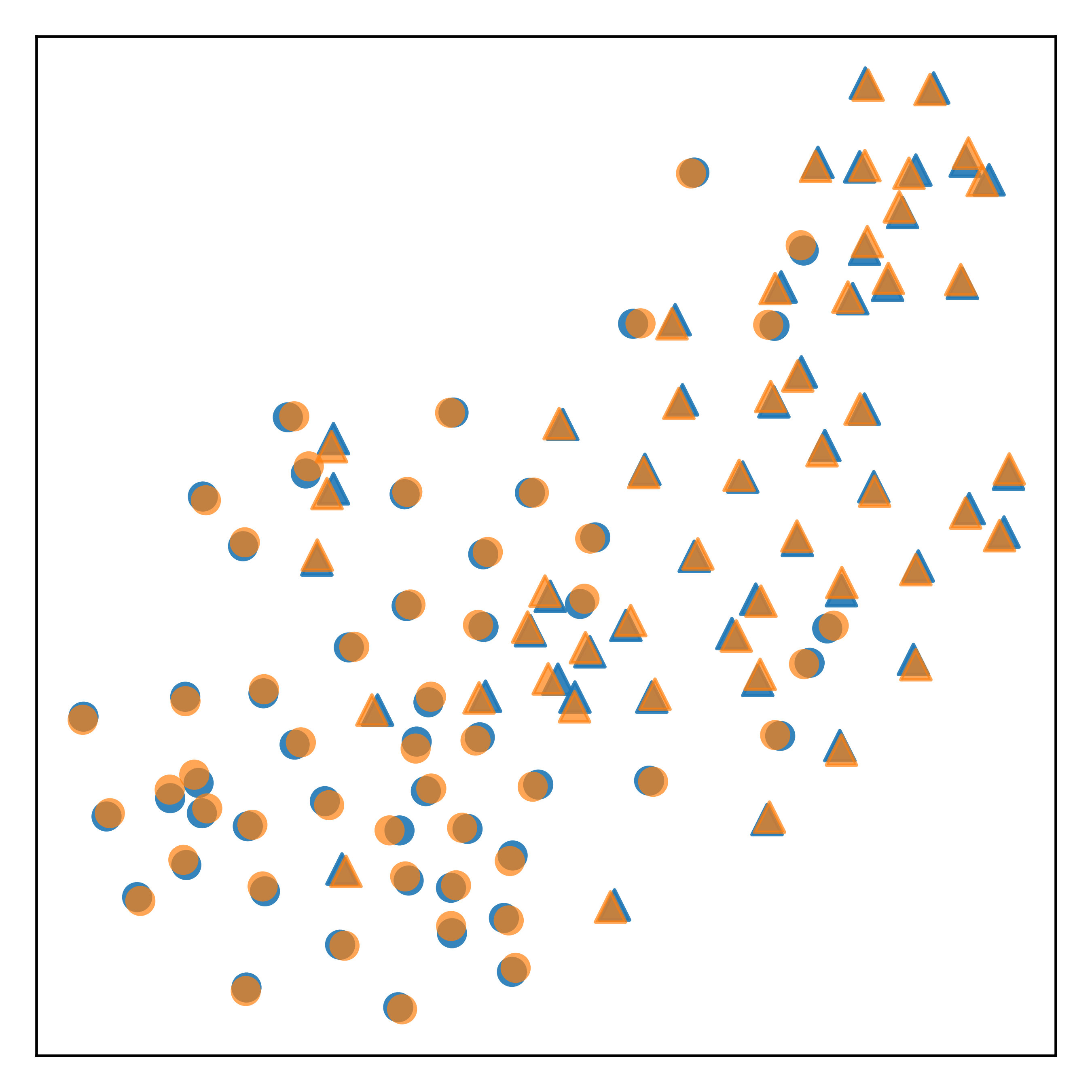} &
    \includegraphics[width=0.3\linewidth, valign=m]{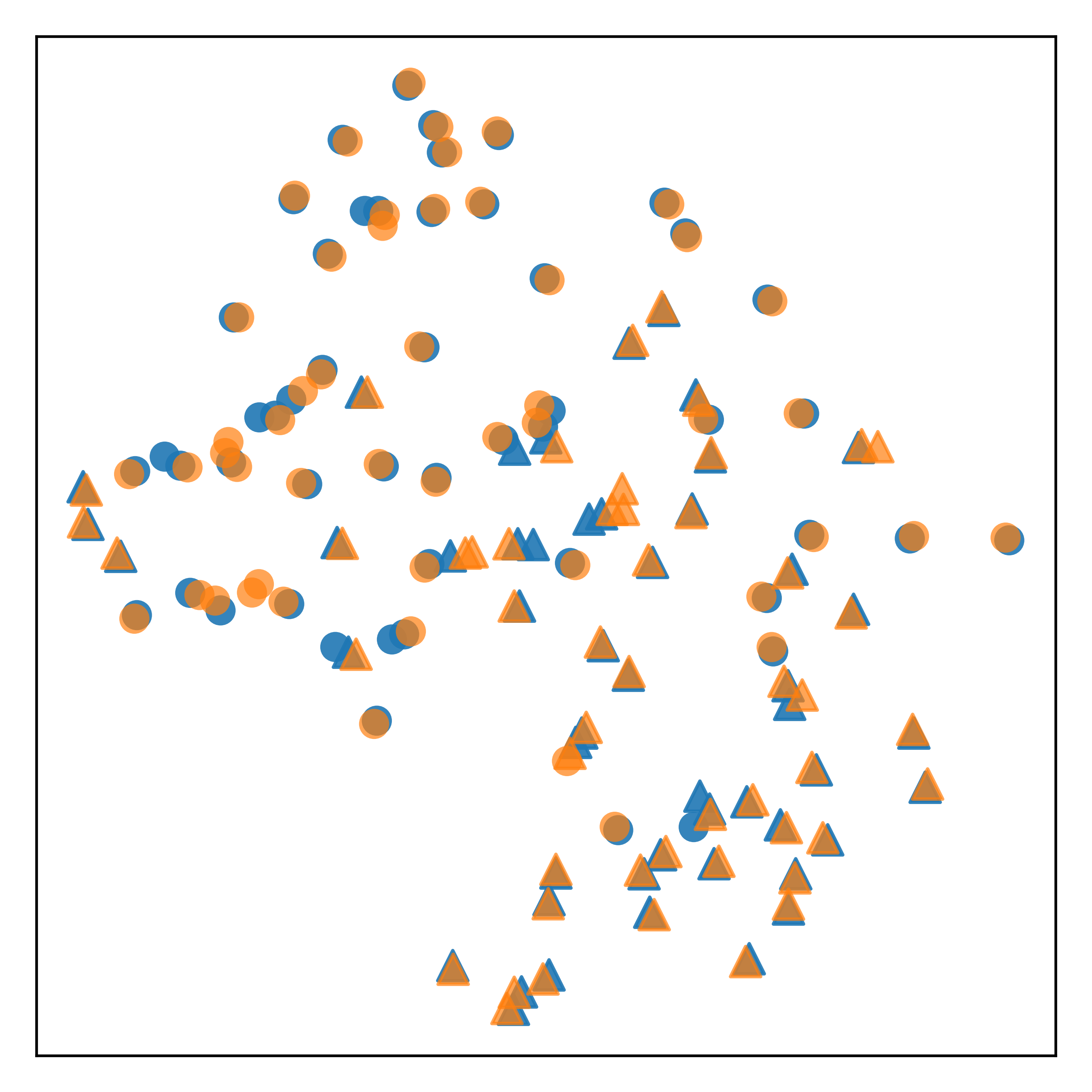}
\end{tabular}

    \vspace{-7pt}
    \caption{\textbf{t-SNE visualization of features before (top) and after (bottom) post-hoc mitigation} for two classes ($\bullet$, $\blacktriangle$) from IN1k  processed by different JPEG parameters (\textcolor[HTML]{1F77B4}{$\blacksquare$}, \textcolor[HTML]{FF7F0E}{$\blacksquare$}). Used models are ResNet50 trained on different versions of IN1k with manipulated JPEG labels from \cref{sec:processing_correlations} (original IN1k/\emph{none}, $p_i=80$ and $p_i=100$ with low, strong and maximal correlations, respectively).
    \vspace{-10pt}
    }
    \label{fig:tsne}
\end{figure}

\noindent\textbf{Representation space visualization before/after post-hoc mitigation.}
\cref{fig:tsne} provides a t-SNE~\cite{van2008visualizing} visualization of ResNet50 features for two IN1k classes, where all images are processed twice with two different sets of JPEG parameters. The models are trained on the original IN1k and on modified versions with strong processing metadata-semantics correlations (\cref{sec:processing_correlations}), shown before and after post-hoc mitigation.
For the model trained on the original IN1k, the classes are decently separated while points with different JPEG parameters overlap well; the small signs of misalignment are consistent with the quantitative measurements in \cref{fig:imagenet_image_prob}, indicating mild sensitivity to metadata. The model thus learns semantic features and is only subtly affected by metadata.
In contrast, the two classes start to mix for models trained under strong correlations, while points with different JPEG parameters become more separated, showing that the model relies on metadata instead of semantics and that metadata leaves strong traces in the representation space.
At $p_i = 100\%$, the classes are almost completely mixed, and the features can instead be separated by JPEG processing.
Finally, post-hoc mitigation improves semantic class separability even for the model trained with $p_i = 100\%$ correlation.

\begin{figure}[t]
    \centering
    \input{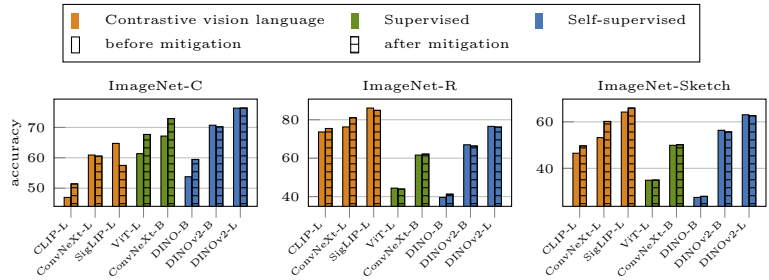}
\pgfplotsset{
    my 4bar style/.style={
        ybar,
        width=0.35\linewidth,
         bar width=2.5pt,
        bar shift=-1.3pt,
        height=3cm,
        enlarge x limits=0.1,
        xtick={0,1,2,3,4,5,6,7}, %
        label style={font=\tiny},
        tick label style={font=\tiny},
        xticklabels = {CLIP-L, ConvNeXt-L, SigLIP-L, ViT-L, ConvNeXt-B, DINO-B, DINOv2-B, DINOv2-L},
        x tick label style={rotate=45, anchor=east, font=\tiny, scale=0.75},
        title style={font=\tiny, yshift=-7pt},
        y label style={yshift=-4pt},
        x label style={yshift=5pt},
        ymajorgrids=true,
        xtick pos=bottom,
        ytick pos=left,
    }
}

\newcommand{\addplotpairsem}[1]{
    \addplot[fill=vlmcolor, draw=black, restrict x to domain=0:2] table[x=idx, y={#1 orig}] {\ood};
    \addplot[preaction={fill=vlmcolor}, draw=black, pattern=horizontal lines, bar shift=1.3pt, restrict x to domain=0:2] table[x=idx, y={#1 deb}] {\ood};
    
    \addplot[fill=supervisedcolor, draw=black, restrict x to domain=3:4] table[x=idx, y={#1 orig}] {\ood};
    \addplot[preaction={fill=supervisedcolor}, draw=black, pattern=horizontal lines, bar shift=1.3pt, restrict x to domain=3:4] table[x=idx, y={#1 deb}] {\ood};
    
    \addplot[fill=sslcolor, draw=black, restrict x to domain=5:7] table[x=idx, y={#1 orig}] {\ood};
    \addplot[preaction={fill=sslcolor}, draw=black, pattern=horizontal lines, bar shift=1.3pt, restrict x to domain=5:7] table[x=idx, y={#1 deb}] {\ood};
}

\pgfplotsset{
    ybar legend/.style={
        legend image code/.code={
            \fill[#1] (0pt,-3pt) rectangle (3pt,3pt);
        }
    }
}

\begin{tikzpicture}
    \begin{axis}[
        hide axis,
        scale only axis,
        height=0cm,
        width=0cm,
        legend style={
            draw=black,
            cells={anchor=west},
            legend columns=3,
            column sep=0.8em,
            font=\tiny,
            anchor=north west
        },
        legend entries={
            {Contrastive vision language},
            {Supervised},
            {Self-supervised},
            {before mitigation},
            {after mitigation}
        }
    ]
        \addlegendimage{ybar, fill=vlmcolor, draw=none, ybar legend}
        \addlegendimage{ybar, fill=supervisedcolor, draw=none, ybar legend}
        \addlegendimage{ybar, fill=sslcolor, draw=none, ybar legend}
        \addlegendimage{ybar, fill=white, draw=black, ybar legend}
        \addlegendimage{ybar, fill=white, draw=black, pattern=horizontal lines, ybar legend}
        
        \addplot[draw=none] coordinates {(0,0)};
    \end{axis}
\end{tikzpicture}
\vspace{3pt}

\centering
    \begin{tabular}{ccc}
        \begin{tikzpicture}
            \begin{axis}[my 4bar style, title={ImageNet-C}, ylabel={accuracy}]
                \addplotpairsem{in-c}
            \end{axis}
        \end{tikzpicture} &

        \begin{tikzpicture}
            \begin{axis}[my 4bar style, title={ImageNet-R}, ylabel={}]
                \addplotpairsem{in-r}
            \end{axis}
        \end{tikzpicture} &

        \begin{tikzpicture}
            \begin{axis}[my 4bar style, title={ImageNet-Sketch}]
                \addplotpairsem{in-sketch}
            \end{axis}
        \end{tikzpicture} \\

    \end{tabular}

    \caption{\textbf{Post-hoc adversarial mitigation with a linear layer -- impact on ImageNet OOD test sets}. 
    Performance is reported for the original frozen foundational models and for the models adjusted by an additional linear layer trained to mitigate the metadata sensitivity guided by manipulated JPEG labels on IN1k. 
    Accuracy reported on ImageNet-C instead of the commonly reported error.}
    \label{fig:ood}
\end{figure}

\noindent\textbf{Impact of mitigation on OOD tasks.}
\label{sec:ood_supp}
Since metadata traces can be a spurious low-level signal that models exploit as a shortcut, reducing reliance on them should be most beneficial under distribution shift, where such shortcuts no longer hold. We therefore investigate whether reducing the models' metadata sensitivity improves performance on common out-of-distribution (OOD) ImageNet test sets: ImageNet-C~\cite{hendrycks2019benchmarking}, ImageNet-R~\cite{hendrycks2021many}, and ImageNet-Sketch~\cite{wang19sketch}. \cref{fig:ood} reports results without and with the post-hoc mitigation. The mitigation helps in many cases, sometimes considerably, and the gains are largest for the models most sensitive to metadata (according to \cref{fig:debias}: CVLs and supervised ConvNeXt), while sometimes hurting less sensitive ones like DINOv2. We observe similar trends for ResNet50 trained on the original IN1k (ImageNet-C: $44.2 \!\rightarrow\! 45.0$, ImageNet-R: $34.5 \!\rightarrow\! 37.7$, ImageNet-Sketch: $26.4 \!\rightarrow\! 30.0$) when mitigating during training with augmentations.

\section{Related work}
\label{sec:related}

\subsection{Shortcut learning}
\vspace{5pt}
In empirical risk minimization (ERM) settings, neural networks have been shown to rely on simple input features as shortcuts to solve the task being learned, even if the shortcuts are only spuriously correlated with features more relevant to the task \cite{geirhos2020shortcut, suhail2025shortcut}. This is because under ERM, the objective is simply to minimize error, meaning that shortcut features will be learned as long as they help with the training objective \cite{nagarajan2021understanding,sagawa2020investigation}. Not only are shortcut features learned, but they are even preferred by models due to their simplicity. Even in the presence of features that are more salient and more informative, neural networks will still rely more strongly on the simple spurious features if the salient features are more complex to learn \cite{shah2020pitfalls}.

A frequent origin of shortcuts is dataset bias. Due to data sources and collection methods \cite{liu2024decade,shirali2025what}, as well as their properties such as confounding factors \cite{zech2018variable}, data scarcity \cite{parashar2024neglected}, collection location \cite{gaviria2022dollar}, and collection timeframe \cite{recht2019imagenet},
resulting datasets may exhibit unique characteristics to the point that it is possible to distinguish them from one another despite having similar design goals \cite{torralba2011unbiased,liu2024decade}. More importantly, among these idiosyncrasies may be correlations that create exploitable shortcuts.
While these commonly manifest as object-background correlations, as is the case with ImageNet \cite{Neuhaus_2023_ICCV, singla2022salient, hendrycks2021natural}, \eg the side of freight cars being associated with graffiti, non-object-centric correlations also exist. In LAION, images of cardboard boxes are commonly accompanied by watermarks \cite{li2023whac}. 
Even spatial characteristics, such as object location, and lower-level features, such as color distributions, have been found to be correlated with different semantic concepts~\cite{Meister_2023_ICCV, Jahanian*2020On}.
Models trained on these datasets in standard ERM settings typically learn these shortcuts and predict the semantic label even when the image contains only the shortcut feature~\cite{Neuhaus_2023_ICCV, singla2022salient, hendrycks2021natural, li2023whac}.

In contrast to the aforementioned immediately perceptible cues, we focus on largely imperceptible shortcuts, mainly JPEG compression artifacts. 
Previous studies that have examined their role in shortcut learning focus on specialized tasks \cite{grommelt2024fake} or exist outside the context of their correlation with semantic labels \cite{Doersch_2015_ICCV},
while our work investigates their presence in large general-purpose training datasets and consequent ability to interfere with foundation models.

Additionally, neural networks have been shown to rely on frequency-based shortcuts~\cite{wang23frequency,wang25imagenetfrequency}, \eg zebra patterns being easily distinguishable in the spectrum, and such shortcuts arise even in models trained at ImageNet scale. Although some types of metadata could also introduce frequency-related shortcuts, these works do not account for them.

\subsection{Shortcut mitigation}
\vspace{5pt}
A diverse range of strategies has been proposed to mitigate shortcut learning.
One prominent direction is \emph{data-centric}, where the training distribution is modified to reduce spurious correlations.
These approaches include data augmentation techniques~\cite{geirhos2018imagenet,yao2022improving,puli2022nuisances}, such as image stylization or mixup, which expand the dataset with semantically consistent transformations.
Other methods rely on resampling or reweighting strategies~\cite{li2019repair,labonte2024group} to balance class distributions, as well as concept discovery and pseudo-label extraction techniques~\cite{wu2023discover,nam2022spread,sarridis2025mavias} that incorporate additional supervisory signals.
A second line of work mitigates shortcuts directly during training.
Some approaches introduce additional classification heads~\cite{kim2019learning,alvi2018turning} or train parallel models~\cite{wang2019learning,sarridis2024flac,sarridis2025mavias} to explicitly capture shortcut-related features, thereby encouraging the main model to ignore them.

Post-training approaches have also been explored, typically in a two-stage framework.
These methods often leverage biased models to identify potential shortcuts~\cite{yenamandra2023facts,yang2024identifying,varma2024ravl}, employ ensembles of bias models~\cite{nam2020learning,kim2022learning,bahng2020learning,lee2023diversify}, or focus on misclassified examples using contrastive learning techniques~\cite{zhang2022correct}.
In addition, optimization-based methods modify the training objective to improve robustness to spurious correlations~\cite{liu2021just,sagawa2020distributionally,liu2022avoiding}.
Other approaches fine-tune biased models either entirely~\cite{taghanaki2022masktune} or only at the final layer~\cite{kirichenko2022last,izmailov2022feature,qiu2023simple,du2021fairness}, often using feature reweighting strategies.
Recent work has also begun exploring bias mitigation in foundation models, particularly in vision--language systems~\cite{yang2023mitigating,varma2024ravl}.
Our mitigation strategy draws inspiration from these directions. Our post-hoc contrastive separation term is similar to prior work~\cite{alvi2018turning,zhang2022correct}, but
unlike prior methods, our approach specifically targets \emph{metadata traces} in vision encoders.

\section{Discussion and conclusions}
\label{sec:conclusion}
We study why vision encoders become sensitive to image metadata and show that
this sensitivity is the result of shortcut learning, emerging when metadata
correlates with the semantic supervision used during pretraining. This accounts
for its presence under categorical supervision on ImageNet and, as our
controlled LAION experiments
indicate,
under caption supervision in
vision-language pretraining. Self-supervised methods such as the DINO family are
less sensitive, since they do not optimize for semantic targets, yet not fully
invariant, as metadata still serves as a shortcut for matching views of the same
image. More broadly, we establish the image formation and processing pipeline
itself as a source of bias, imperceptible to humans yet readily exploited by
models.

Crucially, metadata sensitivity is not purely a disadvantage. The tendency to
capture pixel-level signal that harms robustness is what makes an encoder a
strong detector of generated images: models trained under stronger
metadata-semantics correlations, and thus more sensitive, detect synthetic
images noticeably better. This reframes recent observations that frozen CLIP
features excel at fake-image
detection as a direct
consequence of metadata encoding, making sensitivity double-edged, a
disadvantage for semantic generalization but an asset for forensics. 
In our analysis we identify further connections than detection: even images synthesized by Stable Diffusion, which never pass through a physical acquisition pipeline, seem to carry pixel-level metadata traces correlated with semantics, suggesting generators implicitly imitate the acquisition and processing signatures of their training data.

Since metadata distributions drift across cameras, capture conditions, and
processing pipelines, encoders relying on such shortcuts are fragile precisely
where deployment demands robustness. We show this sensitivity can be mitigated
both during and after pretraining, generalizing across attributes while
preserving semantic utility. Consistently, reducing it improves
out-of-distribution generalization on ImageNet-C, ImageNet-R, and ImageNet-Sketch, with larger gains for the more sensitive models,
suggesting metadata invariance may need to be a first-class objective when
training or adapting foundation encoders.

We believe this work opens a new horizon, shifting attention from visible,
semantic biases to the invisible traces left by the imaging pipeline. It remains
open how strongly metadata sensitivity contaminates downstream tasks and the
benchmarks used to evaluate encoders.

\noindent\textbf{Acknowledgments:}
This work was supported by the GACR grant GA26-24228S, the Horizon MSCA-PF grant No. 101154126, the Czech Technical University in Prague grant No. SGS26/074/OHK3/1T/13, the JSPS KAKENHI No. 23H00497 and No. 22K12091. We acknowledge VSB – Technical University of Ostrava, IT4Innovations National Supercomputing Center, Czech Republic, for awarding this project (OPEN-35-7) access to the LUMI supercomputer, owned by the EuroHPC Joint Undertaking, hosted by CSC (Finland) and the LUMI consortium through the Ministry of Education, Youth and Sports of the Czech Republic through the e-INFRA CZ (grant ID: 90254).

\bibliographystyle{splncs04}
\bibliography{main}

\clearpage

\begin{center}
    {\LARGE \textbf{Supplementary Material}}
\end{center}

\setcounter{section}{0}
\setcounter{figure}{0}
\setcounter{table}{0}
\renewcommand{\thesection}{\Alph{section}}
\renewcommand{\thefigure}{\Roman{figure}}
\renewcommand{\thetable}{\Roman{table}}

\begin{figure}[h]
    \centering
    \input{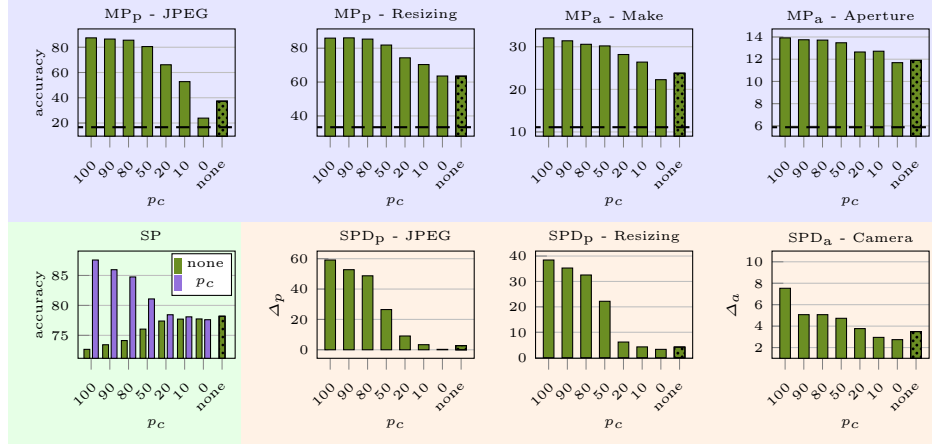}
\pgfplotsset{
    mybarstyle2/.style={
        ybar=1pt, 
        bar width=4pt,
        width=0.3\linewidth,
        height=3cm,
        enlarge x limits=0.1,
        xtick=data,
        xticklabels from table={\resnetimagenetjpegprobclassinversion}{p},
        x tick label style={font=\tiny, rotate=45, anchor=north east},
        title style={font=\tiny, yshift=-7pt},
        label style={font=\tiny},
        tick label style={font=\tiny},
        y label style={yshift=-4pt},
        x label style={yshift=5pt},
        ymajorgrids=true,
        xtick pos=bottom,
        ytick pos=left,
    }
}

\centering
    \begin{tabular}{cccc}

        \cellcolor{blue!10}
        \begin{tikzpicture}
            \begin{axis}[mybarstyle2, title={MP$_\text{p}$ - JPEG}, ylabel={accuracy}, xlabel={$p_c$}]
            \addplot[fill=supervisedcolor, draw=black, bar shift=0pt] table[x expr=\coordindex, y=jpeg] {\resnetimagenetjpegprobclassinversion};
            \addplot[fill=supervisedcolor, draw=black, postaction={pattern=crosshatch dots}, bar shift=0pt, restrict x to domain=6.5:7.5] table[x expr=\coordindex, y=jpeg] {\resnetimagenetjpegprobclassinversion};
            \addplot[
                black, 
                dashed, 
                thick, 
                sharp plot, %
                dash pattern=on 5pt off 3pt,
                mark=none,  %
                shorten <=-8pt, %
                shorten >=-8pt  %
            ] coordinates {(0, 16.67) (7, 16.67)};
            \end{axis}
        \end{tikzpicture} &

        \cellcolor{blue!10}
        \begin{tikzpicture}
            \begin{axis}[mybarstyle2, title={MP$_\text{p}$ - Resizing}, xlabel={$p_c$}, ylabel={\phantom{$\Delta_p$}}]
            \addplot[fill=supervisedcolor, draw=black, bar shift=0pt] table[x expr=\coordindex, y=resizing] {\resnetimagenetjpegprobclassinversion};
            \addplot[fill=supervisedcolor, draw=black, postaction={pattern=crosshatch dots}, bar shift=0pt, restrict x to domain=6.5:7.5] table[x expr=\coordindex, y=resizing] {\resnetimagenetjpegprobclassinversion};
            \addplot[
                black, 
                dashed, 
                thick, 
                sharp plot, %
                dash pattern=on 5pt off 3pt,
                mark=none,  %
                shorten <=-8pt, %
                shorten >=-8pt  %
            ] coordinates {(0, 33.33) (7, 33.33)};
            \end{axis}
        \end{tikzpicture} &

        \cellcolor{blue!10}
        \begin{tikzpicture}
            \begin{axis}[mybarstyle2, title={MP$_\text{a}$ - Make}, xlabel={$p_c$}]
            \addplot[fill=supervisedcolor, draw=black, bar shift=0pt] table[x expr=\coordindex, y=make] {\resnetimagenetjpegprobclassinversion};
            \addplot[fill=supervisedcolor, draw=black, postaction={pattern=crosshatch dots}, bar shift=0pt, restrict x to domain=6.5:7.5] table[x expr=\coordindex, y=make] {\resnetimagenetjpegprobclassinversion};
            \addplot[
                black, 
                dashed, 
                thick, 
                sharp plot, %
                dash pattern=on 5pt off 3pt,
                mark=none,  %
                shorten <=-8pt, %
                shorten >=-8pt  %
            ] coordinates {(0, 11.11) (7, 11.11)};
            \end{axis}
        \end{tikzpicture} &

        \cellcolor{blue!10}
        \begin{tikzpicture}
            \begin{axis}[mybarstyle2, title={MP$_\text{a}$ - Aperture}, xlabel={$p_c$}, ylabel={\phantom{$\Delta_a$}}]
            \addplot[fill=supervisedcolor, draw=black, bar shift=0pt] table[x expr=\coordindex, y=aperture] {\resnetimagenetjpegprobclassinversion};
            \addplot[fill=supervisedcolor, draw=black, postaction={pattern=crosshatch dots}, bar shift=0pt, restrict x to domain=6.5:7.5] table[x expr=\coordindex, y=aperture] {\resnetimagenetjpegprobclassinversion};
            \addplot[
                black, 
                dashed, 
                thick, 
                sharp plot, %
                dash pattern=on 5pt off 3pt,
                mark=none,  %
                shorten <=-8pt, %
                shorten >=-8pt  %
            ] coordinates {(0, 5.88) (7, 5.88)};
            \end{axis}
        \end{tikzpicture} \\

        \cellcolor{green!10}
        \begin{tikzpicture}
            \begin{axis}[mybarstyle2, title={SP\phantom{$_\text{p}$}}, ylabel={accuracy}, xlabel={$p_c$}, bar width=2.2pt, bar shift=0pt, legend style={font=\tiny, scale=0.6, /tikz/every even column/.append style={column sep=0.1cm}, inner sep=0.8pt, row sep=-1pt}, legend image code/.code={
                    \draw[#1, draw=none] (0cm,-0.1cm) rectangle (0.1cm,0.1cm); %
                }]
            \addplot[draw=none, fill=none, bar shift=0pt, forget plot] table[x expr=\coordindex, y=val-knn] {\resnetimagenetjpegprobclassinversion};
            \addplot[fill=supervisedcolor, draw=black, bar shift=-1.6pt, restrict x to domain=-0.5:6.5] table[x expr=\coordindex, y=val-knn] {\resnetimagenetjpegprobclassinversion};\addlegendentry{none};
            \addplot[fill=probcolor, draw=black, bar shift=1.6pt, restrict x to domain=-0.5:6.5] table[x expr=\coordindex, y=val-bias] {\resnetimagenetjpegprobclassinversion};\addlegendentry{$p_c$};
            \addplot[fill=supervisedcolor, draw=black, bar shift=0pt, restrict x to domain=6.5:7.5, forget plot] table[x expr=\coordindex, y=val-knn] {\resnetimagenetjpegprobclassinversion};
            \addplot[fill=supervisedcolor, draw=black, postaction={pattern=crosshatch dots}, bar shift=0pt, restrict x to domain=6.5:7.5, forget plot] table[x expr=\coordindex, y=val-knn] {\resnetimagenetjpegprobclassinversion};
            \end{axis}
        \end{tikzpicture} &

        \cellcolor{orange!10}
        \begin{tikzpicture}
            \begin{axis}[mybarstyle2, title={SPD$_\text{p}$ - JPEG}, ylabel={$\Delta_p$}, xlabel={$p_c$}]
            \addplot[fill=supervisedcolor, draw=black, bar shift=0pt] table[x expr=\coordindex, y=jpeg] {\resnetimagenetjpegprobclasssemantics};
            \addplot[fill=supervisedcolor, draw=black, postaction={pattern=crosshatch dots}, bar shift=0pt, restrict x to domain=6.5:7.5] table[x expr=\coordindex, y=jpeg] {\resnetimagenetjpegprobclasssemantics};
            \end{axis}
        \end{tikzpicture} &

        \cellcolor{orange!10}
        \begin{tikzpicture}
            \begin{axis}[mybarstyle2, title={SPD$_\text{p}$ - Resizing}, xlabel={$p_c$}]
            \addplot[fill=supervisedcolor, draw=black, bar shift=0pt] table[x expr=\coordindex, y=resizing] {\resnetimagenetjpegprobclasssemantics};
            \addplot[fill=supervisedcolor, draw=black, postaction={pattern=crosshatch dots}, bar shift=0pt, restrict x to domain=6.5:7.5] table[x expr=\coordindex, y=resizing] {\resnetimagenetjpegprobclasssemantics};
            \end{axis}
        \end{tikzpicture} &

        \cellcolor{orange!10}
        \begin{tikzpicture}
            \begin{axis}[mybarstyle2, title={SPD$_\text{a}$ - Camera}, ylabel={$\Delta_a$}, xlabel={$p_c$}, ymin=1, ymax=11]
            \addplot[fill=supervisedcolor, draw=black, bar shift=0pt] table[x expr=\coordindex, y=spd-a] {\resnetimagenetjpegprobclassinversion};
            \addplot[fill=supervisedcolor, draw=black, postaction={pattern=crosshatch dots}, bar shift=0pt, restrict x to domain=6.5:7.5] table[x expr=\coordindex, y=spd-a] {\resnetimagenetjpegprobclassinversion};
            \end{axis}
        \end{tikzpicture} \\

    \end{tabular}
    \caption{\textbf{Impact of correlations between semantics and processing metadata on the model's sensitivity to metadata in a class controlled setting.}
    ResNet50 trained on different versions of IN1k with 
    manipulated JPEG labels. 
    We use probability $p_c$ to control the fraction of classes that exhibit correlations between JPEG labels and semantic labels, instead of using $p_i$ shown in \cref{fig:imagenet_image_prob}. 
    The larger the $p_c$, the stronger the correlation between JPEG labels and semantic labels. 
    \emph{None}: model trained on original IN1k. 
    Difference between $p_c=0$ and \emph{none}: the former has no JPEG-semantics correlations, while the latter has the original IN1k correlations.
    SP: two types of test sets are used, the original one (none, \textcolor{supervisedcolor}{green} color) which is fixed across values of $p_c$, and 7 different versions (\textcolor{probcolor}{purple} color) created by manipulating the JPEG labels of the original test set in the same way as the corresponding training set for each value of $p_c$.
    }
    \label{fig:imagenet_class_prob}
\end{figure}

\section{Class controlled correlations of semantics and processing metadata}
\label{sec:class_controlled_supp}
In~\cref{sec:processing_correlations}, we use probability $p_i$ to control the percentage of images of each class that exibit correlations of semantics and processing metadata. We now take a different approach and we randomly assign one specific metadata label only to a fraction $p_c$ of classes.
Images belonging to the remaining classes are randomly assigned to their metadata label.
Thus, $p_c$ controls the proportion of classes exhibiting correlations of semantics and processing metadata.

\cref{fig:imagenet_class_prob} presents the results of this experiment, where we observe similar trends as in~\cref{fig:imagenet_image_prob} (\cref{sec:processing_correlations}), that the models trained on data with stronger correlations between metadata and semantic labels are more sensitive to metadata.

Additionally, among the two ways ($p_i$ and $p_c$) of creating correlations, the one that controls the number of correlated classes seems to have a stronger effect on the model's ability to encode metadata information and on the sensitivity of semantic prediction abilities to metadata. This suggests that having whole classes exhibiting correlation between metadata and semantics has a stronger effect on the model's ability to encode metadata information than having only a fraction of images of each class exhibiting such correlation.

\begin{figure}[t]
    \centering
    \vspace{10pt}
    \input{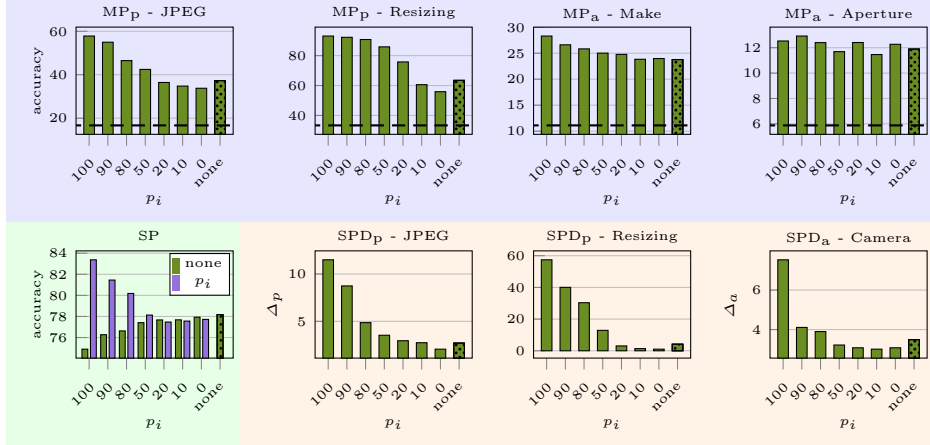}
\pgfplotsset{
    mybarstyle2/.style={
        ybar=1pt, 
        bar width=4pt,
        width=0.3\linewidth,
        height=3cm,
        enlarge x limits=0.1,
        xtick=data,
        xticklabels from table={\resnetimagenetjpegprobimageinversion}{p},
        x tick label style={font=\tiny, rotate=45, anchor=north east},
        title style={font=\tiny, yshift=-7pt},
        label style={font=\tiny},
        tick label style={font=\tiny},
        y label style={yshift=-4pt},
        x label style={yshift=5pt},
        ymajorgrids=true,
        xtick pos=bottom,
        ytick pos=left,
    }
}

\centering
    \begin{tabular}{cccc}

        \cellcolor{blue!10}
        \begin{tikzpicture}
            \begin{axis}[mybarstyle2, title={MP$_\text{p}$ - JPEG}, ylabel={accuracy}, xlabel={$p_{i}$}]
            \addplot[fill=supervisedcolor, draw=black, bar shift=0pt] table[x expr=\coordindex, y=jpeg] {\resnetimagenetresizeprobimageinversion};
            \addplot[fill=supervisedcolor, draw=black, postaction={pattern=crosshatch dots}, bar shift=0pt, restrict x to domain=6.5:7.5] table[x expr=\coordindex, y=jpeg] {\resnetimagenetresizeprobimageinversion};
            \addplot[
                black, 
                dashed, 
                thick, 
                sharp plot, %
                dash pattern=on 5pt off 3pt,
                mark=none,  %
                shorten <=-8pt, %
                shorten >=-8pt  %
            ] coordinates {(0, 16.67) (7, 16.67)};
            \end{axis}
        \end{tikzpicture} &
        
        \cellcolor{blue!10}
        \begin{tikzpicture}
            \begin{axis}[mybarstyle2, title={MP$_\text{p}$ - Resizing}, xlabel={$p_{i}$}, ylabel={\phantom{$\Delta_p$}}]
            \addplot[fill=supervisedcolor, draw=black, bar shift=0pt] table[x expr=\coordindex, y=resizing] {\resnetimagenetresizeprobimageinversion};
            \addplot[fill=supervisedcolor, draw=black, postaction={pattern=crosshatch dots}, bar shift=0pt, restrict x to domain=6.5:7.5] table[x expr=\coordindex, y=resizing] {\resnetimagenetresizeprobimageinversion};
            \addplot[
                black, 
                dashed, 
                thick, 
                sharp plot, %
                dash pattern=on 5pt off 3pt,
                mark=none,  %
                shorten <=-8pt, %
                shorten >=-8pt  %
            ] coordinates {(0, 33.33) (7, 33.33)};
            \end{axis}
        \end{tikzpicture} &

        \cellcolor{blue!10}
        \begin{tikzpicture}
            \begin{axis}[mybarstyle2, title={MP$_\text{a}$ - Make}, xlabel={$p_{i}$}]
            \addplot[fill=supervisedcolor, draw=black, bar shift=0pt] table[x expr=\coordindex, y=make] {\resnetimagenetresizeprobimageinversion};
            \addplot[fill=supervisedcolor, draw=black, postaction={pattern=crosshatch dots}, bar shift=0pt, restrict x to domain=6.5:7.5] table[x expr=\coordindex, y=make] {\resnetimagenetresizeprobimageinversion};
            \addplot[
                black, 
                dashed, 
                thick, 
                sharp plot, %
                dash pattern=on 5pt off 3pt,
                mark=none,  %
                shorten <=-8pt, %
                shorten >=-8pt  %
            ] coordinates {(0, 11.11) (7, 11.11)};
            \end{axis}
        \end{tikzpicture} &

        \cellcolor{blue!10}
        \begin{tikzpicture}
            \begin{axis}[mybarstyle2, title={MP$_\text{a}$ - Aperture}, xlabel={$p_{i}$}, ylabel={\phantom{$\Delta_a$}}]
            \addplot[fill=supervisedcolor, draw=black, bar shift=0pt] table[x expr=\coordindex, y=aperture] {\resnetimagenetresizeprobimageinversion};
            \addplot[fill=supervisedcolor, draw=black, postaction={pattern=crosshatch dots}, bar shift=0pt, restrict x to domain=6.5:7.5] table[x expr=\coordindex, y=aperture] {\resnetimagenetresizeprobimageinversion};
            \addplot[
                black, 
                dashed, 
                thick, 
                sharp plot, %
                dash pattern=on 5pt off 3pt,
                mark=none,  %
                shorten <=-8pt, %
                shorten >=-8pt  %
            ] coordinates {(0, 5.88) (7, 5.88)};
            \end{axis}
        \end{tikzpicture} \\

        \cellcolor{green!10}
        \begin{tikzpicture}
            \begin{axis}[mybarstyle2, title={SP\phantom{$_\text{p}$}}, ylabel={accuracy}, xlabel={$p_{i}$}, bar width=2.2pt, bar shift=0pt, legend style={font=\tiny, scale=0.6, /tikz/every even column/.append style={column sep=0.1cm}, inner sep=0.8pt, row sep=-1pt}, legend image code/.code={
                    \draw[#1, draw=none] (0cm,-0.1cm) rectangle (0.1cm,0.1cm); %
                }]
            \addplot[draw=none, fill=none, bar shift=0pt, forget plot] table[x expr=\coordindex, y=val-knn] {\resnetimagenetresizeprobimageinversion};
            \addplot[fill=supervisedcolor, draw=black, bar shift=-1.6pt, restrict x to domain=-0.5:6.5] table[x expr=\coordindex, y=val-knn] {\resnetimagenetresizeprobimageinversion};\addlegendentry{none};
            \addplot[fill=probcolor, draw=black, bar shift=1.6pt, restrict x to domain=-0.5:6.5] table[x expr=\coordindex, y=val-bias]{\resnetimagenetresizeprobimageinversion};\addlegendentry{$p_i$};
            \addplot[fill=supervisedcolor, draw=black, bar shift=0pt, restrict x to domain=6.5:7.5, forget plot] table[x expr=\coordindex, y=val-knn] {\resnetimagenetresizeprobimageinversion};
            \addplot[fill=supervisedcolor, draw=black, postaction={pattern=crosshatch dots}, bar shift=0pt, restrict x to domain=6.5:7.5, forget plot] table[x expr=\coordindex, y=val-knn] {\resnetimagenetresizeprobimageinversion};
            \end{axis}
        \end{tikzpicture} &

        \cellcolor{orange!10}
        \begin{tikzpicture}
            \begin{axis}[mybarstyle2, title={SPD$_\text{p}$ - JPEG}, ylabel={$\Delta_p$}, xlabel={$p_{i}$}]
            \addplot[fill=supervisedcolor, draw=black, bar shift=0pt] table[x expr=\coordindex, y=jpeg] {\resnetimagenetresizeprobimagesemantics};
            \addplot[fill=supervisedcolor, draw=black, postaction={pattern=crosshatch dots}, bar shift=0pt, restrict x to domain=6.5:7.5] table[x expr=\coordindex, y=jpeg] {\resnetimagenetresizeprobimagesemantics};
            \end{axis}
        \end{tikzpicture} &

        \cellcolor{orange!10}
        \begin{tikzpicture}
            \begin{axis}[mybarstyle2, title={SPD$_\text{p}$ - Resizing}, xlabel={$p_{i}$}]
            \addplot[fill=supervisedcolor, draw=black, bar shift=0pt] table[x expr=\coordindex, y=resizing] {\resnetimagenetresizeprobimagesemantics};
            \addplot[fill=supervisedcolor, draw=black, postaction={pattern=crosshatch dots}, bar shift=0pt, restrict x to domain=6.5:7.5] table[x expr=\coordindex, y=resizing] {\resnetimagenetresizeprobimagesemantics};
            \end{axis}
        \end{tikzpicture} &

        \cellcolor{orange!10}
        \begin{tikzpicture}
            \begin{axis}[mybarstyle2, title={SPD$_\text{a}$ - Camera}, ylabel={$\Delta_a$}, xlabel={$p_{i}$}, ytick={4,6}, yticklabels={\phantom{0}4,6}]
            \addplot[fill=supervisedcolor, draw=black, bar shift=0pt] table[x expr=\coordindex, y=spd-a] {\resnetimagenetresizeprobimagesemantics};
            \addplot[fill=supervisedcolor, draw=black, postaction={pattern=crosshatch dots}, bar shift=0pt, restrict x to domain=6.5:7.5] table[x expr=\coordindex, y=spd-a] {\resnetimagenetresizeprobimagesemantics};
            \end{axis}
        \end{tikzpicture} \\
        
    \end{tabular}
    \caption{\textbf{Impact of correlations between semantics and processing metadata on the model's sensitivity to metadata.
    } ResNet50 trained on different versions of IN1k with 
    manipulated resize labels instead of JPEG labels in \cref{fig:imagenet_image_prob}. 
    The larger the $p_i$, the stronger the correlation between resize labels and semantic labels. 
    \emph{None}: model trained on original IN1k. 
    Difference between $p_i=0$ and \emph{none}: the former has no resize-semantics correlations, while the latter has the original IN1k correlations.
    SP: two types of test sets are used, the original one (none, \textcolor{supervisedcolor}{green} color) which is fixed across values of $p_i$, and 7 different versions (\textcolor{probcolor}{purple} color) created by manipulating the resize labels of the original test set in the same way as the corresponding training set for each value of $p_i$.
    }
    \label{fig:imagenet_image_prob_resize}
\end{figure}

\begin{figure}[t]
    \centering
    \vspace{15pt}
    \input{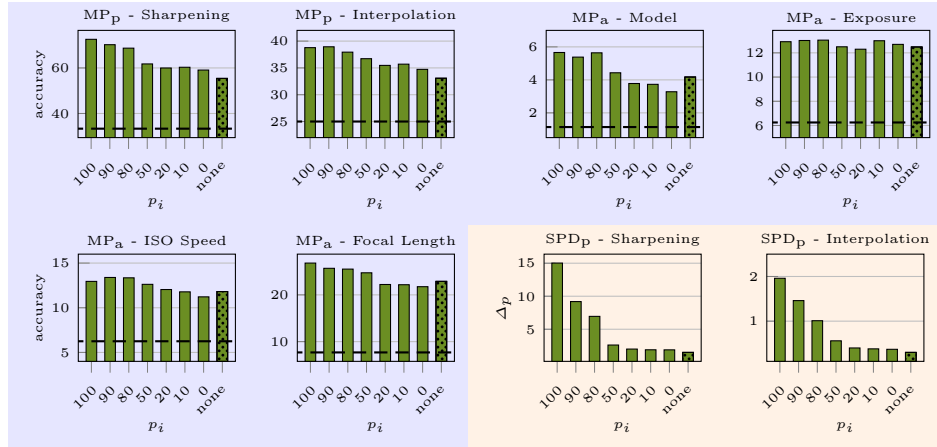}
\pgfplotsset{
    mybarstyle2/.style={
        ybar=1pt, 
        bar width=4pt,
        width=0.3\linewidth,
        height=3cm,
        enlarge x limits=0.1,
        xtick=data,
        xticklabels from table={\resnetimagenetjpegprobimageinversion}{p},
        x tick label style={font=\tiny, rotate=45, anchor=north east},
        title style={font=\tiny, yshift=-7pt},
        label style={font=\tiny},
        tick label style={font=\tiny},
        y label style={yshift=-4pt},
        x label style={yshift=5pt},
        ymajorgrids=true,
        xtick pos=bottom,
        ytick pos=left,
    }
}

\centering
    \begin{tabular}{cccc}

        \cellcolor{blue!10}
        \begin{tikzpicture}
            \begin{axis}[mybarstyle2, title={MP$_\text{p}$ - Sharpening}, ylabel={accuracy}, xlabel={$p_{i}$}]
            \addplot[fill=supervisedcolor, draw=black, bar shift=0pt] table[x expr=\coordindex, y=sharpening] {\resnetimagenetjpegprobimageinversion};
            \addplot[fill=supervisedcolor, draw=black, postaction={pattern=crosshatch dots}, bar shift=0pt, restrict x to domain=6.5:7.5] table[x expr=\coordindex, y=sharpening] {\resnetimagenetjpegprobimageinversion};
            \addplot[
                black, 
                dashed, 
                thick, 
                sharp plot, %
                dash pattern=on 5pt off 3pt,
                mark=none,  %
                shorten <=-8pt, %
                shorten >=-8pt  %
            ] coordinates {(0, 33.33) (7, 33.33)};
            \end{axis}
        \end{tikzpicture} &
        
        \cellcolor{blue!10}
        \begin{tikzpicture}
            \begin{axis}[mybarstyle2, title={MP$_\text{p}$ - Interpolation}, xlabel={$p_{i}$}, ymin=22, ymax=42]
            \addplot[fill=supervisedcolor, draw=black, bar shift=0pt] table[x expr=\coordindex, y=interpolation] {\resnetimagenetjpegprobimageinversion};
            \addplot[fill=supervisedcolor, draw=black, postaction={pattern=crosshatch dots}, bar shift=0pt, restrict x to domain=6.5:7.5] table[x expr=\coordindex, y=interpolation] {\resnetimagenetjpegprobimageinversion};
            \addplot[
                black, 
                dashed, 
                thick, 
                sharp plot, %
                dash pattern=on 5pt off 3pt,
                mark=none,  %
                shorten <=-8pt, %
                shorten >=-8pt  %
            ] coordinates {(0, 25) (7, 25)};
            \end{axis}
        \end{tikzpicture} &

        \cellcolor{blue!10}
        \begin{tikzpicture}
            \begin{axis}[mybarstyle2, title={MP$_\text{a}$ - Model}, xlabel={$p_{i}$}, ylabel={\phantom{$\Delta_p$}}, ymin=0.5, ymax=7, ytick={2,4,6}, yticklabels={\phantom{1}2,4,6}]
            \addplot[fill=supervisedcolor, draw=black, bar shift=0pt] table[x expr=\coordindex, y=model] {\resnetimagenetjpegprobimageinversion};
            \addplot[fill=supervisedcolor, draw=black, postaction={pattern=crosshatch dots}, bar shift=0pt, restrict x to domain=6.5:7.5] table[x expr=\coordindex, y=model] {\resnetimagenetjpegprobimageinversion};
            \addplot[
                black, 
                dashed, 
                thick, 
                sharp plot, %
                dash pattern=on 5pt off 3pt,
                mark=none,  %
                shorten <=-8pt, %
                shorten >=-8pt  %
            ] coordinates {(0, 1.14) (7, 1.14)};
            \end{axis}
        \end{tikzpicture} &

        \cellcolor{blue!10}
        \begin{tikzpicture}
            \begin{axis}[mybarstyle2, title={MP$_\text{a}$ - Exposure}, xlabel={$p_{i}$}, ymin=5]
            \addplot[fill=supervisedcolor, draw=black, bar shift=0pt] table[x expr=\coordindex, y=exposure] {\resnetimagenetjpegprobimageinversion};
            \addplot[fill=supervisedcolor, draw=black, postaction={pattern=crosshatch dots}, bar shift=0pt, restrict x to domain=6.5:7.5] table[x expr=\coordindex, y=exposure] {\resnetimagenetjpegprobimageinversion};
            \addplot[
                black, 
                dashed, 
                thick, 
                sharp plot, %
                dash pattern=on 5pt off 3pt,
                mark=none,  %
                shorten <=-8pt, %
                shorten >=-8pt  %
            ] coordinates {(0, 6.25) (7, 6.25)};
            \end{axis}
        \end{tikzpicture} \\

        \cellcolor{blue!10}
        \begin{tikzpicture}
            \begin{axis}[mybarstyle2, title={MP$_\text{a}$ - ISO Speed}, xlabel={$p_{i}$}, ylabel={accuracy}, ymin=4, ymax=16]
            \addplot[fill=supervisedcolor, draw=black, bar shift=0pt] table[x expr=\coordindex, y=iso] {\resnetimagenetjpegprobimageinversion};
            \addplot[fill=supervisedcolor, draw=black, postaction={pattern=crosshatch dots}, bar shift=0pt, restrict x to domain=6.5:7.5] table[x expr=\coordindex, y=iso] {\resnetimagenetjpegprobimageinversion};
            \addplot[
                black, 
                dashed, 
                thick, 
                sharp plot, %
                dash pattern=on 5pt off 3pt,
                mark=none,  %
                shorten <=-8pt, %
                shorten >=-8pt  %
            ] coordinates {(0, 6.25) (7, 6.25)};
            \end{axis}
        \end{tikzpicture} &

        \cellcolor{blue!10}
        \begin{tikzpicture}
            \begin{axis}[mybarstyle2, title={MP$_\text{a}$ - Focal Length}, xlabel={$p_{i}$}]
            \addplot[fill=supervisedcolor, draw=black, bar shift=0pt] table[x expr=\coordindex, y=focal] {\resnetimagenetjpegprobimageinversion};
            \addplot[fill=supervisedcolor, draw=black, postaction={pattern=crosshatch dots}, bar shift=0pt, restrict x to domain=6.5:7.5] table[x expr=\coordindex, y=focal] {\resnetimagenetjpegprobimageinversion};
            \addplot[
                black, 
                dashed, 
                thick, 
                sharp plot, %
                dash pattern=on 5pt off 3pt,
                mark=none,  %
                shorten <=-8pt, %
                shorten >=-8pt  %
            ] coordinates {(0, 7.69) (7, 7.69)};
            \end{axis}
        \end{tikzpicture} &

        \cellcolor{orange!10}
        \begin{tikzpicture}
            \begin{axis}[mybarstyle2, title={SPD$_\text{p}$ - Sharpening}, ylabel={$\Delta_p$}, xlabel={$p_{i}$}]
            \addplot[fill=supervisedcolor, draw=black, bar shift=0pt] table[x expr=\coordindex, y=sharpening] {\resnetimagenetjpegprobimagesemantics};
            \addplot[fill=supervisedcolor, draw=black, postaction={pattern=crosshatch dots}, bar shift=0pt, restrict x to domain=6.5:7.5] table[x expr=\coordindex, y=sharpening] {\resnetimagenetjpegprobimagesemantics};
            \end{axis}
        \end{tikzpicture} &

        \cellcolor{orange!10}
        \begin{tikzpicture}
            \begin{axis}[mybarstyle2, title={SPD$_\text{p}$ - Interpolation}, xlabel={$p_{i}$}, ymin=0.1, ymax=2.5, ytick={1,2}, yticklabels={\phantom{0}1,2}]
            \addplot[fill=supervisedcolor, draw=black, bar shift=0pt] table[x expr=\coordindex, y=interpolation] {\resnetimagenetjpegprobimagesemantics};
            \addplot[fill=supervisedcolor, draw=black, postaction={pattern=crosshatch dots}, bar shift=0pt, restrict x to domain=6.5:7.5] table[x expr=\coordindex, y=interpolation] {\resnetimagenetjpegprobimagesemantics};
            \end{axis}
        \end{tikzpicture} \\
        
    \end{tabular}
    \vspace{8pt}
    \caption{\textbf{Impact of correlations between semantics and processing metadata on the model's sensitivity to metadata.
    } Diagnostic metrics for additional metadata attributes compared to \cref{fig:imagenet_image_prob} ResNet50 trained on different versions of IN1k with 
    manipulated JPEG labels. 
    The larger the $p_i$, the stronger the correlation between JPEG labels and semantic labels. 
    \emph{None}: model trained on original IN1k. 
    Difference between $p_i=0$ and \emph{none}: the former has no JPEG-semantics correlations, while the latter has the original IN1k correlations.
    SP: two types of test sets are used, the original one (none, \textcolor{supervisedcolor}{green} color) which is fixed across values of $p_i$, and 7 different versions (\textcolor{probcolor}{purple} color) created by manipulating the JPEG labels of the original test set in the same way as the corresponding training set for each value of $p_i$.
    }
    \vspace{15pt}
    \label{fig:imagenet_image_prob_other}
\end{figure}

\section{Controlling correlations of semantics and processing metadata using resize}
\label{sec:resize_supp}

In~\cref{sec:processing_correlations}, we introduce correlations of semantics and processing metadata by recompressing images based on their semantic label. To show that our observations are not dependent on the choice of processing metadata, instead of recompressing, we resize images based on their semantic label. We define $13$ metadata labels that control the amount of upsampling or downsampling of the original image. All images are then %
resized according to one of these labels, creating a correlation between semantic labels and metadata labels. \cref{fig:imagenet_image_prob_resize} presents the results of this experiment. We observe that models trained on data with stronger correlations between metadata and semantic labels are more sensitive to metadata.

\section{Diagnostic metrics for additional metadata attributes}
\label{sec:other_supp}

\begin{figure}[t]
    \centering
    \vspace{-5pt}
    \input{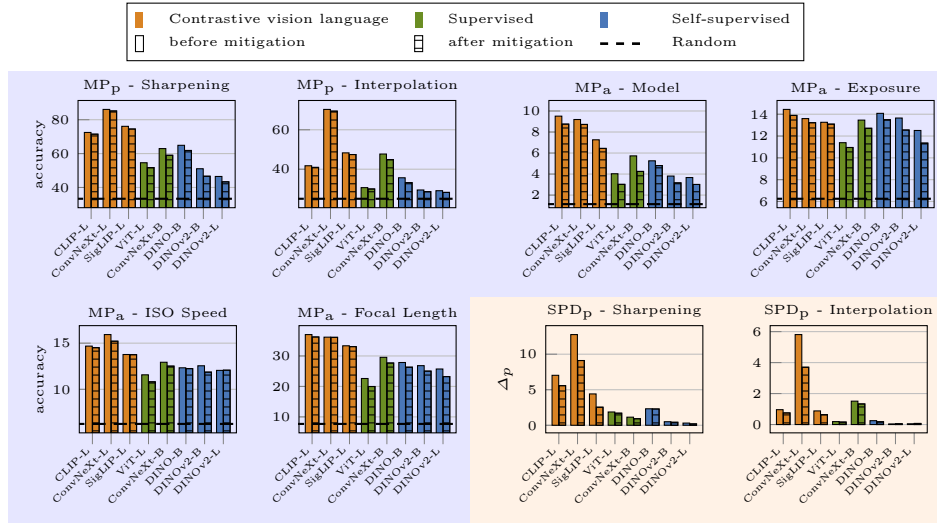}
\pgfplotsset{
    my 4bar style/.style={
        ybar,
        width=0.3\linewidth,
        bar width=2.5pt,
        bar shift=-1.3pt,
        height=3cm,
        enlarge x limits=0.1,
        xtick={0,1,2,3,4,5,6,7}, %
        label style={font=\tiny},
        tick label style={font=\tiny},
        xticklabels = {CLIP-L, ConvNeXt-L, SigLIP-L, ViT-L, ConvNeXt-B, DINO-B, DINOv2-B, DINOv2-L},
        x tick label style={rotate=45, anchor=east, font=\tiny, scale=0.75},
        title style={font=\tiny, yshift=-7pt},
        y label style={yshift=-4pt},
        x label style={yshift=5pt},
        ymajorgrids=true,
        xtick pos=bottom,
        ytick pos=left,
    }
}

\newcommand{\addplotpair}[1]{
    \addplot[fill=vlmcolor, draw=black, restrict x to domain=0:2] table[x=idx, y={#1 orig}] {\debiasprocessinginversion};
    \addplot[preaction={fill=vlmcolor}, draw=black, pattern=horizontal lines, bar shift=1.3pt, restrict x to domain=0:2] table[x=idx, y={#1 deb}] {\debiasprocessinginversion};
    
    \addplot[fill=supervisedcolor, draw=black, restrict x to domain=3:4] table[x=idx, y={#1 orig}] {\debiasprocessinginversion};
    \addplot[preaction={fill=supervisedcolor}, draw=black, pattern=horizontal lines, bar shift=1.3pt, restrict x to domain=3:4] table[x=idx, y={#1 deb}] {\debiasprocessinginversion};
    
    \addplot[fill=sslcolor, draw=black, restrict x to domain=5:7] table[x=idx, y={#1 orig}] {\debiasprocessinginversion};
    \addplot[preaction={fill=sslcolor}, draw=black, pattern=horizontal lines, bar shift=1.3pt, restrict x to domain=5:7] table[x=idx, y={#1 deb}] {\debiasprocessinginversion};
}

\newcommand{\addplotpairsem}[1]{
    \addplot[fill=vlmcolor, draw=black, restrict x to domain=0:2] table[x=idx, y={#1 orig}] {\debiasprocessingsemantics};
    \addplot[preaction={fill=vlmcolor}, draw=black, pattern=horizontal lines, bar shift=1.3pt, restrict x to domain=0:2] table[x=idx, y={#1 deb}] {\debiasprocessingsemantics};
    
    \addplot[fill=supervisedcolor, draw=black, restrict x to domain=3:4] table[x=idx, y={#1 orig}] {\debiasprocessingsemantics};
    \addplot[preaction={fill=supervisedcolor}, draw=black, pattern=horizontal lines, bar shift=1.3pt, restrict x to domain=3:4] table[x=idx, y={#1 deb}] {\debiasprocessingsemantics};
    
    \addplot[fill=sslcolor, draw=black, restrict x to domain=5:7] table[x=idx, y={#1 orig}] {\debiasprocessingsemantics};
    \addplot[preaction={fill=sslcolor}, draw=black, pattern=horizontal lines, bar shift=1.3pt, restrict x to domain=5:7] table[x=idx, y={#1 deb}] {\debiasprocessingsemantics};
}

\pgfplotsset{
    ybar legend/.style={
        legend image code/.code={
            \fill[#1] (0pt,-3pt) rectangle (3pt,3pt);
        }
    }
}

\begin{tikzpicture}
    \begin{axis}[
        hide axis,
        scale only axis,
        height=0cm,
        width=0cm,
        legend style={
            draw=black,
            cells={anchor=west},
            legend columns=3,
            column sep=0.8em,
            font=\tiny,
            anchor=north west
        },
        legend entries={
            {Contrastive vision language},
            {Supervised},
            {Self-supervised},
            {before mitigation},
            {after mitigation},
            {Random}
        }
    ]
        \addlegendimage{ybar, fill=vlmcolor, draw=none, ybar legend}
        \addlegendimage{ybar, fill=supervisedcolor, draw=none, ybar legend}
        \addlegendimage{ybar, fill=sslcolor, draw=none, ybar legend}
        \addlegendimage{ybar, fill=white, draw=black, ybar legend}
        \addlegendimage{ybar, fill=white, draw=black, pattern=horizontal lines, ybar legend}
        \addlegendimage{color=black, line width=1, dashed}
        
        \addplot[draw=none] coordinates {(0,0)};
    \end{axis}
\end{tikzpicture}
\vspace{3pt}

\centering
    \begin{tabular}{cccc}

        \cellcolor{blue!10}
        \begin{tikzpicture}
            \begin{axis}[my 4bar style, title={MP$_\text{p}$ - Sharpening}, ylabel={accuracy}]
                \addplotpair{sharpening};
                \addplot[
                black, 
                dashed, 
                thick, 
                sharp plot, %
                dash pattern=on 5pt off 3pt,
                mark=none,  %
                shorten <=-8pt, %
                shorten >=-8pt  %
            ] coordinates {(0, 33.33) (7, 33.33)};
            \end{axis}
        \end{tikzpicture} &

        \cellcolor{blue!10}
        \begin{tikzpicture}
            \begin{axis}[my 4bar style, title={MP$_\text{p}$ - Interpolation}]
                \addplotpair{interpolation};
                \addplot[
                black, 
                dashed, 
                thick, 
                sharp plot, %
                dash pattern=on 5pt off 3pt,
                mark=none,  %
                shorten <=-8pt, %
                shorten >=-8pt  %
            ] coordinates {(0, 25) (7, 25)};
            \end{axis}
        \end{tikzpicture} &

        \cellcolor{blue!10}
        \begin{tikzpicture}
            \begin{axis}[my 4bar style, title={MP$_\text{a}$ - Model}, ylabel={\phantom{$\Delta_p$}}, ymin=0.8, ymax=11]
                \addplotpair{model};
                \addplot[
                black, 
                dashed, 
                thick, 
                sharp plot, %
                dash pattern=on 5pt off 3pt,
                mark=none,  %
                shorten <=-8pt, %
                shorten >=-8pt  %
            ] coordinates {(0, 1.14) (7, 1.14)};
            \end{axis}
        \end{tikzpicture} &

        \cellcolor{blue!10}
        \begin{tikzpicture}
            \begin{axis}[my 4bar style, title={MP$_\text{a}$ - Exposure}]
                \addplotpair{exposure};
                \addplot[
                black, 
                dashed, 
                thick, 
                sharp plot, %
                dash pattern=on 5pt off 3pt,
                mark=none,  %
                shorten <=-8pt, %
                shorten >=-8pt  %
            ] coordinates {(0, 6.25) (7, 6.25)};
            \end{axis}
        \end{tikzpicture} \\

        \cellcolor{blue!10}
        \begin{tikzpicture}
            \begin{axis}[my 4bar style, title={MP$_\text{a}$ - ISO Speed}, ylabel={accuracy}]
                \addplotpair{iso};
                \addplot[
                black, 
                dashed, 
                thick, 
                sharp plot, %
                dash pattern=on 5pt off 3pt,
                mark=none,  %
                shorten <=-8pt, %
                shorten >=-8pt  %
            ] coordinates {(0, 6.25) (7, 6.25)};
            \end{axis}
        \end{tikzpicture} &

        \cellcolor{blue!10}
        \begin{tikzpicture}
            \begin{axis}[my 4bar style, title={MP$_\text{a}$ - Focal Length}]
                \addplotpair{focal};
                \addplot[
                black, 
                dashed, 
                thick, 
                sharp plot, %
                dash pattern=on 5pt off 3pt,
                mark=none,  %
                shorten <=-8pt, %
                shorten >=-8pt  %
            ] coordinates {(0, 7.69) (7, 7.69)};
            \end{axis}
        \end{tikzpicture} &

        \cellcolor{orange!10}
        \begin{tikzpicture}
            \begin{axis}[my 4bar style, title={SPD$_\text{p}$ - Sharpening}, ylabel={$\Delta_p$}]
                \addplotpairsem{sharpening sem}
            \end{axis}
        \end{tikzpicture} &

        \cellcolor{orange!10}
         \begin{tikzpicture}
            \begin{axis}[my 4bar style, title={SPD$_\text{p}$ - Interpolation}, ytick={0,2,4,6}, yticklabels={\phantom{0}0,2,4,6}]
                \addplotpairsem{interpolation sem}
            \end{axis}
        \end{tikzpicture} \\
        
    \end{tabular}
    \vspace{-5pt}
    \caption{\textbf{Post-hoc adversarial mitigation with a linear layer -- impact on model sensitivity to metadata}. Diagnostic metrics for additional metadata attributes compared to \cref{fig:debias}. 
    }
    \vspace{-10pt}
    \label{fig:debias_other}
\end{figure}

In the main paper, we evaluate models on two processing attributes (JPEG and resizing) and two acquisition attributes (make and aperture). Here, we present results for other processing and acquisition attributes from the benchmark by Ramos \etal~\cite{ramos2025traces}. \cref{fig:imagenet_image_prob_other} presents the results for the ResNet50 model trained on the data with correlations between JPEG labels and semantics labels (\cref{sec:processing_correlations}) for additional metadata attributes. We observe that models trained on data with stronger correlations are more sensitive to metadata, showing that our observations are not dependent on the metadata attribute used during evaluation. Additionally, in \cref{fig:debias_other}, we present the results of our post-hoc adversarial mitigation approach (\cref{sec:mitigation_posthoc}) when applied to these other metadata attributes. We observe that our mitigation approach reduces sensitivity to metadata, demonstrating that it generalizes across metadata attributes.

\begin{table}[b]
    \centering
    
\resizebox{\linewidth}{!}{
\begin{tabular}{cccccccccc}
        \toprule
        aug & post-hoc & \cellcolor{blue!10} MP$_\text{p}$ - JPEG & \cellcolor{blue!10} MP$_\text{p}$ - Resizing & \cellcolor{blue!10} MP$_\text{a}$ - Make & \cellcolor{blue!10} MP$_\text{a}$ - Aperture & \cellcolor{green!10} SP & \cellcolor{orange!10} SPD$_\text{p}$ - JPEG & \cellcolor{orange!10} SPD$_\text{p}$ - Resizing & \cellcolor{orange!10} SPD$_\text{a}$ - Camera \\
        \midrule
        \ding{55} & \ding{55} & \cellcolor{blue!10} 37.3 & \cellcolor{blue!10} 63.4 & \cellcolor{blue!10} 23.8 & \cellcolor{blue!10} 11.9 & \cellcolor{green!10} 78.2 & \cellcolor{orange!10} 2.7 & \cellcolor{orange!10} 4.2 & \cellcolor{orange!10} 3.5 \\
        \ding{51} & \ding{55} & \cellcolor{blue!10} 30.7 & \cellcolor{blue!10} 53.1 & \cellcolor{blue!10} 21.4 & \cellcolor{blue!10} 11.1 & \cellcolor{green!10} 77.3 & \cellcolor{orange!10} 1.4 & \cellcolor{orange!10} 1.8 & \cellcolor{orange!10} 1.6 \\
        \ding{55} & \ding{51} & \cellcolor{blue!10} 31.2 & \cellcolor{blue!10} 56.1 & \cellcolor{blue!10} 19.4 & \cellcolor{blue!10} 11.4 & \cellcolor{green!10} 77.2 & \cellcolor{orange!10} 0.8 & \cellcolor{orange!10} 2.3 & \cellcolor{orange!10} 2.3 \\
        \ding{51} & \ding{51} & \cellcolor{blue!10} 24.7 & \cellcolor{blue!10} 46.9 & \cellcolor{blue!10} 18.9 & \cellcolor{blue!10} 10.5 & \cellcolor{green!10} 76.5 & \cellcolor{orange!10} 0.3 & \cellcolor{orange!10} 0.7 & \cellcolor{orange!10} 1.6 \\
        \bottomrule
\end{tabular}
}

    \caption{\textbf{Comparison and combination of the mitigation methods.} ResNet50 trained on original IN1k. 
    }
    \label{tab:mitigation_compare}
\end{table}

\section{Comparison of mitigation approaches}
\label{sec:mitigation_compare_supp}

The two proposed mitigation approaches are applicable in different settings and assumptions: \emph{augmentations} during the original model training and \emph{post-hoc} on pre-trained models. However, the results in \cref{tab:mitigation_compare} indicate they complement each other, while performing roughly the same if used separately.

\section{Ablation of the post-hoc adversarial mitigation}
\label{sec:ablation_supp}

\begin{figure}
    \centering
    \input{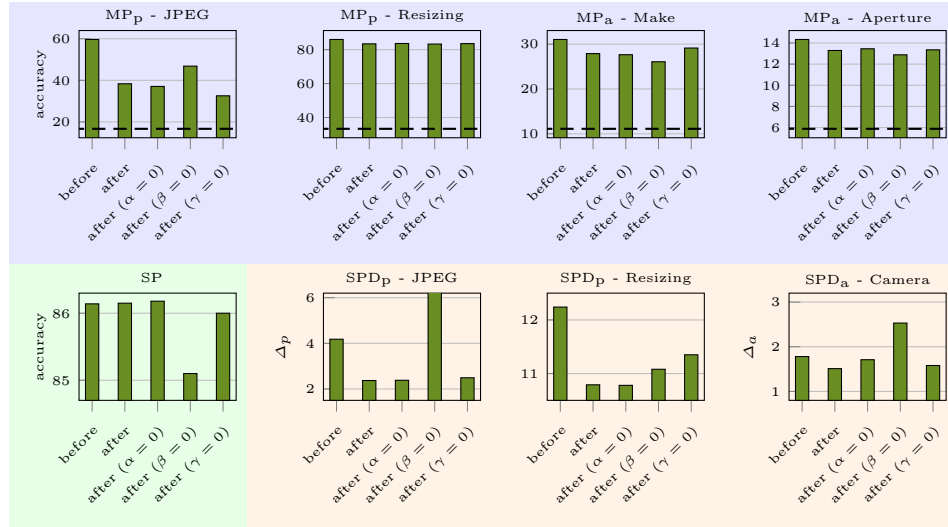}
\pgfplotsset{
    mybarstyle2/.style={
        ybar=1pt, 
        bar width=5pt,
        width=0.3\linewidth,
        height=3cm,
        enlarge x limits=0.1,
        xtick=data,
        xticklabels from table={\mitigationablation}{variant},
        x tick label style={font=\tiny, rotate=45, anchor=north east},
        title style={font=\tiny, yshift=-7pt},
        label style={font=\tiny},
        tick label style={font=\tiny},
        y label style={yshift=-4pt},
        x label style={yshift=5pt},
        ymajorgrids=true,
        xtick pos=bottom,
        ytick pos=left,
    }
}

\centering
    \begin{tabular}{cccc}

        \cellcolor{blue!10}
        \begin{tikzpicture}
            \begin{axis}[mybarstyle2, title={MP$_\text{p}$ - JPEG}, ylabel={accuracy}, xlabel={}]
            \addplot[fill=supervisedcolor, draw=black, bar shift=0pt] table[x expr=\coordindex, y=jpeg] {\mitigationablation};
            \addplot[
                black, 
                dashed, 
                thick, 
                sharp plot, %
                dash pattern=on 5pt off 3pt,
                mark=none,  %
                shorten <=-8pt, %
                shorten >=-8pt  %
            ] coordinates {(0, 16.67) (4, 16.67)};
            \end{axis}
        \end{tikzpicture} &
        
        \cellcolor{blue!10}
        \begin{tikzpicture}
            \begin{axis}[mybarstyle2, title={MP$_\text{p}$ - Resizing}, xlabel={}, ylabel={\phantom{$\Delta_p$}}]
            \addplot[fill=supervisedcolor, draw=black, bar shift=0pt] table[x expr=\coordindex, y=resizing] {\mitigationablation};
            \addplot[
                black, 
                dashed, 
                thick, 
                sharp plot, %
                dash pattern=on 5pt off 3pt,
                mark=none,  %
                shorten <=-8pt, %
                shorten >=-8pt  %
            ] coordinates {(0, 33.33) (4, 33.33)};
            \end{axis}
        \end{tikzpicture} &

        \cellcolor{blue!10}
        \begin{tikzpicture}
            \begin{axis}[mybarstyle2, title={MP$_\text{a}$ - Make}, xlabel={}]
            \addplot[fill=supervisedcolor, draw=black, bar shift=0pt] table[x expr=\coordindex, y=make] {\mitigationablation};
            \addplot[
                black, 
                dashed, 
                thick, 
                sharp plot, %
                dash pattern=on 5pt off 3pt,
                mark=none,  %
                shorten <=-8pt, %
                shorten >=-8pt  %
            ] coordinates {(0, 11.11) (4, 11.11)};
            \end{axis}
        \end{tikzpicture} &

        \cellcolor{blue!10}
        \begin{tikzpicture}
            \begin{axis}[mybarstyle2, title={MP$_\text{a}$ - Aperture}, xlabel={}, ylabel={\phantom{$\Delta_a$}}]
            \addplot[fill=supervisedcolor, draw=black, bar shift=0pt] table[x expr=\coordindex, y=aperture] {\mitigationablation};
            \addplot[
                black, 
                dashed, 
                thick, 
                sharp plot, %
                dash pattern=on 5pt off 3pt,
                mark=none,  %
                shorten <=-8pt, %
                shorten >=-8pt  %
            ] coordinates {(0, 5.88) (4, 5.88)};
            \end{axis}
        \end{tikzpicture} \\

        \cellcolor{green!10}
        \begin{tikzpicture}
            \begin{axis}[mybarstyle2, title={SP\phantom{$_\text{p}$}}, ylabel={accuracy}, xlabel={}, ymin=84.7, ymax=86.3, ytick={85,86}]
            \addplot[fill=supervisedcolor, draw=black] table[x expr=\coordindex, y=val-knn] {\mitigationablation};
            \end{axis}
        \end{tikzpicture} &

        \cellcolor{orange!10}
        \begin{tikzpicture}
            \begin{axis}[mybarstyle2, title={SPD$_\text{p}$ - JPEG}, ylabel={$\Delta_p$}, xlabel={}, ymin=1.5, ymax=6.2, ytick={2,4,6}, yticklabels={\phantom{0}2,4,6}]
            \addplot[fill=supervisedcolor, draw=black, bar shift=0pt] table[x expr=\coordindex, y=jpeg] {\mitigationablationsemantics};
            \end{axis}
        \end{tikzpicture} &

        \cellcolor{orange!10}
        \begin{tikzpicture}
            \begin{axis}[mybarstyle2, title={SPD$_\text{p}$ - Resizing}, xlabel={}, ymin=10.5, ymax=12.5, ytick={11,12}]
            \addplot[fill=supervisedcolor, draw=black, bar shift=0pt] table[x expr=\coordindex, y=resizing] {\mitigationablationsemantics};
            \end{axis}
        \end{tikzpicture} &

        \cellcolor{orange!10}
        \begin{tikzpicture}
            \begin{axis}[mybarstyle2, title={SPD$_\text{a}$ - Camera}, ylabel={$\Delta_a$}, xlabel={}, ymin=0.8, ymax=3.2, ytick={1,2,3}, yticklabels={\phantom{0}1,2,3}]
            \addplot[fill=supervisedcolor, draw=black, bar shift=0pt] table[x expr=\coordindex, y=spd-a] {\mitigationablation};
            \end{axis}
        \end{tikzpicture} \\
        
    \end{tabular}
    \caption{\textbf{Impact of loss terms of the post-hoc adversarial mitigation.} Mitigation applied on top of pretrained supervised ConvNeXt-B~\cite{liu2022convnet}. 
    }
    \vspace{-10pt}
    \label{fig:mitigation_ablation}
\end{figure}

\cref{fig:mitigation_ablation} presents the results of an ablation study of the post-hoc adversarial mitigation loss. We observe that both terms of the loss related to metadata mitigation ($\beta$, $\gamma$) contribute to reducing sensitivity to metadata. Interestingly, we observe that the feature preservation term ($\alpha$) does not significantly affect the semantic prediction (SP). However, we attribute this to the fact that we use a simple linear layer as the mitigation component. When using a more complex mitigation component such as MLP, we observe that this term becomes crucial for preserving semantic information.

\begin{table}[!b]
    \centering
    \resizebox{\linewidth}{!}{
    \begin{tabular}{l@{\lsp}l@{\lsp}l@{\lsp}r@{\lsp}l}
        \toprule
         Attribute & Type & Metric & Num. classes & Description \\
        \midrule
        JPEG & processing & MP, SPD & 6 & amount of JPEG compression \\
        Resizing & processing & MP, SPD & 3 & amount of image resizing \\
        Sharpening & processing & MP, SPD & 3 & amount of sharpening \\
        Interpolation & processing & MP, SPD & 4 & type of interpolation during resizing \\
        \midrule
        Make & acquisition & MP & 9 & manufacturer of the camera \\
        Aperture & acquisition & MP & 17 & size of the opening in the lens \\
        Model & acquisition & MP & 88 & specific camera model used \\
        Exposure & acquisition & MP & 16 & amount of light captured by sensor \\
        ISO speed & acquisition & MP & 16 & camera sensor’s sensitivity to light \\
        Focal length & acquisition & MP & 13 & light convergence strength of lens \\
        Camera & acquisition & SPD & 2 & type of camera used \\
    \bottomrule
    \end{tabular}
}

    \caption{\textbf{Image processing and acquisition attributes used in diagnostic metrics.}}
    \label{tab:diagnostics}
\end{table}

\section{Implementation details}
\label{sec:implementation_supp}

\subsection{Diagnostic metrics}

For diagnostic metrics, we follow the setup of Ramos \etal~\cite{ramos2025traces} for all processing and acquisition attributes. An overview of the used processing and acquisition attributes, along with the number of classes in each case, is given in \cref{tab:diagnostics}; the full details can be found in the original paper~\cite{ramos2025traces}.

\subsection{Training on IN1k}

We train all of our models for $300$ epochs on IN1k. For ResNet50, we follow the A2 recipe of Wightman \etal~\cite{wightman2021resnet}. ViT-S/16 and ViT-B/16 are trained using the AdamW~\cite{loshchilov19adamw} optimizer with a batch size of $4096$ and a cosine-scheduled learning rate of $1e-3$, with a warmup of $40$ and $32$ epochs, respectively. 
We additionally utilize $0.1$ weight decay, $0.1$ dropout, RandomAugment~\cite{cubuk20randaugment}, and $0.5$ mixup~\cite{zhang18mixup}.

\begin{figure}[t]
    \centering
    \includegraphics[width=\textwidth]{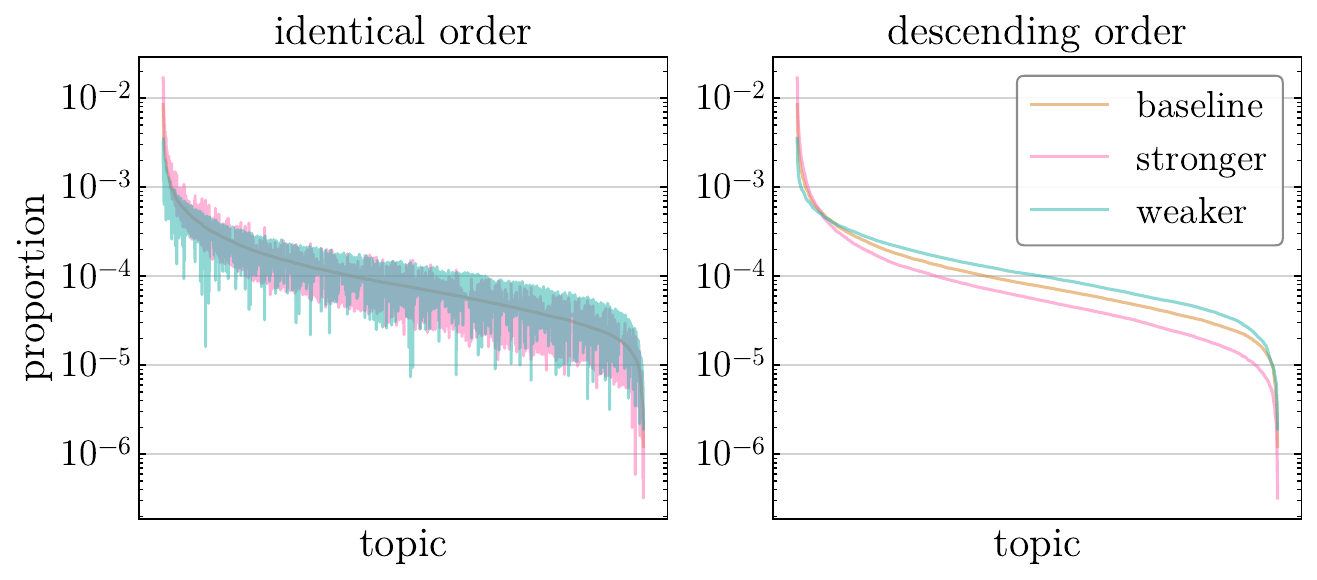}
    \caption{\textbf{Topic proportions within the Re-LAION-2B subsets with \textcolor{vlmcolor}{\textbf{baseline}}, \textcolor{vlmanothercolor}{\textbf{stronger}}, \textcolor{vlmanothercolor2}{\textbf{weaker}} correlations, discussed in Sec 3.4.} \emph{Left:} the values plotted at the same x-value correspond to the same topic. Topics are sorted in descending order by the proportion in the baseline subset. \emph{Right:} the values plotted at the same x-value correspond to the same topic \textit{rank}. Proportions are sorted in descending order.
    }
    \label{fig:topic_proportions}
\end{figure}

\subsection{Training on Re-LAION-2B}

We use BERTopic\cite{grootendorst2022bertopic} for topic extraction, where the captions are embedded with a sentence encoder\footnote{We use the model available at \url{https://huggingface.co/sentence-transformers/all-MiniLM-L6-v2}}, then dimensionality-reduced via UMAP\cite{mcinnes2018umap} and clustered into topics via HDBSCAN\cite{mcinnes2017hdbscan}. Topic embeddings are then calculated by averaging the embeddings of the assigned captions, which is used to categorize new captions via cosine similarity. By first fitting our topic model on 1M captions, then applying the provided auto topic reduction method, we obtain $6,238$ topics.
The proportions of topic representation are shown in \cref{fig:topic_proportions}. In general, topics are represented consistently across the three subsets, with topics highly represented in the baseline subset also being among the most represented topics in the other subsets.

All of our models are trained for 32 epochs using the AdamW\cite{loshchilov19adamw} optimizer with a batch size of 320 and a cosine-scheduled learning rate of $5e-4$, with a warmup of $10,000$ warmup steps. We additionally utilize 0.2 weight decay and automatic mixed precision.

\subsection{Details about synthesizing correlation between semantics and processing metadata}

We introduce controlled correlation between semantics and processing metadata in IN1k by either JPEG compression or resizing.

\noindent\emph{JPEG compression}: We consider a set $\cJ = \cQ \times \cC = \{(q_1, c_1), \cdots, (q_J, c_J)\}$ of $J$ different JPEG metadata labels defined by the quality $q \in \cQ$ and chroma subsampling $c \in \cC$. We use $\cQ = \{75, 77, \cdots, 95\}$ and $\cC = \{\text{4:2:0}, \text{4:4:4}\}$, which yields $J = 22$ distinct values of JPEG metadata labels.

\noindent\emph{Resizing}: We consider a set $\cR = \{r_1,  \cdots, r_R\} = \{\frac{1}{4}, \cdots, \frac{1}{1.5}, 1, 1.5, \cdots, 4\}$ where $r_i$ defines the change in image area while preserving the aspect ratio.

\subsection{Post-hoc adversarial mitigation}

Training of our post-hoc mitigation model proceeds in two steps, with each step optimized for $10$ epochs using the AdamW~\cite{loshchilov19adamw} optimizer with $1.0$ learning rate.
\begin{enumerate}
    \item \emph{Train the metadata prediction model $h$ while keeping the mitigation component frozen} by optimizing cross-entropy loss with supervision from metadata labels.
    \item \emph{Train the mitigation component $f$ while keeping the metadata prediction model frozen} by optimizing a loss
    \begin{equation}
        \begin{split}
            \mathcal{L} &= \underbrace{-\frac{\alpha}{N} \sum_{i=1}^{N}{\log\left(\sigma\left(\frac{f(x_i)^{\top}x_i}{\tau}\right)\right)}}_{\text{keep close to the original feature}} \\
                &\quad \underbrace{-\frac{\beta}{N^2} \sum_{i=1}^{N}{\sum_{j=1}^{N}{\mathbb{I}\left[m(x_i) = m(x_j) \land i \ne j\right]\log\left(1 - \sigma\left(\frac{f(x_i)^{\top}f(x_j)}{\tau}\right)\right)}}}_{\text{push apart features that share the same metadata label}} \\
                &\quad + \underbrace{\frac{\gamma}{N} \sum_{i=1}^{N}{\text{KL}(\mathbf{1} / M \parallel h(f(x_i)))}}_{\text{predict uniform distribution}},
        \end{split}
    \end{equation}
    where $\alpha$, $\beta$, and $\gamma$ are weights, $\sigma$ is a sigmoid function, $\tau$ is a temperature, $\mathbb{I}$ is an indicator function, $m(x)$ outputs the metadata label associated with $x$, KL is KL-divergence, and $\mathbf{1}$ is a vector of all ones.
\end{enumerate}
We initialize $f$ to an identity, while we initialize $h$ randomly. We fix $\alpha=40$, $\beta=20$, $\gamma = 1.0$, $\tau=0.1$, and repeat the steps $L=10$ times.

\clearpage
\section{Experiments with other backbones}
\label{sec:vit_supp}

\begin{figure}[t]
    \centering
    \input{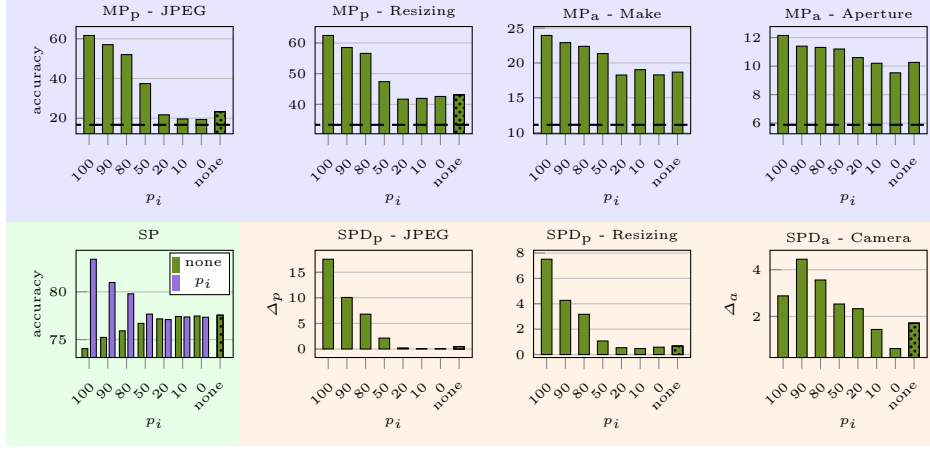}
\pgfplotsset{
    mybarstyle2/.style={
        ybar=1pt, 
        bar width=4pt,
        width=0.3\linewidth,
        height=3cm,
        enlarge x limits=0.1,
        xtick=data,
        xticklabels from table={\resnetimagenetjpegprobimageinversion}{p},
        x tick label style={font=\tiny, rotate=45, anchor=north east},
        title style={font=\tiny, yshift=-7pt},
        label style={font=\tiny},
        tick label style={font=\tiny},
        y label style={yshift=-4pt},
        x label style={yshift=5pt},
        ymajorgrids=true,
        xtick pos=bottom,
        ytick pos=left,
    }
}

\centering
    \begin{tabular}{cccc}

        \cellcolor{blue!10}
        \begin{tikzpicture}
            \begin{axis}[mybarstyle2, title={MP$_\text{p}$ - JPEG}, ylabel={accuracy}, xlabel={$p_{i}$}]
            \addplot[fill=supervisedcolor, draw=black, bar shift=0pt] table[x expr=\coordindex, y=jpeg] {\vitimagenetjpegprobimageinversion};
            \addplot[fill=supervisedcolor, draw=black, postaction={pattern=crosshatch dots}, bar shift=0pt, restrict x to domain=6.5:7.5] table[x expr=\coordindex, y=jpeg] {\vitimagenetjpegprobimageinversion};
            \addplot[
                black, 
                dashed, 
                thick, 
                sharp plot, %
                dash pattern=on 5pt off 3pt,
                mark=none,  %
                shorten <=-8pt, %
                shorten >=-8pt  %
            ] coordinates {(0, 16.67) (7, 16.67)};
            \end{axis}
        \end{tikzpicture} &
        
        \cellcolor{blue!10}
        \begin{tikzpicture}
            \begin{axis}[mybarstyle2, title={MP$_\text{p}$ - Resizing}, xlabel={$p_{i}$}, ylabel={\phantom{$\Delta_p$}}]
            \addplot[fill=supervisedcolor, draw=black, bar shift=0pt] table[x expr=\coordindex, y=resizing] {\vitimagenetjpegprobimageinversion};
            \addplot[fill=supervisedcolor, draw=black, postaction={pattern=crosshatch dots}, bar shift=0pt, restrict x to domain=6.5:7.5] table[x expr=\coordindex, y=resizing] {\vitimagenetjpegprobimageinversion};
            \addplot[
                black, 
                dashed, 
                thick, 
                sharp plot, %
                dash pattern=on 5pt off 3pt,
                mark=none,  %
                shorten <=-8pt, %
                shorten >=-8pt  %
            ] coordinates {(0, 33.33) (7, 33.33)};
            \end{axis}
        \end{tikzpicture} &

        \cellcolor{blue!10}
        \begin{tikzpicture}
            \begin{axis}[mybarstyle2, title={MP$_\text{a}$ - Make}, xlabel={$p_{i}$}]
            \addplot[fill=supervisedcolor, draw=black, bar shift=0pt] table[x expr=\coordindex, y=make] {\vitimagenetjpegprobimageinversion};
            \addplot[
                black, 
                dashed, 
                thick, 
                sharp plot, %
                dash pattern=on 5pt off 3pt,
                mark=none,  %
                shorten <=-8pt, %
                shorten >=-8pt  %
            ] coordinates {(0, 11.11) (7, 11.11)};
            \end{axis}
        \end{tikzpicture} &

        \cellcolor{blue!10}
        \begin{tikzpicture}
            \begin{axis}[mybarstyle2, title={MP$_\text{a}$ - Aperture}, xlabel={$p_{i}$}, ylabel={\phantom{$\Delta_a$}}]
            \addplot[fill=supervisedcolor, draw=black, bar shift=0pt] table[x expr=\coordindex, y=aperture] {\vitimagenetjpegprobimageinversion};
            \addplot[
                black, 
                dashed, 
                thick, 
                sharp plot, %
                dash pattern=on 5pt off 3pt,
                mark=none,  %
                shorten <=-8pt, %
                shorten >=-8pt  %
            ] coordinates {(0, 5.88) (7, 5.88)};
            \end{axis}
        \end{tikzpicture} \\

        \cellcolor{green!10}
        \begin{tikzpicture}
            \begin{axis}[mybarstyle2, title={SP\phantom{$_\text{p}$}}, ylabel={accuracy}, xlabel={$p_{i}$}, bar width=2.2pt, bar shift=0pt, legend style={font=\tiny, scale=0.6, /tikz/every even column/.append style={column sep=0.1cm}, inner sep=0.8pt, row sep=-1pt}, legend image code/.code={
                    \draw[#1, draw=none] (0cm,-0.1cm) rectangle (0.1cm,0.1cm); %
                }]
            \addplot[draw=none, fill=none, bar shift=0pt, forget plot] table[x expr=\coordindex, y=val-knn] {\vitimagenetjpegprobimageinversion};
            \addplot[fill=supervisedcolor, draw=black, bar shift=-1.6pt, restrict x to domain=-0.5:6.5] table[x expr=\coordindex, y=val-knn] {\vitimagenetjpegprobimageinversion};\addlegendentry{none};
            \addplot[fill=probcolor, draw=black, bar shift=1.6pt, restrict x to domain=-0.5:6.5] table[x expr=\coordindex, y=val-bias]{\vitimagenetjpegprobimageinversion};\addlegendentry{$p_i$};
            \addplot[fill=supervisedcolor, draw=black, bar shift=0pt, restrict x to domain=6.5:7.5, forget plot] table[x expr=\coordindex, y=val-knn] {\vitimagenetjpegprobimageinversion};
            \addplot[fill=supervisedcolor, draw=black, postaction={pattern=crosshatch dots}, bar shift=0pt, restrict x to domain=6.5:7.5, forget plot] table[x expr=\coordindex, y=val-knn] {\vitimagenetjpegprobimageinversion};
            \end{axis}
        \end{tikzpicture} &

        \cellcolor{orange!10}
        \begin{tikzpicture}
            \begin{axis}[mybarstyle2, title={SPD$_\text{p}$ - JPEG}, ylabel={$\Delta_p$}, xlabel={$p_{i}$}]
            \addplot[fill=supervisedcolor, draw=black, bar shift=0pt] table[x expr=\coordindex, y=jpeg] {\vitimagenetjpegprobimagesemantics};
            \addplot[fill=supervisedcolor, draw=black, postaction={pattern=crosshatch dots}, bar shift=0pt, restrict x to domain=6.5:7.5] table[x expr=\coordindex, y=jpeg] {\vitimagenetjpegprobimagesemantics};
            \end{axis}
        \end{tikzpicture} &

        \cellcolor{orange!10}
        \begin{tikzpicture}
            \begin{axis}[mybarstyle2, title={SPD$_\text{p}$ - Resizing}, xlabel={$p_{i}$}, ytick={0,2,4,6,8}, yticklabels={\phantom{0}0,2,4,6,8}]
            \addplot[fill=supervisedcolor, draw=black, bar shift=0pt] table[x expr=\coordindex, y=resizing] {\vitimagenetjpegprobimagesemantics};
            \addplot[fill=supervisedcolor, draw=black, postaction={pattern=crosshatch dots}, bar shift=0pt, restrict x to domain=6.5:7.5] table[x expr=\coordindex, y=resizing] {\vitimagenetjpegprobimagesemantics};
            \end{axis}
        \end{tikzpicture} &

        \cellcolor{orange!10}
        \begin{tikzpicture}
            \begin{axis}[mybarstyle2, title={SPD$_\text{a}$ - Camera}, ylabel={$\Delta_a$}, xlabel={$p_{i}$}, ytick={2,4}, yticklabels={\phantom{0}2,4}]
            \addplot[fill=supervisedcolor, draw=black, bar shift=0pt] table[x expr=\coordindex, y=spd-a] {\vitimagenetjpegprobimagesemantics};
            \addplot[fill=supervisedcolor, draw=black, postaction={pattern=crosshatch dots}, bar shift=0pt, restrict x to domain=6.5:7.5] table[x expr=\coordindex, y=spd-a] {\vitimagenetjpegprobimagesemantics};
            \end{axis}
        \end{tikzpicture} \\
        
    \end{tabular}
    \caption{\textbf{Impact of correlations between semantics and processing metadata on the model's sensitivity to metadata.
    } ViT-S/16, instead of ResNet50 (\cref{fig:imagenet_image_prob}), trained on different versions of IN1k with 
    manipulated JPEG labels. 
    The larger the $p_i$, the stronger the correlation between JPEG labels and semantic labels. 
    \emph{None}: model trained on original IN1k. 
    Difference between $p_i=0$ and \emph{none}: the former has no JPEG-semantics correlations, while the latter has the original IN1k correlations.
    SP: two types of test sets are used, the original one (none, \textcolor{supervisedcolor}{green} color) which is fixed across values of $p_i$, and 7 different versions (\textcolor{probcolor}{purple} color) created by manipulating the JPEG labels of the original test set in the same way as the corresponding training set for each value of $p_i$.
    }
    \label{fig:imagenet_image_prob_vit}
\end{figure}

In \cref{fig:imagenet_image_prob_vit,fig:imagenet_image_prob_vit_b}, we present the results of controlling correlations between semantics and processing metadata, as in \cref{sec:processing_correlations}, but for ViT-S/16 and ViT-B/16, respectively, instead of ResNet50. We observe that the introduction of JPEG-semantics correlation during training makes the model more sensitive to metadata, same as we observed for ResNet50 in \cref{fig:imagenet_image_prob}. This shows that our observations are not model-dependent and that models of different architectures exhibit the same behaviour.

Additionally, \cref{fig:clip_training_vits,fig:clip_training_convnextt} show results parallel to our experiment in \cref{sec:causes_laion}, where we train a model using CLIP loss~\cite{radford2021learning} on subsets of Re-LAION-2B with different amount of correlations between semantics and acquisition metadata. This time we use ViT-S/16 and ConvNeXt-T backbones, instead of ResNet50. Results are consistent, as training with stronger correlations between semantics and  acquisition metadata leads to worse SPD and SP, with the opposite being true for training with weaker correlations.

\begin{figure}[b]
    \centering
    \input{figures/pgfplotsdata}
\pgfplotsset{
    mybarstyle2/.style={
        ybar=1pt, 
        bar width=4pt,
        width=0.3\linewidth,
        height=3cm,
        enlarge x limits=0.1,
        xtick=data,
        xticklabels from table={\resnetimagenetjpegprobimageinversion}{p},
        x tick label style={font=\tiny, rotate=45, anchor=north east},
        title style={font=\tiny, yshift=-7pt},
        label style={font=\tiny},
        tick label style={font=\tiny},
        y label style={yshift=-4pt},
        x label style={yshift=5pt},
        ymajorgrids=true,
        xtick pos=bottom,
        ytick pos=left,
    }
}

\centering
    \begin{tabular}{cccc}

        \cellcolor{blue!10}
        \begin{tikzpicture}
            \begin{axis}[mybarstyle2, title={MP$_\text{p}$ - JPEG}, ylabel={accuracy}, xlabel={$p_{i}$}]
            \addplot[fill=supervisedcolor, draw=black, bar shift=0pt] table[x expr=\coordindex, y=jpeg] {\vitbimagenetjpegprobimageinversion};
            \addplot[fill=supervisedcolor, draw=black, postaction={pattern=crosshatch dots}, bar shift=0pt, restrict x to domain=6.5:7.5] table[x expr=\coordindex, y=jpeg] {\vitbimagenetjpegprobimageinversion};
            \addplot[
                black, 
                dashed, 
                thick, 
                sharp plot, %
                dash pattern=on 5pt off 3pt,
                mark=none,  %
                shorten <=-8pt, %
                shorten >=-8pt  %
            ] coordinates {(0, 16.67) (7, 16.67)};
            \end{axis}
        \end{tikzpicture} &
        
        \cellcolor{blue!10}
        \begin{tikzpicture}
            \begin{axis}[mybarstyle2, title={MP$_\text{p}$ - Resizing}, xlabel={$p_{i}$}, ylabel={\phantom{$\Delta_p$}}]
            \addplot[fill=supervisedcolor, draw=black, bar shift=0pt] table[x expr=\coordindex, y=resizing] {\vitbimagenetjpegprobimageinversion};
            \addplot[fill=supervisedcolor, draw=black, postaction={pattern=crosshatch dots}, bar shift=0pt, restrict x to domain=6.5:7.5] table[x expr=\coordindex, y=resizing] {\vitbimagenetjpegprobimageinversion};
            \addplot[
                black, 
                dashed, 
                thick, 
                sharp plot, %
                dash pattern=on 5pt off 3pt,
                mark=none,  %
                shorten <=-8pt, %
                shorten >=-8pt  %
            ] coordinates {(0, 33.33) (7, 33.33)};
            \end{axis}
        \end{tikzpicture} &

        \cellcolor{blue!10}
        \begin{tikzpicture}
            \begin{axis}[mybarstyle2, title={MP$_\text{a}$ - Make}, xlabel={$p_{i}$}]
            \addplot[fill=supervisedcolor, draw=black, bar shift=0pt] table[x expr=\coordindex, y=make] {\vitbimagenetjpegprobimageinversion};
            \addplot[
                black, 
                dashed, 
                thick, 
                sharp plot, %
                dash pattern=on 5pt off 3pt,
                mark=none,  %
                shorten <=-8pt, %
                shorten >=-8pt  %
            ] coordinates {(0, 11.11) (7, 11.11)};
            \end{axis}
        \end{tikzpicture} &

        \cellcolor{blue!10}
        \begin{tikzpicture}
            \begin{axis}[mybarstyle2, title={MP$_\text{a}$ - Aperture}, xlabel={$p_{i}$}, ylabel={\phantom{$\Delta_a$}}]
            \addplot[fill=supervisedcolor, draw=black, bar shift=0pt] table[x expr=\coordindex, y=aperture] {\vitbimagenetjpegprobimageinversion};
            \addplot[
                black, 
                dashed, 
                thick, 
                sharp plot, %
                dash pattern=on 5pt off 3pt,
                mark=none,  %
                shorten <=-8pt, %
                shorten >=-8pt  %
            ] coordinates {(0, 5.88) (7, 5.88)};
            \end{axis}
        \end{tikzpicture} \\

        \cellcolor{green!10}
        \begin{tikzpicture}
            \begin{axis}[mybarstyle2, title={SP\phantom{$_\text{p}$}}, ylabel={accuracy}, xlabel={$p_{i}$}, bar width=2.2pt, bar shift=0pt, legend style={font=\tiny, scale=0.6, /tikz/every even column/.append style={column sep=0.1cm}, inner sep=0.8pt, row sep=-1pt}, legend image code/.code={
                    \draw[#1, draw=none] (0cm,-0.1cm) rectangle (0.1cm,0.1cm); %
                }]
            \addplot[draw=none, fill=none, bar shift=0pt, forget plot] table[x expr=\coordindex, y=val-knn] {\vitbimagenetjpegprobimageinversion};
            \addplot[fill=supervisedcolor, draw=black, bar shift=-1.6pt, restrict x to domain=-0.5:6.5] table[x expr=\coordindex, y=val-knn] {\vitbimagenetjpegprobimageinversion};\addlegendentry{none};
            \addplot[fill=probcolor, draw=black, bar shift=1.6pt, restrict x to domain=-0.5:6.5] table[x expr=\coordindex, y=val-bias]{\vitbimagenetjpegprobimageinversion};\addlegendentry{$p_i$};
            \addplot[fill=supervisedcolor, draw=black, bar shift=0pt, restrict x to domain=6.5:7.5, forget plot] table[x expr=\coordindex, y=val-knn] {\vitbimagenetjpegprobimageinversion};
            \addplot[fill=supervisedcolor, draw=black, postaction={pattern=crosshatch dots}, bar shift=0pt, restrict x to domain=6.5:7.5, forget plot] table[x expr=\coordindex, y=val-knn] {\vitbimagenetjpegprobimageinversion};
            \end{axis}
        \end{tikzpicture} &

        \cellcolor{orange!10}
        \begin{tikzpicture}
            \begin{axis}[mybarstyle2, title={SPD$_\text{p}$ - JPEG}, ylabel={$\Delta_p$}, xlabel={$p_{i}$}, ymin=0]
            \addplot[fill=supervisedcolor, draw=black, bar shift=0pt] table[x expr=\coordindex, y=jpeg] {\vitbimagenetjpegprobimagesemantics};
            \addplot[fill=supervisedcolor, draw=black, postaction={pattern=crosshatch dots}, bar shift=0pt, restrict x to domain=6.5:7.5] table[x expr=\coordindex, y=jpeg] {\vitbimagenetjpegprobimagesemantics};
            \end{axis}
        \end{tikzpicture} &

        \cellcolor{orange!10}
        \begin{tikzpicture}
            \begin{axis}[mybarstyle2, title={SPD$_\text{p}$ - Resizing}, xlabel={$p_{i}$}, ymin=1.5, ymax=9.5, ytick={2,4,6,8}, yticklabels={\phantom{0}2,4,6,8}]
            \addplot[fill=supervisedcolor, draw=black, bar shift=0pt] table[x expr=\coordindex, y=resizing] {\vitbimagenetjpegprobimagesemantics};
            \addplot[fill=supervisedcolor, draw=black, postaction={pattern=crosshatch dots}, bar shift=0pt, restrict x to domain=6.5:7.5] table[x expr=\coordindex, y=resizing] {\vitbimagenetjpegprobimagesemantics};
            \end{axis}
        \end{tikzpicture} &

        \cellcolor{orange!10}
        \begin{tikzpicture}
            \begin{axis}[mybarstyle2, title={SPD$_\text{a}$ - Camera}, ylabel={$\Delta_a$}, xlabel={$p_{i}$}, ymin=2.8, ymax=6.5, ytick={4,6}, yticklabels={\phantom{0}4,6}]
            \addplot[fill=supervisedcolor, draw=black, bar shift=0pt] table[x expr=\coordindex, y=spd-a] {\vitbimagenetjpegprobimagesemantics};
            \addplot[fill=supervisedcolor, draw=black, postaction={pattern=crosshatch dots}, bar shift=0pt, restrict x to domain=6.5:7.5] table[x expr=\coordindex, y=spd-a] {\vitbimagenetjpegprobimagesemantics};
            \end{axis}
        \end{tikzpicture} \\
        
    \end{tabular}
    \caption{\textbf{Impact of correlations between semantics and processing metadata on the model's sensitivity to metadata.
    } ViT-B/16, instead of ResNet50 (\cref{fig:imagenet_image_prob}), trained on different versions of IN1k with 
    manipulated JPEG labels. 
    The larger the $p_i$, the stronger the correlation between JPEG labels and semantic labels. 
    \emph{None}: model trained on original IN1k. 
    Difference between $p_i=0$ and \emph{none}: the former has no JPEG-semantics correlations, while the latter has the original IN1k correlations.
    SP: two types of test sets are used, the original one (none, \textcolor{supervisedcolor}{green} color) which is fixed across values of $p_i$, and 7 different versions (\textcolor{probcolor}{purple} color) created by manipulating the JPEG labels of the original test set in the same way as the corresponding training set for each value of $p_i$.
    }
    \label{fig:imagenet_image_prob_vit_b}
\end{figure}

\begin{figure}[t]
    \centering
    \input{figures/pgfplotsdata}
\pgfplotsset{
    mybarstyle/.style={
        ybar=1pt, 
        width=0.29\linewidth,
        height=3cm,
        enlarge x limits=0.2,
        ymin=0, 
        symbolic x coords={stronger, baseline, weaker},
        xtick=data,
        x tick label style={rotate=45, anchor=east, font=\tiny, scale=0.75},
        title style={font=\tiny, yshift=-7pt},
        label style={font=\tiny},
        tick label style={font=\tiny},
        y label style={yshift=-4pt},
        ymajorgrids=true,
        xtick pos=bottom,
        ytick pos=left,
    },
    stronger/.style={restrict expr to domain={mod(\coordindex,3)}{0:0}},
    baseline/.style={restrict expr to domain={mod(\coordindex,3)}{1:1}},
    weaker/.style={restrict expr to domain={mod(\coordindex,3)}{2:2}},
}

\centering
    \begin{tabular}{cccc}
    
        \cellcolor{blue!10}
        
        \begin{tikzpicture}
            \begin{axis}[mybarstyle, title={MP$_\text{p}$ - JPEG}, ylabel={accuracy}, ymin=10]
            \addplot[fill=vlmcolor, draw=black] table[x=subset, y=jpeg] {\customclipcomparisonvits};
            \addplot[
                black, 
                dashed, 
                thick, 
                sharp plot, %
                dash pattern=on 5pt off 3pt,
                mark=none,  %
                shorten <=-8pt, %
                shorten >=-8pt  %
            ] coordinates {(stronger, 16.67) (weaker, 16.67)};
            \end{axis}
        \end{tikzpicture} &

        \cellcolor{blue!10}
        \begin{tikzpicture}
            \begin{axis}[mybarstyle, title={MP$_\text{p}$ - Resizing}, ylabel={\phantom{$\Delta_p$}}, ymin=30]
            \addplot[fill=vlmcolor, draw=black] table[x=subset, y=resizing] {\customclipcomparisonvits};
            \addplot[
                black, 
                dashed, 
                thick, 
                sharp plot,
                dash pattern=on 5pt off 3pt,
                mark=none,
                shorten <=-8pt,
                shorten >=-8pt 
            ] coordinates {(stronger, 33.33) (weaker, 33.33)};
            \end{axis}
        \end{tikzpicture} &

        \cellcolor{blue!10}
        \begin{tikzpicture}
            \begin{axis}[mybarstyle, title={MP$_\text{a}$ - Make}, ymin=5, ymax=35]
            \addplot[fill=vlmcolor, draw=black] table[x=subset, y=make] {\customclipcomparisonvits};
            \addplot[
                black, 
                dashed, 
                thick, 
                sharp plot, %
                dash pattern=on 5pt off 3pt,
                mark=none,  %
                shorten <=-8pt, %
                shorten >=-8pt  %
            ] coordinates {(stronger, 11.11) (weaker, 11.11)};
            \end{axis}
        \end{tikzpicture} &

        \cellcolor{blue!10}
        \begin{tikzpicture}
            \begin{axis}[mybarstyle, title={MP$_\text{a}$ - Aperture}, ymin=2, ymax=17, ylabel={\phantom{$\Delta_a$}}]
            \addplot[fill=vlmcolor, draw=black] table[x=subset, y=aperture] {\customclipcomparisonvits};
            \addplot[
                black, 
                dashed, 
                thick, 
                sharp plot, %
                dash pattern=on 5pt off 3pt,
                mark=none,  %
                shorten <=-8pt, %
                shorten >=-8pt  %
            ] coordinates {(stronger, 5.88) (weaker, 5.88)};
            \end{axis}
        \end{tikzpicture} \\

        \cellcolor{green!10}
        \begin{tikzpicture}
            \begin{axis}[mybarstyle, title={SP\phantom{$_\text{p}$}}, ylabel={accuracy}, ymin=42]
            \addplot[fill=vlmcolor, draw=black] table[x=subset, y=val-knn] {\customclipcomparisonvits};
            \end{axis}
        \end{tikzpicture} &

        \cellcolor{orange!10}
        \begin{tikzpicture}
            \begin{axis}[mybarstyle, title={SPD$_\text{p}$ - JPEG}, ylabel={$\Delta_p$}, ymin=11]
            \addplot[fill=vlmcolor, draw=black] table[x=subset, y=jpeg-sem] {\customclipcomparisonvits};
            \end{axis}
        \end{tikzpicture} &

        \cellcolor{orange!10}
        \begin{tikzpicture}
            \begin{axis}[mybarstyle, title={SPD$_\text{p}$ - Resizing}, ymin=15]
            \addplot[fill=vlmcolor, draw=black] table[x=subset, y=resizing-sem] {\customclipcomparisonvits};
            \end{axis}
        \end{tikzpicture}  &

        \cellcolor{orange!10}
        \begin{tikzpicture}
            \begin{axis}[mybarstyle, title={SPD$_\text{a}$ - Camera}, ylabel={$\Delta_a$}, ymin=3.4, ytick={4, 5}, yticklabels={\phantom{0}4,5}]
            \addplot[fill=vlmcolor, draw=black] table[x=subset, y=spd-a] {\customclipcomparisonvits};
            \end{axis}
        \end{tikzpicture} \\

    \end{tabular}
    \caption{
    \textbf{Impact of correlations between semantics and acquisition metadata on the model's sensitivity to metadata.} ViT-S/16, instead of ResNet50 (\cref{fig:clip_training}), trained from scratch using CLIP loss~\cite{radford2021learning} on three Re-LAION-2B subsets.
    }
    \label{fig:clip_training_vits}
\end{figure}

\begin{figure}[t]
    \centering
    \input{figures/pgfplotsdata}
\pgfplotsset{
    mybarstyle/.style={
        ybar=1pt, 
        width=0.29\linewidth,
        height=3cm,
        enlarge x limits=0.2,
        ymin=0, 
        symbolic x coords={stronger, baseline, weaker},
        xtick=data,
        x tick label style={rotate=45, anchor=east, font=\tiny, scale=0.75},
        title style={font=\tiny, yshift=-7pt},
        label style={font=\tiny},
        tick label style={font=\tiny},
        y label style={yshift=-4pt},
        ymajorgrids=true,
        xtick pos=bottom,
        ytick pos=left,
    },
    stronger/.style={restrict expr to domain={mod(\coordindex,3)}{0:0}},
    baseline/.style={restrict expr to domain={mod(\coordindex,3)}{1:1}},
    weaker/.style={restrict expr to domain={mod(\coordindex,3)}{2:2}},
}

\centering
    \begin{tabular}{cccc}
    
        \cellcolor{blue!10}
        
        \begin{tikzpicture}
            \begin{axis}[mybarstyle, title={MP$_\text{p}$ - JPEG}, ylabel={accuracy}, ymin=10]
            \addplot[fill=vlmcolor, draw=black] table[x=subset, y=jpeg] {\customclipcomparisonconvnextt};
            \addplot[
                black, 
                dashed, 
                thick, 
                sharp plot, %
                dash pattern=on 5pt off 3pt,
                mark=none,  %
                shorten <=-8pt, %
                shorten >=-8pt  %
            ] coordinates {(stronger, 16.67) (weaker, 16.67)};
            \end{axis}
        \end{tikzpicture} &

        \cellcolor{blue!10}
        \begin{tikzpicture}
            \begin{axis}[mybarstyle, title={MP$_\text{p}$ - Resizing}, ylabel={\phantom{$\Delta_p$}}, ymin=30]
            \addplot[fill=vlmcolor, draw=black] table[x=subset, y=resizing] {\customclipcomparisonconvnextt};
            \addplot[
                black, 
                dashed, 
                thick, 
                sharp plot,
                dash pattern=on 5pt off 3pt,
                mark=none,
                shorten <=-8pt,
                shorten >=-8pt 
            ] coordinates {(stronger, 33.33) (weaker, 33.33)};
            \end{axis}
        \end{tikzpicture} &

        \cellcolor{blue!10}
        \begin{tikzpicture}
            \begin{axis}[mybarstyle, title={MP$_\text{a}$ - Make}, ymax=35]
            \addplot[fill=vlmcolor, draw=black] table[x=subset, y=make] {\customclipcomparisonconvnextt};
            \addplot[
                black, 
                dashed, 
                thick, 
                sharp plot, %
                dash pattern=on 5pt off 3pt,
                mark=none,  %
                shorten <=-8pt, %
                shorten >=-8pt  %
            ] coordinates {(stronger, 11.11) (weaker, 11.11)};
            \end{axis}
        \end{tikzpicture} &

        \cellcolor{blue!10}
        \begin{tikzpicture}
            \begin{axis}[mybarstyle, title={MP$_\text{a}$ - Aperture}, ymin=2, ymax=17, ylabel={\phantom{$\Delta_a$}}]
            \addplot[fill=vlmcolor, draw=black] table[x=subset, y=aperture] {\customclipcomparisonconvnextt};
            \addplot[
                black, 
                dashed, 
                thick, 
                sharp plot, %
                dash pattern=on 5pt off 3pt,
                mark=none,  %
                shorten <=-8pt, %
                shorten >=-8pt  %
            ] coordinates {(stronger, 5.88) (weaker, 5.88)};
            \end{axis}
        \end{tikzpicture} \\

        \cellcolor{green!10}
        \begin{tikzpicture}
            \begin{axis}[mybarstyle, title={SP\phantom{$_\text{p}$}}, ylabel={accuracy}, ymin=36, ymax=63]
            \addplot[fill=vlmcolor, draw=black] table[x=subset, y=val-knn] {\customclipcomparisonconvnextt};
            \end{axis}
        \end{tikzpicture} &

        \cellcolor{orange!10}
        \begin{tikzpicture}
            \begin{axis}[mybarstyle, title={SPD$_\text{p}$ - JPEG}, ylabel={$\Delta_p$}, ymin=12.5]
            \addplot[fill=vlmcolor, draw=black] table[x=subset, y=jpeg-sem] {\customclipcomparisonconvnextt};
            \end{axis}
        \end{tikzpicture} &

        \cellcolor{orange!10}
        \begin{tikzpicture}
            \begin{axis}[mybarstyle, title={SPD$_\text{p}$ - Resizing}, ymin=17]
            \addplot[fill=vlmcolor, draw=black] table[x=subset, y=resizing-sem] {\customclipcomparisonconvnextt};
            \end{axis}
        \end{tikzpicture}  &

        \cellcolor{orange!10}
        \begin{tikzpicture}
            \begin{axis}[mybarstyle, title={SPD$_\text{a}$ - Camera}, ylabel={$\Delta_a$}, ymin=1.7, ymax=8.3, ytick={2,4,6,8}, yticklabels={\phantom{0}2,4,6,8}]
            \addplot[fill=vlmcolor, draw=black] table[x=subset, y=spd-a] {\customclipcomparisonconvnextt};
            \end{axis}
        \end{tikzpicture} \\

    \end{tabular}
    \caption{
    \textbf{Impact of correlations between semantics and acquisition metadata on the model's sensitivity to metadata.} ConvNeXt-T, instead of ResNet50 (\cref{fig:clip_training}), trained from scratch using CLIP loss~\cite{radford2021learning} on three Re-LAION-2B subsets.
    }
    \label{fig:clip_training_convnextt}
\end{figure}

\end{document}